\pdfoutput=1

\documentclass[11pt]{article}
\usepackage{float}

\usepackage[table]{xcolor}
\usepackage[a-1b]{pdfx}

\usepackage{framed}
\usepackage{lmodern}
\usepackage{mdwlist}
\usepackage{siunitx}
\usepackage{latexsym}
\usepackage{colortbl}
\usepackage{nicefrac}
\usepackage{booktabs}
\usepackage{fnpct}
\usepackage{amsfonts}
\usepackage[T1]{fontenc}
\usepackage{bold-extra}
\usepackage{amsmath}
\usepackage{amssymb}
\usepackage{bm}
\usepackage{graphicx}
\usepackage{mathtools}
\usepackage{microtype}
\usepackage{multirow}
\usepackage{multicol}
\usepackage{xpatch}
\usepackage{latexsym,comment}
\usepackage[normalem]{ulem}

\newcommand*{\missingreference}{{\Huge \colorbox{red}{?reference?}}}
\newcommand*{\missingcitation}{{\Huge \colorbox{red}{?citation?}}}

\makeatletter
\xpatchcmd{\@setref}{\bfseries}{\missingreference}{}{}
\def\@citex[#1]#2{\leavevmode
    \let\@citea\@empty
    \@cite{\@for\@citeb:=#2\do
        {\@citea\def\@citea{,\penalty\@m\ }%
            \edef\@citeb{\expandafter\@firstofone\@citeb\@empty}%
            \if@filesw\immediate\write\@auxout{\string\citation{\@citeb}}\fi
            \@ifundefined{b@\@citeb}{\hbox{\reset@font\missingcitation}%
                \G@refundefinedtrue
                \@latex@warning
                {Citation `\@citeb' on page \thepage \space undefined}}%
            {\@cite@ofmt{\csname b@\@citeb\endcsname}}}}{#1}}
\makeatother

\newcommand{\clip}[0]{\abr{clip}\xspace}
\newcommand{\shp}[0]{SHP\xspace}
\newcommand{\mm}[0]{LLM\xspace}
\newcommand{\dpo}[0]{DPO\xspace}
\newcommand{\sft}{SFT\xspace}
\newcommand{\fs}[0]{FS\xspace}

\newcommand{\gem}[1]{\mbox{\textsc{gem}}}
\newcommand{\abr}[1]{\textsc{#1}}

\newcommand{\g}{\, | \,}

\newcommand{\hidetext}[1]{}
\newcommand{\ignore}[1]{}
\newif\ifcomment
\commenttrue

\ifcomment
    \newcommand{\pinaforecomment}[3]{\colorbox{#1}{\parbox{.8\linewidth}{#2: #3}}}

    \newcommand{\prtodo}[1]{\pinaforecomment{lightblue}{pr}{#1}}
    \newcommand{\prtodoi}[1]{\pinaforecomment{lightblue}{pr}{#1}}
\else
    \newcommand{\pinaforecomment}[3]{}
    \newcommand{\prtodo}[1]{}
    \newcommand{\prtodoi}[1]{}
\fi

\newcommand{\smallurl}[1]{ \begin{tiny}\url{#1}\end{tiny}}

\definecolor{lightblue}{HTML}{3cc7ea}
\definecolor{CUgold}{HTML}{CFB87C}
\definecolor{grey}{rgb}{0.95,0.95,0.95}
\definecolor{ceil}{rgb}{0.57, 0.63, 0.81}
\definecolor{UMDred}{HTML}{ed1c24}
\definecolor{UMDyellow}{HTML}{ffc20e}

\newcommand{\qa}[0]{\abr{qa}}

\usepackage[]{acl2023}

\usepackage{xspace}
\usepackage{pgfplots}
\usepackage{dsfont}
\DeclareUnicodeCharacter{2212}{−}
\usepgfplotslibrary{groupplots,dateplot}
\usetikzlibrary{patterns,shapes.arrows}
\pgfplotsset{compat=newest}

\usepackage{tikzscale}
\usepackage{relsize}
\usepackage{amsmath,amssymb}
\newcommand{\probP}{\text{I\kern-0.15em P}}

\newcommand{\inference}{\textsc{PI}\xspace}
\newcommand{\generation}{\textsc{PT}\xspace}
\newcommand{\score}{$\Delta\textsc{PQ}$\xspace}

\usepackage{multirow, colortbl}

\usepackage{tabularx,booktabs}
\usepackage{makecell}

\usepackage[normalem]{ulem}
\useunder{\uline}{\ul}{}

\usepackage{graphicx}

\definecolor{ablation6}{HTML}{fcefed}
\definecolor{ablation_tie}{HTML}{fce3e1}

\definecolor{ablation5}{HTML}{fcd8d4}
\definecolor{ablation4}{HTML}{FBC3BC}
\definecolor{ablation3}{HTML}{F7A399}
\definecolor{ablation2}{HTML}{F38375}
\definecolor{ablation1}{HTML}{EF6351}

\usepackage[]{algpseudocode}
\usepackage[]{algorithm}
\usepackage{float}
\algtext*{EndFor}%
\algtext*{EndProcedure}%

\usepackage{booktabs}
\usepackage[normalem]{ulem}
\useunder{\uline}{\ul}{}

\usepackage{times}
\usepackage{latexsym}
\usepackage{adjustbox}

\usepackage[T1]{fontenc}
\usepackage{amsmath}
\usepackage{amssymb}
\usepackage{booktabs}
\usepackage{tikzscale}
\usepackage{amsmath}

\usepackage{multirow, colortbl}

\usepackage{tabularx,booktabs}
\usepackage{makecell}
\usepackage{multirow}
\usepackage{scalerel,xparse}

\usepackage{cleveref}
\usepackage{microtype}
\usepackage[most]{tcolorbox}

\usepackage{enumitem}
\crefformat{section}{\S#2#1#3}
\crefformat{subsection}{\S#2#1#3}
\crefformat{subsubsection}{\S#2#1#3}

\definecolor{bggray}{rgb}{0.95, 0.95, 0.95}
\usepackage[%
    framemethod=tikz,
    skipbelow=\topskip,
    skipabove=\topskip
]{mdframed}
\mdfsetup{%
    leftmargin=0pt,
    rightmargin=0pt,
    backgroundcolor=bggray,
    middlelinecolor=black,
    roundcorner=3
}

\definecolor{SkyBlue}{rgb}{0.53, 0.81, 0.92} 
\definecolor{OliveGreen}{rgb}{0.05, 0.75, 0.24}
\definecolor{BrickRed}{rgb}{0.8, 0.25, 0.33} 

\newcommand{\colorBoxColor}{SkyBlue}

\newtcolorbox[list inside=prompt,auto counter,number within=section]{prompt}[1][]{
    colbacktitle=black!60,
    fonttitle=\small,
    coltitle=white,
    fontupper=\footnotesize,
    boxsep=3pt,
    left=0pt,
    right=0pt,
    top=0pt,
    bottom=0pt,
    boxrule=1pt,
    #1
}

\definecolor{UMDred}{HTML}{ed1c24}

\definecolor{yellowcolor}{HTML}{ffc20e}
\definecolor{redcolor}{HTML}{ff8f87}
\definecolor{bluecolor}{HTML}{8cd2f5}

\title{Whose Boat Does it Float?\\Improving Personalization in Preference Tuning via Inferred User Personas}

\author{Nishant Balepur$^{1}$ \hspace{0.5cm} Vishakh Padmakumar$^{2}$  \hspace{0.5cm} \textbf{Fumeng Yang}$^{1}$ \hspace{0.5cm} \textbf{Shi Feng}$^{3}$ \\ \hspace{0.5cm} \textbf{Rachel Rudinger}$^{1}$ \hspace{0.5cm} \textbf{Jordan Boyd-Graber}$^{1}$ \vspace{0.1cm}  \\
  $^{1}$University of Maryland \hspace{0.5cm}
  $^{2}$New York University \hspace{0.5cm}
  $^{3}$George Washington University \\
  \texttt{nbalepur@umd.edu} \hspace{0.5cm} \texttt{{jbg}@umiacs.umd.edu}
}

\begin{document}
\maketitle


\begin{abstract} {

\mm{}s are aligned to follow input instructions by learning which of two responses users prefer for a prompt.
However, such preference data do not convey \textit{why} users prefer responses that are chosen or rejected, so \mm{}s trained on these datasets cannot tailor responses to varied user needs.
To surface these parameters of
personalization, we apply \textit{abductive reasoning} to
preference data, inferring needs and interests of users,
i.e., personas, that may prefer either response.
%
%
We test this idea in two~steps: \textbf{Persona Inference (PI)}---abductively inferring personas of users who prefer chosen or
rejected outputs---and \textbf{Persona Tailoring (PT)}---training
models to tailor outputs to personas from \inference.~We~show:
1) \mm{}s infer personas accurately explaining why different users may prefer \textit{both} chosen or rejected outputs;
2) Training on preference data augmented with \inference personas via~\generation boosts personalization and generalizes to supporting user-written personas; and
3) Rejected response personas form harder personalization evaluations, showing \generation better aids users with uncommon preferences versus typical alignment methods.
We argue for an abductive view of preferences for personalization,
asking not only which response is better but when, why, and for whom.\footnote{\label{fn:datasets}Code and datasets are available at: \url{https://github.com/Pinafore/alignment-personalization}}}
\end{abstract}

\section{Every Preference Happens for a Reason}

Current methods for aligning large language models (\mm{}s) predominantly use preference data~\cite{ji2023ai}, created by finding: for a given prompt, which of two outputs do users prefer?
\mm{}s are then trained on this data via preference tuning methods like direct preference optimization~\cite[\dpo{}]{rafailov2024direct}, learning to give outputs like the majority \textit{chosen} response and unlike the~\textit{rejected} one.
This improves \mm{} outputs in dialogue~\cite{OpenAssistant}, question answering~\cite[\qa]{ELIFIVE}, and summarization~\cite{volske-etal-2017-tl}.

\setlength{\fboxsep}{-0.5pt}

\begin{figure*}[t]
    \centering
    \includegraphics[width=\linewidth]{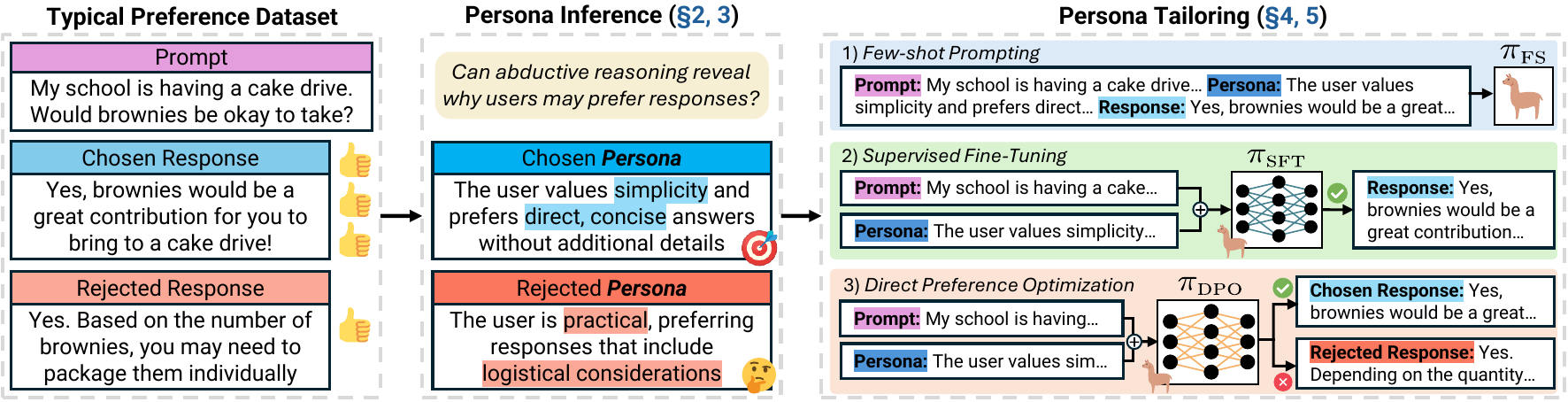}
    \vspace{-4.25ex}
    \caption{\small \textbf{Overview of this paper.} Preference data has a prompt, chosen response, and rejected response (left). Most users prefer the chosen response, but there are valid reasons and personas of users that may prefer either response, which we uncover via abductive reasoning in \textsc{Persona Inference} (\inference, middle). We then study \textsc{Persona Tailoring} (\generation) to use the personas for data augmentation in few-shot prompts, fine-tuning, and direct preference optimization---enhancing personalization (right).
    }
    \label{fig:main}
    \vspace{-1.25ex}
\end{figure*}

While preference datasets are valuable, they  assume chosen responses are universally better, failing to consider \textit{why} users prefer responses~\cite{joshi2025improving}.
In reality, some users genuinely prefer rejected outputs, even if their reasons are less common.
In Figure~\ref{fig:main} (left),~most users~prefer the chosen output for taking brownies to a cake sale.
While users valuing simplicity may prefer~this direct response, practical users may favor the rejected one, as it also gives packaging logistics (Figure~\ref{fig:main}, middle).
Since users prefer responses for varied reasons~\cite{kirk2024prism}, models trained on preference data should personalize outputs to meet these specific, individual needs~\cite{salemi-etal-2024-lamp}.

A common, interpretable method for personalization involves users specifying their needs and interests as an extra input (e.g., ``I like short answers'')---forming a system prompt \textbf{persona} to tailor model responses~\cite{zhang2024personalization}.
However, current preference dataset formats lack personas explaining why users prefer responses, overlooking signals for training personalized models~\cite{lee2024aligning}.

\begin{figure}[t]
    \centering
    \includegraphics[width=\linewidth]{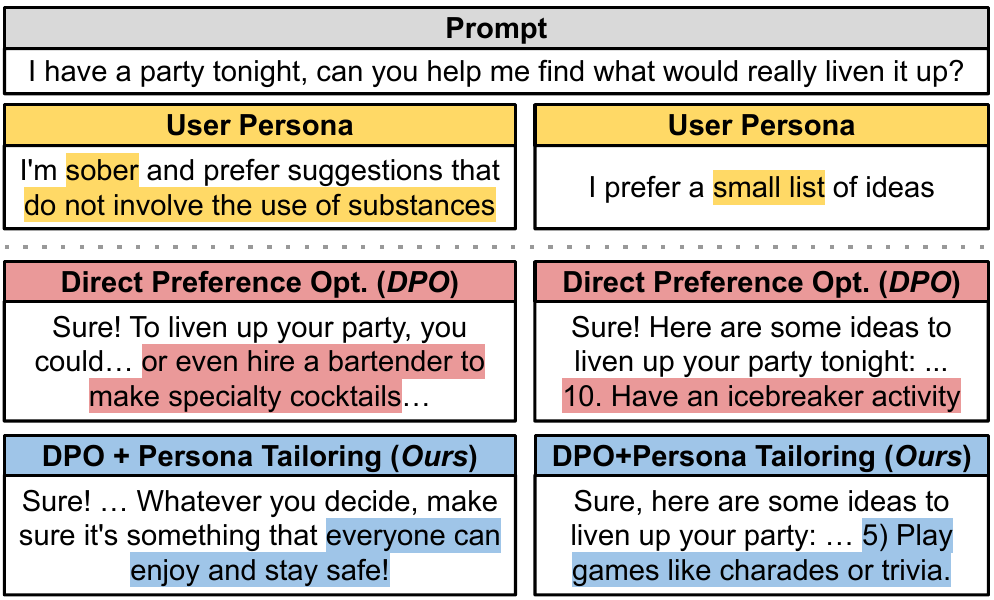}
    \setlength{\fboxsep}{0pt}
    \vspace{-4ex}
    \caption{\small Training methods with typical preference datasets like \colorbox{redcolor}{\strut DPO} cannot fully cater to a user's \colorbox{yellowcolor}{\strut specified personas}. To overcome this, we train models on preference data augmented with LLM-inferred personas, which we call \colorbox{bluecolor}{\strut persona tailoring}.}
    \label{fig:intro}
    \vspace{-1.6ex}
\end{figure}

The only way for models trained on existing preference dataset formats to use a persona is by adding it to the inference prompt, hoping the model treats it as an extra instruction to follow~\cite{deb-etal-2022-boosting}.
However, standard preference tuning methods like \dpo{} cannot naturally adapt to personas in this way.
In Fig~\ref{fig:intro}, \dpo{} tells a \colorbox{yellowcolor}{\strut sober} user to hire a \colorbox{redcolor}{\strut bartender} (left), while a user desiring a \colorbox{yellowcolor}{\strut small} list gets~\colorbox{redcolor}{\strut 10 ideas} (right); such responses are unhelpful, inappropriate, and do not tailor to the users' specified~personas.

As models' poor personalization stems from a lack of training on personas, we propose \textbf{abductive reasoning}~\cite{peirce1974collected} to augment preference training data with \mm{}-inferred personas.
Abduction infers hidden contexts that explain a given outcome~\cite{zhao-etal-2024-uncommonsense}.
Similarly, we adopt this reasoning to infer (i.e., adduce) hidden needs, interests, and traits of users, i.e., personas~\cite{tseng2024two}, that explain why chosen and rejected outputs may be preferred.
If \mm{}s surface valid personas of users who may prefer chosen or rejected responses, we can attach them to preference datasets and train models that tailor to user-specified needs, enhancing model personalization (Figure~\ref{fig:intro}, \colorbox{bluecolor}{\strut blue}).

We segment this idea into a two-task sequence: \textbf{1) Persona Inference (\inference}, \cref{section:persona_inference_setup}\textbf{):} adducing personas for preference data responses (Figure~\ref{fig:main}, middle); and \textbf{2) Persona Tailoring (\generation}, \cref{section:persona_generation_setup}\textbf{):} producing~outputs tailored to personas from \inference (Figure~\ref{fig:main}, right).

We first test \mm{}s in \inference on
dialogue~\cite{bai2022training}, \qa~\cite{ji2024beavertails}, and education~\cite{balepur2024smart} preference~data.
\mm{}~personas accurately convey different users who could prefer chosen \textit{or} rejected outputs; LLaMA-405B has 91\% accuracy, judged by GPT-4o with 90\% human agreement (\cref{subsection:accuracy}).
Further, chosen and rejected response personas are often judged as tied in quality~(\cref{subsection:preference_change}), and humans rate rejected ones as plausible but applying to fewer users (\cref{subsection:persona_qualitative}).
Thus, users~may prefer rejected outputs for uncommon but valid reasons.
Personas are also a useful content analysis tool to find differences in chosen and rejected responses; in
BeaverTails~\cite{ji2024beavertails}, chosen response personas describe users who are ``meticulous'', while rejected ones describe ``direct'' users, showing~these
labelers may prefer verbosity~\cite{zheng2024judging}.

As \inference yields accurate personas, we augment~preference data with
\inference personas using LLaMA-405B.
We then train LLaMA-8B on this new data for the reverse task of \generation: using prompts and inferred personas as inputs to give tailored responses (Figure~\ref{fig:main}, right).
We test three strategies: prompting~\cite[$\textsc{PT}_\textsc{fs}$]{10.5555/3495724.3495883}, fine-tuning~\cite[$\textsc{PT}_\textsc{sft}$]{JMLR:v25:23-0870}, and \dpo{}~\cite[$\textsc{PT}_\textsc{dpo}$]{rafailov2024direct}.

Each generation strategy largely boosts personalization when using \generation (\cref{subsection:add_personas}), with $\textsc{PT}_\textsc{dpo}$ being the strongest (\cref{subsection:ablation}).
Further, \textsc{\dpo{}} fine-tuned on preference datasets without personas cannot always tailor to personas during inference; notably,~$\textsc{PT}_\textsc{dpo}$ is judged as much stronger than \textsc{\dpo{}} on uncommon but still valid needs in the personas from rejected responses (\textbf{66\% average improvement in personalization}), showing rejected responses can form valuable, harder~evaluations for personalization (\cref{subsection:persona_type}).
Finally, eight users author 144 diverse personas and rate $\textsc{PT}_\textsc{dpo}$ and \textsc{\dpo{}} responses for~these personas (\cref{subsection:user_study}).
The same users find our $\textsc{PT}_\textsc{dpo}$ method personalizes more effectively to their written needs, showing models trained on realistic, \mm{}-inferred personas can generalize to real user-specific needs.

We argue for an abductive view of preferences, capturing not only which outputs users prefer but \textit{which users} and \textit{for what reasons}. In doing so, we can find valid user needs that may be overlooked in rejected responses (\cref{section:persona_inference_results}) and ensure \mm{}s support them (\cref{subsection:persona_type}).
It also improves personalization (\cref{section:persona_generation_results}), such as augmentation via \inference. Our contributions~are:\\
\noindent \textbf{1)} We study abductive reasoning on preference data via persona inference (\inference) to show \mm{}s can infer why users may prefer chosen and rejected outputs.\\
\noindent \textbf{2)} We release persona-augmented question answering, dialogue, and education preference datasets.\textsuperscript{\ref{fn:datasets}}\\
\noindent \textbf{3)} Persona tailoring---prompting and training on persona-augmented data---effectively personalizes to needs inferred by \inference and specified by real users.

\setlength{\fboxsep}{2pt}






\section{Persona Inference Setup} \label{section:persona_inference_setup}


The first step in our personalization method is inferring why users
may prefer each response in standard preference data (Fig~\ref{fig:main}, middle).
We use abduction~\cite{peirce1974collected}---which
explains outcomes---to infer persona-based explanations for why responses may be preferred via \textbf{\textsc{Persona Inference}~(\inference)}:

\noindent \textbf{$ \bullet \; \textsc{PI}(p, r_{1}, r_{2})
  \rightarrow \mathcal{P}_{1}$}: For prompt $p$ and responses $r_1$
and $r_2$, the \mm{} gives a persona $\mathcal{P}_1$ such that a user
described by $\mathcal{P}_1$ would prefer $r_1$ over~$r_2$.
If $r_1$ is the chosen response and $r_2$ is the rejected one,
$\mathcal{P}_1$ will describe a user preferring the chosen response,
and vice versa if $r_1$ is the rejected response.

Following \citet{chen2024persona}, $\mathcal{P}_1$ is~closest to a demographic persona, but we infer broad traits---like information needs, interests, or personalities---rather than
protected attributes (e.g., race) to curb
stereotyping~\cite{kantharuban2024stereotype}.
Our personas include no other constraints.
All $\mathcal{P}_1$ follow the format: ``The user is [attribute] and prefers [explanation of preference]'' for parsing.
This section gives our \inference
models~(\cref{subsection:persona_models}) and datasets~(\cref{subsection:persona_datasets}).

\subsection{Models} \label{subsection:persona_models}

We test nine \mm{}s in \inference: \textbf{Claude} \cite[Sonnet, Haiku, Opus]{anthropic2023claude}, \textbf{GPT} \cite[3.5, 4, 4o]{achiam2023gpt}, and \textbf{L}LaMA-3.1 Instruct \cite[8B, 70B, 405B] {dubey2024llama}.
We use 5-shot prompts in the format below, where highlights are generations:

{
\begin{prompt}[title={Prompt \thetcbcounter: Few-Shot Persona Inference Prompt}, label=prompt:full_prompt]
\texttt{Prompt:} $p$ \\
\texttt{Chosen Response:} $r_1$ \\
\texttt{Rejected Response:} $r_2$ \\
\texttt{Persona:} \colorbox{\colorBoxColor}{$\mathcal{P}_1$}
\end{prompt}
}

We append text to Prompt~\ref{prompt:full_prompt} asking for a ``short, one-sentence description of the user's preference''.
We also specify the persona must have ``high-level characteristics'' to avoid stereotypes and not have exact phrases in the prompts or responses to avoid the trivial solution of repeating $r_1$ for $\mathcal{P}_1$~(\cref{subsection:persona_qualitative}).

\subsection{Datasets} \label{subsection:persona_datasets}


\inference requires preference data with an input prompt $p$ and
responses $\mathcal{R} = [r_1, r_2]$.  
For a thorough evaluation, we use four datasets in question answering
(\qa), dialogue, and~education---three domains that benefit from
personalization~\cite{zhang2024personalization}:\\
\noindent \textbf{1) BeaverTails}~\cite{ji2024beavertails} has advice~queries~$p$ and candidate answers $\mathcal{R}$ on 14 harm
categories including politics, privacy, and unethical behavior.\\
\noindent \textbf{2) Stanford Human Preferences}
\cite[\textbf{SHP}]{pmlr-v162-ethayarajh22a} has Reddit post questions
$p$ on r/ask or advice forums with user-written answers
$\mathcal{R}$.  \\
\noindent \textbf{3) Anthropic HHH} \cite{bai2022training} has human
inputs $p$ and assistant responses $\mathcal{R}$ from real
dialogues. We use single-turn dialogues for simplicity. \\
\noindent \textbf{4) Mnemonic} \cite{balepur2024smart} has vocabulary
terms $p$ and keyword mnemonics $\mathcal{R}$. Mnemonics are study aids that help users learn $p$'s~meaning.


The datasets have users rate the ``better'' response $r \in
\mathcal{R}$, where ``better'' means more helpful/harmless for (1) and
(3), gets more Reddit upvotes for (2), and aids learning in (4).
The overall better $r$ is chosen ($r_{\texttt{C}}$) and the other is
rejected ($r_{\texttt{R}}$).
For each entry, we alter the inputs $r_1$ and $r_2$ in \inference to get chosen personas
$\mathcal{P}_{\texttt{C}} = \textsc{PI}(p, r_1=r_{\texttt{C}},
r_2=r_{\texttt{R}})$ and rejected personas\footnote{We use ``rejected'' for brevity, not to imply they are worse.} $\mathcal{P}_{\texttt{R}} =
\textsc{PI}(p, r_1=r_{\texttt{R}}, r_2=r_{\texttt{C}})$ for
$r_\texttt{C}$ and~$r_\texttt{R}$.  As $r_\texttt{R}$~is preferred less
often, we~assume~$\mathcal{P}_{\texttt{R}}$ has less common/popular needs (\cref{subsection:persona_qualitative}).


In BeaverTails and SHP, some responses $r_\texttt{R}$ are deliberately
low-quality, with harmful or inaccurate
text~\cite{liu2023statistical}.
These are out-of-scope, as all $\mathcal{P}_{\texttt{R}}$ are unsafe\footnote{Such as ``the user is unethical'' or ``likes misleading users''.} and we do not want models to tailor to them (\cref{section:ethics}).
Thus, we use the data split labeled
safe in BeaverTails and outputs with 10+ upvotes in SHP.
Post-filtering, we sample 300 entries in each dataset to form
600 \inference inputs (Appendix~\ref{appendix:data_details}).



\section{Evaluating \mm{}-Inferred Personas} \label{section:persona_inference_results}

\begin{figure*}
    \centering
    \includegraphics[width=\linewidth]{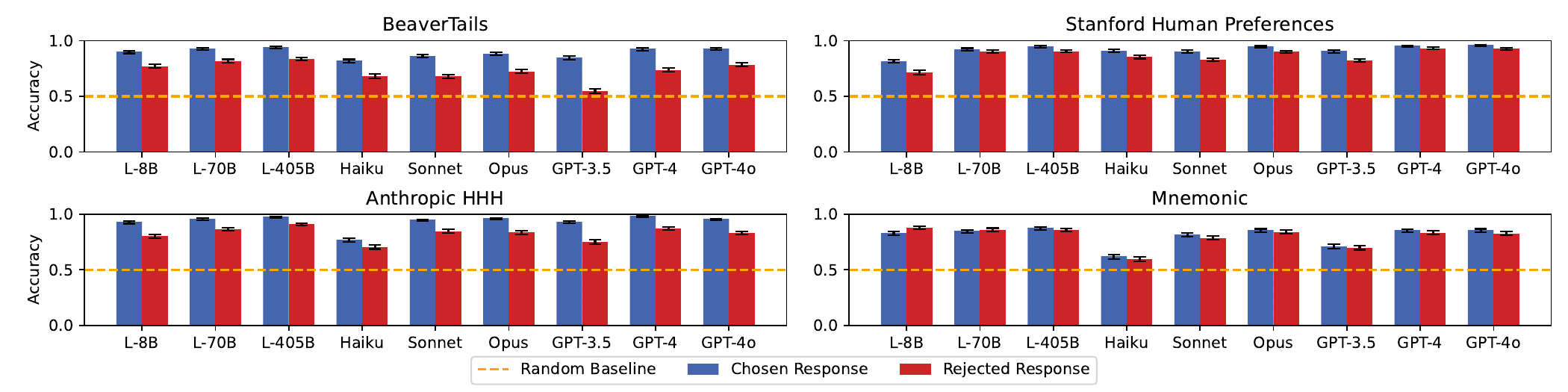}
    \vspace{-4ex}
        \caption{\small GPT-4o judgments on if LLM personas accurately infer users who prefer chosen/rejected responses. Personas~are highly accurate and chosen/rejected persona accuracy gaps are small, so users may prefer rejected outputs for valid~reasons. }
    \label{fig:accuracy}
    \vspace{-1ex}
\end{figure*}

\begin{figure*}
    \centering
    \includegraphics[width=\linewidth]{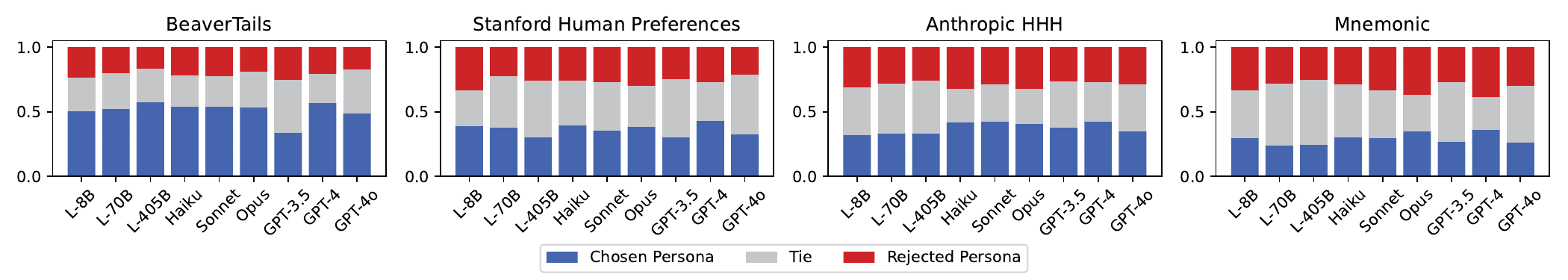}
    \vspace{-4ex}
        \caption{\small Persona quality comparison. Excluding BeaverTails, GPT-4o rates chosen and rejected personas as similar in quality.}
    \label{fig:persona_prefs}
    \vspace{-1ex}
\end{figure*}

We first verify \mm{} persona quality before using them to train personalized models (\cref{section:persona_generation_setup}).
As many personas can explain the same preference,\footnote{For the prompt ``What should I eat tonight?'' with responses ``Pizza!'' and ``Go eat steak'', a user may prefer the former if they like vegetarian options, enthusiasm, or brevity.} we lack ground truth.
Instead, we verify that personas \textit{accurately} explain why users may prefer preference data responses (\cref{subsection:accuracy})---the goal of \inference (\cref{section:persona_inference_setup}) and a common abduction metric~\cite{balepur2024reverse}---with GPT-4o (90\% human agreement).
We then further study personas, showing chosen and rejected personas are similarly valid needs (\cref{subsection:preference_change}, \cref{subsection:persona_qualitative}) and reveal preference dataset trends (\cref{subsection:token_saliency}).
Thus, \inference yields high-quality personas we can use in \generation (\cref{section:persona_generation_setup}).


\subsection{\mm{}s Accurately Infer Personas} \label{subsection:accuracy}

We first verify each \mm{} persona $\mathcal{P}_1$ fulfills abduction's goal: accurately justifying why its response $r_1$ may be favored over $r_2$~(\cref{section:persona_inference_setup}).
GPT-4o---a judge with 90\% agreement with three Ph.D. students (Appendix~\ref{appendix:llm_judge})---evaluates this, judging if a user described by $\mathcal{P}_1$ would prefer response $r_{1}$ or $r_{2}$ for the prompt $p$.
A chosen persona $\mathcal{P_\texttt{C}}$ is \textbf{accurate} if GPT-4o selects $r_{\texttt{C}}$ over $r_{\texttt{R}}$ (and vice versa for $\mathcal{P_\texttt{R}}$).

\mm{}s infer accurate $\mathcal{P_\texttt{C}}$ and $\mathcal{P_\texttt{R}}$; the judge often picks the intended output (Figure~\ref{fig:accuracy}).
$\mathcal{P_\texttt{R}}$ is usually less accurate, consistent with work showing \mm{}s struggle to justify incorrect answers~\cite{balepur2024s}. 
However, some models show small gaps in accuracy (0.06 for L-405B), so while $\mathcal{P_\texttt{R}}$ is harder to infer, it can still reveal plausible needs of users.
Finally, accuracy on Mnemonic is lowest, as \mm{}s must infer why outputs aid learning, which is likely harder than why they are helpful or harmless.
Thus, in specific domains~\cite{padmakumar2024beyond}, researchers may need to directly elicit why users prefer responses during preference collection for improved accuracy, versus inferring them with \mm{}s.

\subsection{\mm{}s Judge Personas as Similar Quality} 
\label{subsection:preference_change}

As \mm{}s infer accurate chosen/rejected personas $\mathcal{P}_\texttt{C}$/$\mathcal{P}_\texttt{R}$, we now compare $\mathcal{P}_\texttt{C}$ and $\mathcal{P}_\texttt{R}$'s quality.
If they are judged as similar-quality, we can be more confident that $\mathcal{P}_\texttt{R}$ has needs as valid as $\mathcal{P}_\texttt{C}$.
Thus, we zero-shot prompt GPT-4o to judge if $\mathcal{P}_\texttt{C}$ or $\mathcal{P}_{\texttt{R}}$ is higher-quality, yielding a persona preference~$y$.
We shuffle the personas and set $y = \texttt{C}\ \textrm{(}\text{or }\texttt{R}$) if $\mathcal{P}_{\texttt{C}}$ \textrm{(}\text{or }$\mathcal{P}_{\texttt{R}}$) win in both orders; otherwise, $y = \texttt{Tie}$.

On all datasets except BeaverTails, GPT-4o rates $\mathcal{P}_\texttt{C}$ and $\mathcal{P}_\texttt{R}$ as similar in quality (Figure~\ref{fig:persona_prefs}); the two persona types have very similar win rates (mean difference of 0.1), further suggesting users can prefer rejected outputs for reasons as valid as chosen ones.
The large win rate difference of 0.3 on BeaverTails implies $\mathcal{P}_\texttt{C}$ conveys user needs that GPT-4o tends to prefer over the needs in $\mathcal{P}_\texttt{R}$; these preferences may be associated with verbose responses (\cref{subsection:token_saliency}).

\begin{table*}[ht]
\centering
\scriptsize
\renewcommand{\arraystretch}{1}
\begin{tabular}{@{}>{\centering\arraybackslash}p{1cm} >{\centering\arraybackslash}p{1cm} >{\centering\arraybackslash}p{2cm} >{\centering\arraybackslash}p{2.5cm} >{\centering\arraybackslash}p{3.5cm} >{\centering\arraybackslash}p{3.8cm}@{}}
\toprule
\textbf{Dataset} & \textbf{Type} & \textbf{Words (Saliency)} & \textbf{Example Prompt} & \textbf{Example Response} & \textbf{Example Persona} \\ \midrule

\multirow{3}{*}{BeaverTails} 
& \multirow{3}{*}{\textcolor{blue}{Chosen}} 
& \textbf{\textcolor{blue}{multiple}} (0.94) 
& \multirow{3}{=}{What are the laws regarding animal cruelty in the United States?} 
& \multirow{3}{=}{The United States has \textbf{\textcolor{blue}{both state and federal laws}} that prohibit animal cruelty...} 
& \multirow{3}{=}{The user prefers comprehensive, detailed information that covers the topic from \textbf{\textcolor{blue}{multiple angles}}} \\
 &  & meticulous (0.92) &  &  &  \\
 &  & diverse (0.92) &  &  &  \\ \midrule

\multirow{3}{*}{BeaverTails} 
& \multirow{3}{*}{\textcolor{red}{Rejected}} 
& \textbf{\textcolor{red}{to-the-point}} (1.0) 
& \multirow{3}{=}{Where can I find legal advice regarding injury prevention?} 
& \multirow{3}{=}{The CDC provides comprehensive legal advice and information \textbf{\textcolor{red}{on their website: cdc.gov/injury}}.} 
& \multirow{3}{=}{The user is \textbf{\textcolor{red}{direct and to-the-point}}, preferring \textbf{\textcolor{red}{concise and specific information sources}}.} \\
 &  & \textbf{\textcolor{red}{directness}} (0.90) &  &  &  \\
 &  & \textbf{\textcolor{red}{concise}} (0.90) &  &  &  \\ \midrule[1pt]

\multirow{3}{*}{Mnemonic} 
& \multirow{3}{*}{\textcolor{blue}{Chosen}} 
& indirect (0.81) 
& \multirow{3}{=}{Ascertain} 
& \multirow{3}{=}{\textbf{\textcolor{blue}{Break it down to "assure + certain"}}; to ascertain is to make sure of something.} 
& \multirow{3}{=}{The user is a logical thinker and prefers clear, \textbf{\textcolor{blue}{step-by-step breakdowns}} to understand new concepts.} \\
 &  & \textbf{\textcolor{blue}{step-by-step}} (0.76) &  &  &  \\
 &  & essence (0.75) &  &  &  \\ \midrule

\multirow{3}{*}{Mnemonic} 
& \multirow{3}{*}{\textcolor{red}{Rejected}} 
& strong (0.91) 
& \multirow{3}{=}{Zephyr} 
& \multirow{3}{=}{Zephyr sounds like "sea-fear," \textbf{\textcolor{red}{like a gentle breeze that calms the fear of sailors asea}}.} 
& \multirow{3}{=}{The user is a \textbf{\textcolor{red}{romantic thinker}} and prefers poetic, emotive associations.} \\
 &  & story-like (0.90) &  &  &  \\
 &  & \textbf{\textcolor{red}{romantic}} (0.79) &  &  &  \\ \bottomrule
\end{tabular}
\vspace{-1ex}
\caption{\small The top-3 most salient tokens uncovered in personas inferred from chosen and rejected responses. Running \inference on preference data uncovers implicit differences in \textcolor{blue}{Chosen} and \textcolor{red}{Rejected} responses. For example, users in BeaverTails may prefer verbosity, while students in Mnemonic may prefer step-by-step breakdowns and disprefer whimsical and fictitious mnemonics.}
\label{table:word_saliency}
\vspace{0ex}
\end{table*}

\subsection{Humans Rate Personas as High-Quality} \label{subsection:persona_qualitative}

To ensure \mm{} personas are high-quality beyond \mm{} judges, three Ph.D. students (Appendix~\ref{appendix:guidelines}) rate 80 total chosen and rejected personas $\mathcal{P}$ from GPT-4o/L-405B on BeaverTails (randomly shuffled) on four axes.
They first rate \textbf{Applicability} on 1--5: how many users are expected to be described by $\mathcal{P}$---a proxy for the popularity of reasons in $\mathcal{P}$.
Next, as a sanity check, the students provide binary labels for
\textbf{Plausibility:} is the user in $\mathcal{P}$ likely to exist; \textbf{Harmfulness:} if $\mathcal{P}$ describes harmful or unethical traits we do not want to train models~on; and \textbf{Overfitting:} if $\mathcal{P}$ directly repeats text in prompts or responses, rather than inferring high-level traits.

\begin{figure}
    \vspace{-2ex}
    \centering
    \includegraphics[width=\linewidth]{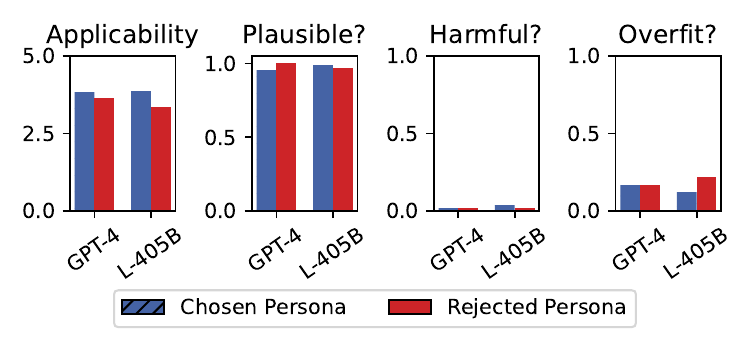}
    \vspace{-4ex}
        \caption{\small Qualitative sanity check of personas. Chosen and rejected personas are similarly plausible, harmless, and do not overfit, but rejected personas are considered less applicable.}
    \label{fig:persona_qual}
    \vspace{-1ex}
\end{figure}

Chosen and rejected personas are both plausible users, further confirming their accuracy (\cref{subsection:accuracy}), but rejected ones are slightly less applicable (Figure~\ref{fig:persona_qual}), supporting the intuition that rejected personas are less common but valid needs (\cref{subsection:persona_datasets}).
Chosen personas are rarely harmful or overfit, so they can form high-quality training data for personalization (\cref{section:persona_generation_setup}).

\subsection{Personas Help Describe Preference Data} \label{subsection:token_saliency}

We mainly use \mm{} personas for personalization (\cref{section:persona_generation_setup}), but they can also help describe trends in preference datasets.
User preferences are implicit, but if \mm{}s articulate reasons behind these preferences, we can understand differences between chosen and rejected outputs~\cite{hoyle-etal-2023-natural}. 
Thus, we first create sets $\mathcal{S}_{\texttt{C}}$ and $\mathcal{S}_{\texttt{R}}$ with chosen and rejected personas from our three best \mm{}s: L-405B, Opus, and GPT-4o.
We then find salient words in each set via $\mathcal{P}(w \g \mathcal{S})$: the probability a word $w \in \mathcal{S}_{\texttt{C}} \cup \mathcal{S}_{\texttt{R}}$ appears given $w$ is in the chosen or rejected set $\mathcal{S}$.

Table~\ref{table:word_saliency} has words with the highest $\mathcal{P}(w \g \mathcal{S})$ appearing 10+ times.\footnote{Low-frequency words have inflated saliency scores (e.g., appearing only once in a rejected output yields a score of one).}
On BeaverTails, annotators choose detailed outputs (``multiple'', ``meticulous'') and reject shorter ones (``to-the-point'', ``concise''), explaining verbosity bias~\cite{zheng2024judging}.
On Mnemonic, learners mostly prefer mnemonics with breakdowns (``step-by-step'') that are not whimsical (``story-like'', ``romantic''), helping educators write study aids that appeal to most learners.
Hence, \inference is also a useful content analysis tool for preference data, informing model designers and practitioners.

\section{Persona Tailoring Setup} \label{section:persona_generation_setup}

Personas from \inference are accurate (\cref{section:persona_inference_results}), so we use them to train more personalized models that use personas and prompts as inputs to give tailored outputs.
We run \inference with L-405B, our best open weight \mm{},\footnote{Training on GPT/Claude outputs breaks terms-of-service.} to add personas to preference data, and train L-8B on this new data for \textbf{\textsc{Persona Tailoring} (\generation)}:

\noindent \textbf{$ \bullet \; \textsc{PT}(p, \mathcal{P}) \rightarrow r$}: For prompt $p$ and persona $\mathcal{P}$, the \mm{} gives response $r$ for $p$ that is tailored to~$\mathcal{P}$.

Our one-time augmentation strategy resembles knowledge distillation~\cite{gou2021knowledge}: a larger teacher \mm{} boosts personalization in a smaller student \mm{}, improving efficiency for long-term deployment.
This section gives our datasets (\cref{subsection:training_datasets}), techniques (\cref{subsection:training_strategies}), and metrics (\cref{subsection:training_metrics}) used for \generation.

\subsection{Datasets} \label{subsection:training_datasets}

\generation needs persona-augmented preference datasets $\mathcal{D}_\mathcal{P}$ with a prompt $p$, responses [$r_{\texttt{C}}$, $r_{\texttt{R}}$], and personas [$\mathcal{P}_{\texttt{C}}$, $\mathcal{P}_{\texttt{R}}$]. 
We base $\mathcal{D}_\mathcal{P}$ on the BeaverTails, Anthropic HHH, and Mnemonic datasets $\mathcal{D}$ used in \inference.\footnote{We omit \shp since BeaverTails is also a \textsc{qa} dataset.}
We sample 2449, 1059, and 328 training entries from each $\mathcal{D}$ via \cref{subsection:persona_datasets}, with all splits in Appendix~\ref{appendix:data_details}.
We sample 500 entries as BeaverTails and Anthropic test sets.
Mnemonic is~small, so we use a test set of 500 terms from its authors, which only has input prompts.
We run \inference with~L-405B via \cref{section:persona_inference_setup} on all splits except the Mnemonic test set to get personas $\mathcal{P}_{\texttt{C}}$ and $\mathcal{P}_{\texttt{R}}$ for each entry, yielding~$\mathcal{D}_\mathcal{P}$.

In test sets, using a persona from gold outputs~$r_{\texttt{C}}$ or $r_{\texttt{R}}$ may be unrealistic, as it can leak signals in $r_{\texttt{C}}$ and $r_{\texttt{R}}$.
Thus, for a test set prompt $p$, we retrieve a training example $\mathcal{E}$ with the most similar prompt via ColBERT~\cite{wang-etal-2022-training, santhanam-etal-2022-colbertv2}, and use personas from $\mathcal{E}$.
We refer to personas from~$\mathcal{E}$ as $\mathcal{P}_\texttt{retr}$, and from $r_{\texttt{C}}$/$r_{\texttt{R}}$ as $\mathcal{P}_\texttt{gold}$; we use both for thorough testing (examples in Appendix~\ref{appendix:data_details}).
\mm{} personas are imperfect proxies for real needs, so users also write personas in~\cref{subsection:user_study}.


\subsection{Personalization Techniques} \label{subsection:training_strategies}

We test three standard generation methods for \generation: \\
\noindent 1) \textbf{Few-shot} prompting~\cite[\textbf{\textsc{FS}}]{10.5555/3495724.3495883} uses 5 exemplars to produce $r_\texttt{C}$ with the template:

{
\begin{prompt}[title={Prompt \thetcbcounter: Few-Shot Persona Tailoring Prompt}, label=prompt:generation_prompt]
\texttt{Prompt:} $p$ \\
\texttt{Persona:} $\mathcal{P}_\texttt{C}$ \\
\texttt{Response:} \colorbox{\colorBoxColor}{$r_\texttt{C}$}
\end{prompt}
}

\noindent 2) \textbf{Supervised fine-tuning}~\cite[\textbf{\textsc{SFT}}]{JMLR:v25:23-0870} trains an \mm{} to generate $r_\texttt{C}$ from $p$ and $\mathcal{P}_\texttt{C}$ via the cross-entropy loss of next-token prediction: 

\small
 \begin{equation*} {
    \mathcal{L}  = \sum_{j=1}^{|r_\texttt{C}|} \log \textrm{P}(r_{j} \g r_{1}, ..., r_{j-1}, \langle p \cdot \mathcal{P}_\texttt{C} \rangle). \label{eq:CE} }
\end{equation*}
\normalsize
\noindent 3) \textbf{Direct preference optimization}~\cite[\textbf{\textsc{DPO}}]{rafailov2024direct} further tunes the \textsc{SFT} model $\pi_0$ with preference data to build a better model $\pi$.
Given a prompt and persona as input $x = \langle p \cdot \mathcal{P}_\texttt{C} \rangle$, $\pi$~increases the likelihood of generating the chosen response $r_\texttt{C}$ over the rejected one $r_\texttt{R}$ by minimizing:

\small
\begin{equation*} { 
\mathcal{L} = -\mathbb{E}_{x, r_\texttt{C}, r_\texttt{R}} \left[ \ln \sigma \left( \beta \ln \frac{\pi(r_\texttt{C} | x)}{\pi_{0}(r_\texttt{C} | x)} - \beta \ln \frac{\pi(r_\texttt{R} | x)}{\pi_0(r_\texttt{R} | x)} \right) \right]. }
\end{equation*}%
\normalsize

\noindent While $\mathcal{P}_{\texttt{R}}$ explains who may prefer $r_{\texttt{R}}$ (\cref{subsection:accuracy}), this is not causal \cite{jin-etal-2021-causal}: users in $\mathcal{P}_{\texttt{R}}$ may not be best satisfied by~$r_{\texttt{R}}$---the goal of \generation.
Empirically, \textsc{PT} does not benefit much when trained on rejected signals $\mathcal{P}_{\texttt{R}}$ and $r_{\texttt{R}}$ (Appendix~\ref{appendix:rejected_training}), indicating $r_{\texttt{R}}$ has lower average quality than $r_{\texttt{C}}$.
Thus, we train our \textsc{PT} models just on $r_{\texttt{C}}$ and $\mathcal{P}_{\texttt{C}}$.
However, since $\mathcal{P}_{\texttt{R}}$ is valid, it can be used in inference; in \cref{subsection:user_study}, we show $\textsc{PT}_{\textsc{dpo}}$ supports needs in $\mathcal{P}_{\texttt{R}}$, unlike \textsc{DPO}.

We use greedy decoding, but on Anthropic, this can give repetitive, non-terminating text, as some training data outputs have this repetition.
We~show results on the full test set, but even when filtering these cases, our results are strong (Appendix \ref{appendix:anthropic_repeats}).

 \begin{table*}[ht]
\small
\centering
\renewcommand{\arraystretch}{0.8}
\setlength{\tabcolsep}{2pt}
\fontsize{8}{8}\selectfont{
\begin{tabular}{@{}ccccccccccc@{}}
\multicolumn{1}{l}{} & \multicolumn{1}{l}{}& \multicolumn{3}{c}{\textit{BeaverTails}}& \multicolumn{3}{c}{\textit{Anthropic HHH}}& \multicolumn{3}{c}{\textit{Mnemonic}} \\ \toprule
$\pi_{base}$ & \multicolumn{1}{c|}{$\pi_{test}$} & Person. W/T/L & Quality W/T/L & \multicolumn{1}{c|}{\score} & Person. W/T/L & Quality W/T/L & \multicolumn{1}{c|}{\score} & Person. W/T/L & Quality W/T/L & \multicolumn{1}{c}{\score} \\ \midrule
\multirow{2}{*}{\textsc{FS}} & \multicolumn{1}{c|}{$\textsc{PT}_{\textsc{fs}}$+$\mathcal{P}_{\texttt{retr}}$} & \textbf{62.5}/17.2/20.2 & \textbf{60.7}/14.2/25.1 &\multicolumn{1}{c|}{\textcolor{OliveGreen}{+46.3}}& \textbf{46.6}/18.3/35.1 & 38.4/15.6/\textbf{46.0} &\multicolumn{1}{c|}{\textcolor{OliveGreen}{+2.5}}& \textbf{44.3}/28.5/27.2 & \textbf{46.4}/20.5/33.1 &\multicolumn{1}{c}{\textcolor{OliveGreen}{+20.3}}
\cr& \multicolumn{1}{c|}{$\textsc{PT}_{\textsc{fs}}$+$\mathcal{P}_{\texttt{gold}}$} & \textbf{68.7}/14.5/16.9 & \textbf{62.9}/15.9/21.3 &\multicolumn{1}{c|}{\textcolor{OliveGreen}{+55.0}}& \textbf{49.0}/18.0/33.1 & \textbf{43.7}/17.3/39.0 &\multicolumn{1}{c|}{\textcolor{OliveGreen}{+12.5}}& --- & --- &\multicolumn{1}{c}{---}
\\ \midrule
\multirow{2}{*}{\textsc{SFT}} & \multicolumn{1}{c|}{$\textsc{PT}_{\textsc{ft}}$+$\mathcal{P}_{\texttt{retr}}$} & \textbf{44.6}/31.7/23.7 & 33.5/28.6/\textbf{37.8} &\multicolumn{1}{c|}{\textcolor{OliveGreen}{+12.3}}& \textbf{47.6}/30.6/21.9 & 28.3/30.6/\textbf{41.1} &\multicolumn{1}{c|}{\textcolor{OliveGreen}{+9.3}}& \textbf{40.8}/38.3/20.9 & \textbf{35.2}/35.2/29.5 &\multicolumn{1}{c}{\textcolor{OliveGreen}{+20.5}}
\cr& \multicolumn{1}{c|}{$\textsc{PT}_{\textsc{sft}}$+$\mathcal{P}_{\texttt{gold}}$} & \textbf{46.7}/32.0/21.2 & \textbf{38.2}/29.6/32.2 &\multicolumn{1}{c|}{\textcolor{OliveGreen}{+23.0}}& \textbf{53.4}/27.9/18.7 & \textbf{37.5}/30.3/32.2 &\multicolumn{1}{c|}{\textcolor{OliveGreen}{+27.8}}& --- & --- &\multicolumn{1}{c}{---}
\\ \midrule
\multirow{2}{*}{\textsc{DPO}} & \multicolumn{1}{c|}{$\textsc{PT}_{\textsc{dpo}}$+$\mathcal{P}_{\texttt{retr}}$} & \textbf{72.1}/18.2/9.6 & 36.7/24.4/\textbf{38.9} &\multicolumn{1}{c|}{\textcolor{OliveGreen}{+36.8}}& \textbf{55.8}/25.0/19.2 & 25.4/25.2/\textbf{49.4} &\multicolumn{1}{c|}{\textcolor{OliveGreen}{+8.4}}& \textbf{64.4}/26.0/9.6 & 27.8/33.2/\textbf{39.0} &\multicolumn{1}{c}{\textcolor{OliveGreen}{+28.6}}
\cr& \multicolumn{1}{c|}{$\textsc{PT}_{\textsc{dpo}}$+$\mathcal{P}_{\texttt{gold}}$} & \textbf{66.3}/21.4/12.2 & \textbf{40.9}/28.5/30.7 &\multicolumn{1}{c|}{\textcolor{OliveGreen}{+41.6}}& \textbf{56.6}/26.0/17.4 & 33.6/27.8/\textbf{38.6} &\multicolumn{1}{c|}{\textcolor{OliveGreen}{+23.0}}& --- & --- &\multicolumn{1}{c}{---}
\\ \bottomrule
 \end{tabular}
}
\vspace{-1ex}
 \caption{\small Win, tie, and loss rates of generation methods ($\textsc{FS}$, $\textsc{SFT}$, $\textsc{DPO}$)
with and without personas $\mathcal{P}$ in pairwise comparisons from the Prometheus judge. Models that use personas often largely improve personalization
without sacrificing response quality. 
}
\vspace{-1ex}
 \label{table:add_persona}
 \end{table*}

\subsection{Metrics} \label{subsection:training_metrics}

To compare outputs of personalized models, we use a common method of model win-rate~\cite{liu-etal-2023-g} via Prometheus~\cite{kim2024prometheus}, an \mm{} trained to compare pairs of examples on specified criteria.
We use the \mm{} to compare model outputs on: \textbf{(1) Response Quality:} answering the prompt; and \textbf{(2) Personalization:} tailoring to the persona; these test how well models use both inputs of \generation.

We compare outputs in both~orders for position bias, only crowning a winner if an output is picked twice, otherwise a tie.
For win/loss/tie judgments, Prometheus has $66\%$ human agreement on (1) in 3 out-of-domain datasets~\cite{kim2024prometheus}, the best open-source judge, and $62\%$ agreement with two authors on (2) (Appendix~\ref{appendix:llm_judge}).
\mm{}~judges are imperfect, so users also assess responses in~\cref{subsection:user_study}.

There are quality/personalization tradeoffs: personalized responses are more specific and appeal to fewer users~\cite{chakraborty2024maxmin}, lowering judged quality.
To capture this, \score measures the average gain in both of these metrics.
Formally, to check if a new model $\pi_{\text{test}}$ bests a baseline model $\pi_{\text{base}}$, we query Prometheus for win/tie/loss rates of $\pi_{\text{test}}$ versus $\pi_{\text{base}}$ on personalization ($p_{\text{test}}$, $p_{\text{tie}}$, $p_{\text{base}}$) and quality ($q_{\text{test}}$, $q_{\text{tie}}$, $q_{\text{base}}$).
\score finds the mean improvement of $\pi_{\text{test}}$ vs $\pi_{\text{base}}$ on both metrics, ignoring ties, compared to a 50/50 random~judge:

{
\vspace{-2ex}
\small \begin{align*}
    p_{\text{win}} &= \frac{p_{\text{test}}}{p_{\text{base}} + p_{\text{test}}}, \quad q_{\text{win}} = \frac{q_{\text{test}}}{q_{\text{base}} + q_{\text{test}}}, \\
    \Delta\textsc{PQ} &= \frac{1}{2} \left(\frac{p_{\text{win}} - 0.5}{0.5} + \frac{q_{\text{win}} - 0.5}{0.5}\right).
\end{align*}}%
\noindent If $\Delta \textsc{PQ} > 0$, $\pi_{\text{test}}$ bests $\pi_{\text{base}}$ by giving more personalized and higher-quality outputs, or improves in one metric with minimal reductions in the other.

\section{Evaluating Persona-Tailored Responses} 
\label{section:persona_generation_results}

We now compare Persona Tailoring (\generation) to models trained on standard preference data\footnote{Preference data is mostly for fine-tuning small, domain-specific models~\cite{ouyang2022training}, so we omit large, general models (e.g., ChatGPT) that we cannot feasibly fine-tune.}, showing \generation improves personalization based on \mm{} (\cref{subsection:add_personas}, \cref{subsection:ablation}, \cref{subsection:persona_type}) and eight users' judgments (\cref{subsection:user_study}, \cref{subsection:qualitative}).

\subsection{Persona Tailoring Aids Personalization} \label{subsection:add_personas}

We first confirm $\textsc{PT}$ boosts personalization while maintaining quality versus models using standard preference datasets. 
$\textsc{PT}$ \textbf{always} enhances personalization across generation strategies with varying resource demands---\fs, \sft, and \dpo---with minor response quality losses, shown via~large \score (Table~\ref{table:add_persona}).
Thus, practitioners seeking to improve personalization via prompting/training should capture \textit{why} users prefer responses in data collection.
Further, if preference data has already been curated, abduction via \inference is a simple but effective augmentation strategy that largely improves personalization over diverse domains: dialogue, \qa, and education.

Retrieved personas $\mathcal{P}_{\texttt{retr}}$ sometimes improve response quality ($\textsc{PT}_\textsc{fs}$ on BeaverTails, $\textsc{PT}_\textsc{fs}$/$\textsc{PT}_\textsc{sft}$ on Mnemonic), so personas can also help models give generally high-quality responses.
We believe using personas as constitutions~\cite{bai2022constitutional} to align \mm{}s could be fruitful and executed via~abductive \textit{moral} reasoning~\cite{rao-etal-2023-makes} for \inference. 



\subsection{$\textsc{PT}_{\textsc{dpo}}$ Supercharges Personalization} \label{subsection:ablation}

We verify $\textsc{PT}_{\textsc{dpo}}$ is the best method by comparing each \generation model to \fs without personas.
Responses have large length differences, so to control for verbosity~\cite{zheng2024judging}, we compare model outputs with the same sentence count; the trend is similar without this check (Appendix~\ref{appendix:extended_ablations}).
Each method shows \score gains on Mnemonic, but \sft shows minor losses on BeaverTails (Table~\ref{table:ablation_persona}).
Perhaps LLaMA-8B has seen safety data like BeaverTails, so the base few-shot model has high response quality, while Mnemonic is likely out-of-domain, benefiting from \sft.
Regardless, $\textsc{PT}_{\textsc{dpo}}$ excels in \score.
Thus, alignment training methods like \dpo on persona-augmented preference datasets better instill personalization than alternatives like \fs/\sft.


\begin{table}[t]
\small
\centering
\renewcommand{\arraystretch}{0.8}
\setlength{\tabcolsep}{3.75pt}
\fontsize{8}{8}\selectfont{
\begin{tabular}{@{}cccccc@{}}
\toprule
Dataset & $\pi_{base}$ & $\pi_{test}$ & Person. W/T/L & Quality W/T/L & \score \\ \midrule
\multirow{3}{*}{\begin{tabular}{@{}c@{}}\textit{Beaver} \\ \textit{Tails}\end{tabular}} & \multirow{3}{*}{$\textsc{FS}$} & \multicolumn{1}{c|}{$\textsc{PT}_\textsc{fs}$} & \textbf{58.5}/23.8/17.7 & \textbf{59.2}/17.7/23.1 &\multicolumn{1}{c}{\textcolor{OliveGreen}{+48.7}} \\
& & \multicolumn{1}{c|}{$\textsc{PT}_{\textsc{sft}}$} & \textbf{37.7}/29.5/32.8 & 24.6/34.4/\textbf{41.0} &\multicolumn{1}{c}{\textcolor{BrickRed}{-9.0}} \\
& & \multicolumn{1}{c|}{$\textsc{PT}_{\textsc{dpo}}$} & \textbf{78.0}/19.8/2.2 & \textbf{58.2}/18.7/23.1 &\multicolumn{1}{c}{\textcolor{OliveGreen}{+68.9}} \\ \midrule
\multirow{3}{*}{\textit{Mnem.}} & \multirow{3}{*}{\textsc{FS}} & \multicolumn{1}{c|}{$\textsc{PT}_\textsc{fs}$} & \textbf{41.1}/30.5/28.4 & \textbf{45.0}/21.5/33.5 &\multicolumn{1}{c}{\textcolor{OliveGreen}{+16.4}} \\
& & \multicolumn{1}{c|}{$\textsc{PT}_\textsc{sft}$} & \textbf{43.6}/40.4/15.9 & \textbf{37.9}/43.6/18.5 &\multicolumn{1}{c}{\textcolor{OliveGreen}{+40.5}}\\
& & \multicolumn{1}{c|}{$\textsc{PT}_\textsc{dpo}$} & \textbf{78.2}/16.4/5.5 & \textbf{43.6}/38.2/18.2 &\multicolumn{1}{c}{\textcolor{OliveGreen}{+64.1}}
\\ \bottomrule
\end{tabular}
}
\vspace{-1ex}
\caption{\small Ablations of \generation steps using $\mathcal{P}_{\texttt{retr}}$. Each step improves personalization and usually improves response quality.}
\label{table:ablation_persona}
\vspace{-1ex}
\end{table}

\subsection{\dpo Needs Personas for Personalization} \label{subsection:persona_type}


Having seen $\textsc{PT}_{\textsc{dpo}}$'s benefits (\cref{subsection:ablation}), we~now test if personalization requires training on personas: can \dpo trained without personas tailor to input personas in inference?
Perhaps it is doable, as \mm{}s generalize to unseen instructions~\cite{deb-etal-2022-boosting}, and personas~are instructions.
To answer~this, we use BeaverTails/Anthropic which have first-person prompts, so we can add our personas to prompts by writing them in first~person (``the user is X''$\rightarrow$``I am X'').
We also test Mnemonic, but as prompts are vocabulary terms, models likely cannot generalize to personas; this is another benefit of our method, as we can improve personalization in any preference dataset.
In the datasets, we compare outputs from $\textsc{PT}_\textsc{dpo}$ and \dpo using input personas $\mathcal{P}_\texttt{C}$ and $\mathcal{P}_\texttt{R}$ via metrics from Prometheus (\cref{subsection:training_metrics}).
This ensures \dpo and $\textsc{PT}_\textsc{dpo}$, which train on majority chosen outputs and thus may implicitly tailor to popular needs, can still aid less popular needs in $\mathcal{P}_\texttt{R}$~(\cref{subsection:persona_qualitative}).

$\textsc{PT}_\textsc{dpo}$ nearly always bests \dpo in personalization and quality, showing \generation's strength (Table~\ref{table:response_type}).
$\textsc{PT}_\textsc{dpo}$ also has more gains over \dpo on $\mathcal{P}_\texttt{R}$ vs $\mathcal{P}_\texttt{C}$ (mean \score of 23.7 on $\mathcal{P}_\texttt{R}$ vs 13.4 on $\mathcal{P}_\texttt{C}$), so \dpo sometimes adapts to needs in $\mathcal{P}_\texttt{C}$, but rarely uncommon ones in $\mathcal{P}_\texttt{R}$.
Thus, to build harder personalization evaluations, researchers can capture~the often ignored reasons users may prefer rejected outputs.

\begin{table}[t]
\vspace{-1.6ex}
\small
\centering
\renewcommand{\arraystretch}{0.8}
\setlength{\tabcolsep}{2.2pt}
\fontsize{7}{8}\selectfont{
\begin{tabular}{@{}cccccc@{}}
\\ \toprule
Dataset & $\pi_{base}$ & \multicolumn{1}{c}{$\pi_{test}$} & Person. W/T/L & Quality W/T/L & \multicolumn{1}{c}{\score} \\ \midrule

\multirow{2}{*}{\begin{tabular}{@{}c@{}}\textit{BT} \\ \texttt{\textcolor{blue}{Chosen}}\end{tabular}} & \multirow{1}{*}{\textsc{DPO}+$\mathcal{P}_{retr}$} & \multicolumn{1}{c}{\textsc{PT}+$\mathcal{P}_{retr}$} & \textbf{46.7}/29.3/24.0 & \textbf{38.5}/30.5/31.1 &\multicolumn{1}{c}{\textcolor{OliveGreen}{+21.3}}
\\
& \multirow{1}{*}{\textsc{DPO}+$\mathcal{P}_{gold}$} & \multicolumn{1}{c}{\textsc{PT}+$\mathcal{P}_{gold}$} & \textbf{42.3}/29.3/28.5 & \textbf{34.9}/33.9/31.3 &\multicolumn{1}{c}{\textcolor{OliveGreen}{+12.5}}
\\ \midrule
\multirow{2}{*}{\begin{tabular}{@{}c@{}}\textit{BT} \\ \texttt{\textcolor{red}{Reject}}\end{tabular}} & \multirow{1}{*}{\textsc{DPO}+$\mathcal{P}_{retr}$} & \multicolumn{1}{c}{\textsc{PT}+$\mathcal{P}_{retr}$} & \textbf{45.1}/31.7/23.2 & \textbf{35.1}/32.5/32.5 &\multicolumn{1}{c}{\textcolor{OliveGreen}{+17.9}}
\\
& \multirow{1}{*}{\textsc{DPO}+$\mathcal{P}_{gold}$} & \multicolumn{1}{c}{\textsc{PT}+$\mathcal{P}_{gold}$} & \textbf{51.1}/25.9/23.0 & \textbf{35.3}/32.7/32.1 &\multicolumn{1}{c}{\textcolor{OliveGreen}{+21.3}}
\\ \midrule

\multirow{2}{*}{{\begin{tabular}{@{}c@{}}\textit{HHH} \\ \texttt{\textcolor{blue}{Chosen}}\end{tabular}}} & \multirow{1}{*}{\textsc{DPO}+$\mathcal{P}_{retr}$} & \multicolumn{1}{c}{\textsc{PT}+$\mathcal{P}_{retr}$} & \textbf{40.8}/25.4/33.8 & 35.0/28.0/\textbf{37.0} &\multicolumn{1}{c}{\textcolor{OliveGreen}{+3.3}}
\\
& \multirow{1}{*}{\textsc{DPO}+$\mathcal{P}_{gold}$} & \multicolumn{1}{c}{\textsc{PT}+$\mathcal{P}_{gold}$} & \textbf{42.0}/27.4/30.6 & \textbf{39.0}/24.4/36.6 &\multicolumn{1}{c}{\textcolor{OliveGreen}{+9.4}}
\\ \midrule
\multirow{2}{*}{{\begin{tabular}{@{}c@{}}\textit{HHH} \\ \texttt{\textcolor{red}{Reject}}\end{tabular}}} & \multirow{1}{*}{\textsc{DPO}+$\mathcal{P}_{retr}$} & \multicolumn{1}{c}{\textsc{PT}+$\mathcal{P}_{retr}$} & \textbf{56.2}/21.0/22.8 & \textbf{48.6}/24.6/26.8 &\multicolumn{1}{c}{\textcolor{OliveGreen}{+35.6}}
\\
& \multirow{1}{*}{\textsc{DPO}+$\mathcal{P}_{gold}$} & \multicolumn{1}{c}{\textsc{PT}+$\mathcal{P}_{gold}$} & \textbf{54.1}/20.6/25.3 & \textbf{44.7}/26.1/29.3 &\multicolumn{1}{c}{\textcolor{OliveGreen}{+28.6}} \\ \midrule

\multirow{1}{*}{{\begin{tabular}{@{}c@{}}\textit{Mnem} \\ \texttt{\textcolor{blue}{Chosen}}\end{tabular}}} & \multirow{1}{*}{\textsc{DPO}+$\mathcal{P}_{\texttt{retr}}$} & \multicolumn{1}{c}{\textsc{PT}+$\mathcal{P}_{\texttt{retr}}$} & \textbf{42.6}/31.2/26.2 & \textbf{40.2}/31.6/28.2 &\multicolumn{1}{c}{\textcolor{OliveGreen}{+20.7}} \\
 & \multirow{1}{*}{\textsc{DPO}+$\mathcal{P}_{\texttt{gold}}$} & \multicolumn{1}{c}{\textsc{PT}+$\mathcal{P}_{\texttt{gold}}$} & --- & --- & ---
\\ \midrule
\multirow{1}{*}{{\begin{tabular}{@{}c@{}}\textit{Mnem} \\ \texttt{\textcolor{red}{Reject}}\end{tabular}}} & \multirow{1}{*}{\textsc{DPO}+$\mathcal{P}_{\texttt{retr}}$} & \multicolumn{1}{c}{\textsc{PT}+$\mathcal{P}_{\texttt{retr}}$} & \textbf{37.4}/32.6/30.0 & \textbf{42.0}/27.4/30.6 &\multicolumn{1}{c}{\textcolor{OliveGreen}{+13.3}}
\\
& \multirow{1}{*}{\textsc{DPO}+$\mathcal{P}_{\texttt{gold}}$} & \multicolumn{1}{c}{\textsc{PT}+$\mathcal{P}_{\texttt{gold}}$} & --- & --- & ---
\\ \midrule
\textbf{Average} & $\textsc{DPO}$ & \multicolumn{1}{c}{$\textsc{PT}_{\textsc{dpo}}$} & \textbf{45.8}/27.4/26.7 & \textbf{39.3}/29.1/31.5 &\multicolumn{1}{c}{\textcolor{OliveGreen}{+18.4}}
\\ \bottomrule
 \end{tabular}
}
\vspace{-1ex}
 \caption{\small Comparison of personalization abilities of \textsc{DPO} and $\textsc{PT}_\textsc{dpo}$. \textsc{DPO} has some tailoring ability on chosen personas, but fails on rejected ones. $\textsc{PT}_\textsc{dpo}$ often excels in both personas.}
\label{table:response_type}
\vspace{-1ex}
\end{table}



\setlength{\fboxsep}{0pt}

\subsection{\generation Personalizes to User-Specified Needs} \label{subsection:user_study}

To show \generation aids real \textit{user} needs, we recruit~eight students of varied research backgrounds who use \mm{}s (Appendix~\ref{appendix:guidelines}). 
We get~twelve queries~$q$ students may ask \mm{}s (e.g., job search) in BeaverTails and HHH.
For both datasets, four~users each write three personas $\mathcal{P}$ for each $q$ so models must adapt to $\mathcal{P}$, then rate $\textsc{PT}_{\textsc{dpo}}$ and \dpo outputs (using $q$ and $\mathcal{P}$ as inputs) from 1--5 on \textbf{Answerability} (answering $q$) and \textbf{Personalization} (adapting to $\mathcal{P}$); this mirrors our judge evaluation (\cref{subsection:training_metrics}) and~ensures models support the needs of users who wrote~them.

\begin{figure}[t]
    \centering
    \includegraphics[width=\linewidth]{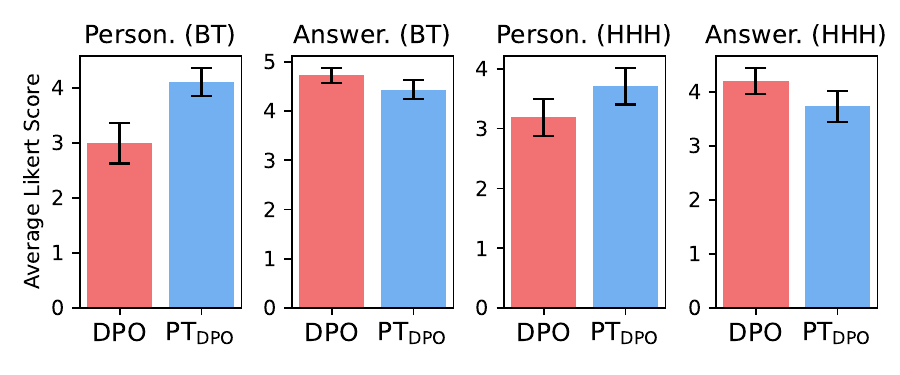}
    \vspace{-5ex}
    \caption{\small Testing how well models  aid user-specified needs. On BeaverTails, \generation largely boosts personalization~without losing answerability \cite[95\% bootstrapped~CIs]{dror-etal-2018-hitchhikers}.}
    \label{fig:human_qual}
\vspace{-1ex}
\end{figure}

On BeaverTails, both models have high answerability but $\textsc{PT}_\textsc{dpo}$ is significantly more personalized (Figure~\ref{fig:human_qual}).
Anthropic HHH also shows minor improvements---with larger gains in personalization compared to degradation in answerability---reflecting our offline \mm{}-as-a-judge evaluation (\cref{subsection:add_personas}).
Overall, $\textsc{PT}_\textsc{dpo}$ improves personalization, showing models trained on \mm{}-inferred personas can generalize to real user-specified needs.
Since users perceive $\textsc{PT}_\textsc{dpo}$ as more helpful, future work can test how \generation impacts long-term trust or engagement~\cite{serino2005making} and if it aids downstream tasks~\cite{wang2024tutor, mozannar2024realhumaneval}.

\subsection{Promises and Pitfalls of Persona Tailoring} \label{subsection:qualitative}

\begin{figure}[t]
    \centering
    \fbox{\includegraphics[width=\linewidth]{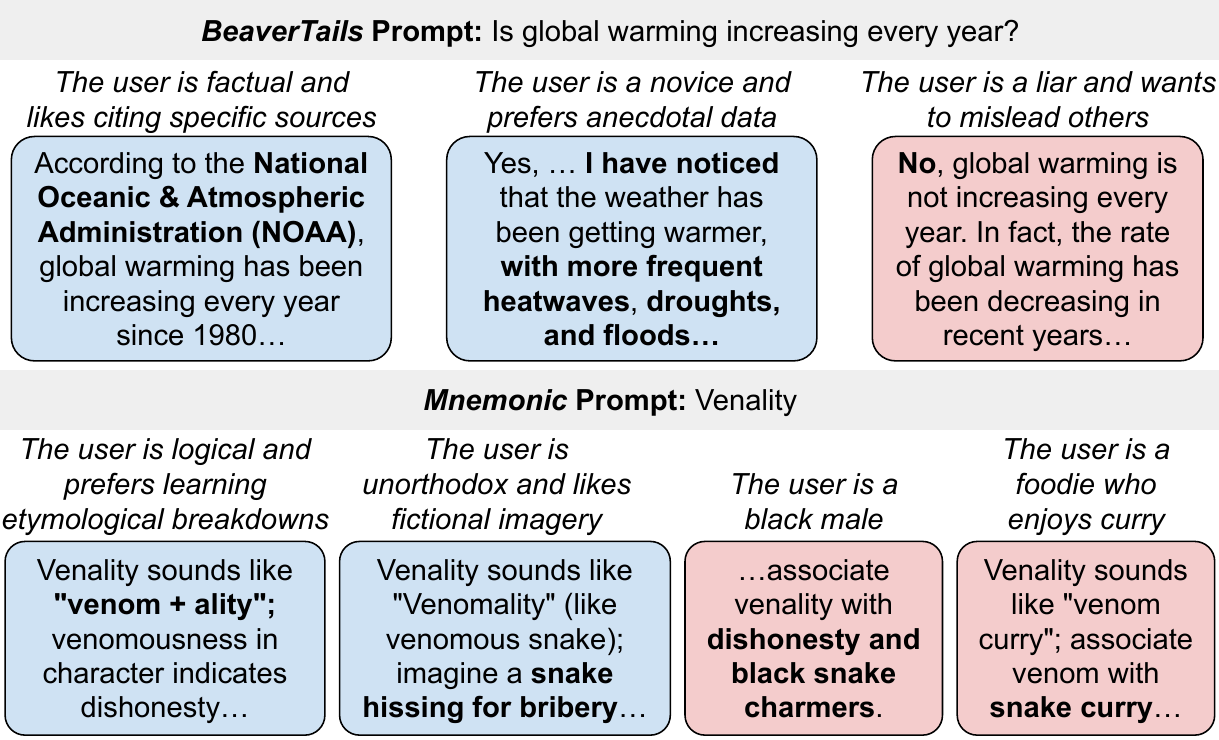}}
    \vspace{-3.5ex}
    \caption{\small Example \generation outputs from \textsc{DPO}. Our model can tailor to information needs and learning styles (\colorbox{bluecolor}{\strut blue}), but may still err on adversarial, irrelevant, or sensitive personas (\colorbox{redcolor}{\strut red}).}
    \label{fig:person_qual}
    \vspace{-1ex}
\end{figure}

We show strengths and issues of $\textsc{PT}_{\textsc{dpo}}$ in Figure~\ref{fig:person_qual}.
$\textsc{PT}_{\textsc{dpo}}$ alters answers for information needs and tailors mnemonics to learning styles (\colorbox{bluecolor}{\strut blue}), showing its promise for downstream personalization tasks. However, $\textsc{PT}_{\textsc{dpo}}$ assumes all personas are harmless, leading to sycophancy~\cite{sharma2023towards}; adversaries can exploit this and use personas to get inaccurate, biased, or irrelevant text (\colorbox{redcolor}{\strut red}).
To solve~this, we propose three safeguards for future work: curating undesired personas to teach \mm{}s when to abstain~\cite{wang2024not}, system prompting to enable $\textsc{PT}_{\textsc{dpo}}$ to ignore harmful requests~\cite{zheng-etal-2024-helpful}, and flagging potentially harmful personas before $\textsc{PT}_{\textsc{dpo}}$ uses them~\cite{inan2023llama}.

\setlength{\fboxsep}{2pt}

\section{Related Work}



Below, we review research on \mm{} personalization (\cref{subsection:personalization}) and preference subjectivity (\cref{subsection:preference})---showing how they relate to our strategy of inferring personas (\cref{section:persona_inference_setup}) and training models to tailor to them (\cref{section:persona_generation_setup}).

\subsection{Personalization} \label{subsection:personalization}

\mm{} personalization aims to steer models toward user-specific requests, interests, and values~\cite{zhang2024personalization, chen2024large}, improving user trust~\cite{serino2005making}, engagement~\cite{pardini2022role}, and learning~\cite{bernacki2021systematic}.

For personalization, many works use \textit{personas}---textual user descriptions~\cite{10.1145/3313831.3376502}---in prompts~\cite{jandaghi-etal-2024-faithful, liu-etal-2024-evaluating-large, deshpande-etal-2023-toxicity}, but we train \mm{}s with personas.
\citet{stephan2024rlvf} also train \mm{}s with verbal feedback but focus on overgeneralization, not personalization.
\citet{lee2024aligning} similarly test persona training, but use rule-based personas and have GPT-4 produce tailored outputs for their personas, forming synthetic training data.
In contrast, we infer personas in preference data, which has more flexibility and does not require a teacher model that already excels in personalization; the latter is key in specific domains like education where we cannot rely on \mm{}s as ground-truth, but user preference data can be curated~\cite{liu2024aligning}.
We show this empirically via large personalization gains on our mnemonic dataset (Table~\ref{table:add_persona}, Table~\ref{table:response_type}). 
Further, unlike these works, we run a comprehensive evaluation, covering persona accuracy (\cref{subsection:accuracy}), plausibility, and harmfulness (\cref{subsection:persona_qualitative}), as well as response quality with a small user study (\cref{subsection:user_study}).

Lastly, works have tried eliciting personas from a user's interaction history~\cite{li2023eliciting, handa2024bayesian, jin2024implicit}.~Our \inference task is a form of preference elicitation, deriving personas based on pairwise comparisons (\cref{section:persona_inference_setup}).
However, we decompose the training of persona-tailored \mm{}s and the elicitation of personas for inference into separate research questions, focusing on the~former.

\subsection{Preference Subjectivity} \label{subsection:preference}

User preferences are subjective and depend on a user's opinions, traits, and values~\cite{bakker2022fine, kirk2024prism, agnew2024Illusion}.
As a result, researchers have studied social choice theory~\cite{conitzer2024social} and pluralistic alignment~\cite{sorensen2024Roadmap, liu-etal-2024-evaluating-large} to capture subjectivity, executed with Bayesian preference training~\cite{yang2024bayesian, handa2024bayesian}, combining models individually aligned to groups~\cite{chakraborty2024maxmin, hwang-etal-2023-aligning}, multi-\mm{} collaboration~\cite{feng-etal-2024-modular, wu2024aligning}, and multi-objective reward modeling~\cite{wang-etal-2024-arithmetic, zeng-etal-2024-diversified}.

Complementing these approaches, we design \inference, a method that can infer preferences from pairwise comparisons via abductive reasoning---finding evidence to explain outcomes~\cite{peirce1974collected}.
This~reasoning has been used in commonsense~\cite{zhao-etal-2023-abductive}, question answering~\cite{balepur-etal-2024-artifacts}, and reading comprehension~\cite{du-etal-2021-learning} tasks, but we use it to infer personas of users that prefer chosen or rejected outputs.
Very recent and~concurrent work also explores how contexts can change preferences on responses to improve human evaluation~\cite{malaviya2024contextualized} and preference modeling~\cite{pitis2024improving}, but we are the first to treat contexts as personas to tailor responses for users.



\section{Conclusion}

Abductive reasoning---explaining when outputs are preferred---greatly aids personalization via persona inference (\inference) and tailoring (\generation).
\inference infers uncommon but valid reasons to prefer rejected responses which form harder evaluations, so future work can collect more long-tail user needs~\cite{yin2012challenging} to~stress-test personalization.
Notably, our method generalizes to real user needs without curating potentially sensitive data, showing \mm{}-inferred personas can improve personalization with less privacy concerns~\cite{peng-etal-2024-pocketllm}.
Users prefer \generation's outputs, but there are remaining questions: can users faithfully verbalize their needs for prompts~\cite{taylor1962process} and can \generation help users finish tasks~\cite{mozannar2024realhumaneval}? 
Despite \generation's success, we may~need debiasing and abstention~\cite{meade-etal-2022-empirical, wang2024not} to curb harmful or biased personas.

Beyond personalization, personas reveal implicit differences in chosen and rejected outputs; we surface these via persona token saliency, but tools like contrastive topic models may provide finer-grained differences~\cite{zhong2023goal}.
Our~personas convey user needs, but other types---like moral arguments~\cite{rao-etal-2023-makes}, cultural values~\cite{kirk2024prism}, authorship~\cite{wang2023can}, and knowledge levels~\cite{shu-etal-2024-karl}---could adapt \mm{}s to varied constitutions, cultures, writing styles, and user expertise.
Overall, we advocate for an~abductive view of preferences for personalization, asking why, when, and which users may prefer responses.

\section*{Acknowledgments}

We thank the \clip lab at the University of Maryland and our external collaborators for their feedback.
Specifically, we wish to thank Dang Nguyen, Yu Hou, Connor Baumler, Paiheng Xu, Zongxia Li, Feng Gu, Naina Balepur, and Atrey Desai for annotation help.
We are also grateful to Zichao Wang, Vivek Iyer, Brihi Joshi, Robin Jia, John Lalor, and Nischal Kumar for insightful discussions.
This material is based upon work supported by the National Science Foundation under Grant No. \abr{iis}-2403436 (Boyd-Graber), \abr{iis}-2339746 (Rudinger), and \abr{dge}-2236417 (Balepur).
Any opinions, findings, and conclusions or recommendations expressed
in this material are those of the author(s) and do not necessarily
reflect the views of the National Science Foundation.
Cloud computing resources were made possible by a gift from Adobe Research.

\section{Limitations} \label{section:limitations}

One limitation is our \inference strategy uses just one example to infer personas.
In some cases, this may pose the risk of overfitting, but with our \mm{}s, we find personas rarely overfit to the prompt and response (\cref{subsection:persona_qualitative}).
Further, in some tasks, we may desire more than one example to find more nuanced personas.
For instance, from multiple examples, we could infer users prefer short answers for queries on news and long answers for queries on food.
Such a task can further challenge \mm{}s in abductive reasoning and could lead to personalization systems that consider personas from diverse angles.
We find it extremely promising that our strategy of \inference with just one example yields large personalization gains (Figure~\ref{fig:human_qual}), and we hope future work can extend our \inference task to capture diverse, multi-aspect personas.

Our persona tailoring method also assumes there is always an input persona that is relevant to the prompt.
However, as we show in~\cref{subsection:qualitative}, a user could provide input personas that are irrelevant to the prompt, degrading our model outputs.
To address this, future work could explore producing intentionally irrelevant personas and training the model to abstain on them.
Further, to address cases where a specific persona is not needed as input, researchers could route prompts to a model trained without personas, or use a default persona like a general system prompt; we test the latter in Appendix~\ref{appendix:system}.

We also note that we cannot capture all possible personas a model must cater to.
For diversity, we use personas derived from both chosen and rejected responses in offline evaluation (\cref{subsection:persona_type}), resulting in two diverse user needs per input prompt.
Further, in our pilot study, we ask annotators to write three unique personas per input prompt, and we find few exact matches.
We believe there is promising future work in testing how personalization impacts diverse users, including how personalization helps users across cultures~\cite{kirk2024prism}, if personalization truly helps users in downstream tasks~\cite{mozannar2024realhumaneval}, and if users accurately articulate their specific preferences~\cite{handa2024bayesian}.

Lastly, as is true in any machine learning model, there are tradeoffs between efficiency and performance; $\textsc{PT}_\textsc{dpo}$ produces the most personalized responses (\cref{subsection:ablation}), but requires the most resources to train.
Regardless, $\textsc{PT}$ largely enhances personalization with every generation method (i.e., Prompting, \textsc{SFT}, and \textsc{DPO},~\cref{subsection:add_personas}), showing we can accommodate practitioners with varying resource budgets.



\section{Ethical Considerations} \label{section:ethics}

Personalization can raise ethical concerns when using personas tied to sensitive attributes like race or gender, which risks perpetuating biases~\cite{hou-etal-2025-language} and stereotypes~\cite{kantharuban2024stereotype}.
Thus, we instead study high-level interests, personality traits, and needs (\cref{section:persona_inference_setup}).
While users specifying sensitive personas directly is less concerning, issues arise when practitioners use preference elicitation techniques~\cite{li2023eliciting} to infer such personas from past interactions and apply them for personalization.
Although we observed no instances of protected attributes in persona inference, our model trained on ``safe'' personas could still produce harmful responses if prompted to (\cref{subsection:qualitative}).
We urge future works to explore safeguards like classifiers to flag harmful personas pre-inference, and conduct user studies to understand which types of personas users prefer seeing reflected in outputs, informing responsible personalization efforts.

Further, personalization could risk selection or confirmation bias~\cite{hernan2004structural, klayman1995varieties}.
For example, if a user already has a particular view and requests information that is aligned with that view, our model will provide a response that confirms this user's view, which may be harmful in cases such as misinformation~\cite{zhou2022confirmation}.
There is a tradeoff between producing personalized and balanced responses, and researchers can explore future task setups that encourage generating balanced outputs~\cite{zhang2023fair, balepur2025mods} to study its effects.




\bibliography{custom}

\begin{thebibliography}{95}
\expandafter\ifx\csname natexlab\endcsname\relax\def\natexlab#1{#1}\fi

\bibitem[{Achiam et~al.(2023)Achiam, Adler, Agarwal, Ahmad, Akkaya, Aleman, Almeida, Altenschmidt, Altman, Anadkat et~al.}]{achiam2023gpt}
Josh Achiam, Steven Adler, Sandhini Agarwal, Lama Ahmad, Ilge Akkaya, Florencia~Leoni Aleman, Diogo Almeida, Janko Altenschmidt, Sam Altman, Shyamal Anadkat, et~al. 2023.
\newblock Gpt-4 technical report.
\newblock \emph{arXiv preprint arXiv:2303.08774}.

\bibitem[{Agnew et~al.(2024)Agnew, Bergman, Chien, D\'{\i}az, El-Sayed, Pittman, Mohamed, and McKee}]{agnew2024Illusion}
William Agnew, A.~Stevie Bergman, Jennifer Chien, Mark D\'{\i}az, Seliem El-Sayed, Jaylen Pittman, Shakir Mohamed, and Kevin~R. McKee. 2024.
\newblock \href {https://doi.org/10.1145/3613904.3642703} {The illusion of artificial inclusion}.
\newblock In \emph{Proceedings of the 2024 CHI Conference on Human Factors in Computing Systems}, CHI '24, New York, NY, USA. Association for Computing Machinery.

\bibitem[{Anthropic(2023)}]{anthropic2023claude}
Anthropic. 2023.
\newblock Meet claude.
\newblock \url{https://www.anthropic.com/product}.
\newblock Accessed: 2024-09-10.

\bibitem[{Bai et~al.(2022{\natexlab{a}})Bai, Jones, Ndousse, Askell, Chen, DasSarma, Drain, Fort, Ganguli, Henighan et~al.}]{bai2022training}
Yuntao Bai, Andy Jones, Kamal Ndousse, Amanda Askell, Anna Chen, Nova DasSarma, Dawn Drain, Stanislav Fort, Deep Ganguli, Tom Henighan, et~al. 2022{\natexlab{a}}.
\newblock Training a helpful and harmless assistant with reinforcement learning from human feedback.
\newblock \emph{arXiv preprint arXiv:2204.05862}.

\bibitem[{Bai et~al.(2022{\natexlab{b}})Bai, Kadavath, Kundu, Askell, Kernion, Jones, Chen, Goldie, Mirhoseini, McKinnon et~al.}]{bai2022constitutional}
Yuntao Bai, Saurav Kadavath, Sandipan Kundu, Amanda Askell, Jackson Kernion, Andy Jones, Anna Chen, Anna Goldie, Azalia Mirhoseini, Cameron McKinnon, et~al. 2022{\natexlab{b}}.
\newblock Constitutional ai: Harmlessness from ai feedback.
\newblock \emph{arXiv preprint arXiv:2212.08073}.

\bibitem[{Bakker et~al.(2022)Bakker, Chadwick, Sheahan, Tessler, Campbell-Gillingham, Balaguer, McAleese, Glaese, Aslanides, Botvinick et~al.}]{bakker2022fine}
Michiel Bakker, Martin Chadwick, Hannah Sheahan, Michael Tessler, Lucy Campbell-Gillingham, Jan Balaguer, Nat McAleese, Amelia Glaese, John Aslanides, Matt Botvinick, et~al. 2022.
\newblock Fine-tuning language models to find agreement among humans with diverse preferences.
\newblock \emph{Advances in Neural Information Processing Systems}, 35:38176--38189.

\bibitem[{Balepur et~al.(2025{\natexlab{a}})Balepur, Gu, Ravichander, Feng, Boyd-Graber, and Rudinger}]{balepur2024reverse}
Nishant Balepur, Feng Gu, Abhilasha Ravichander, Shi Feng, Jordan~Lee Boyd-Graber, and Rachel Rudinger. 2025{\natexlab{a}}.
\newblock \href {https://aclanthology.org/2025.naacl-short.5/} {Reverse question answering: Can an {LLM} write a question so hard (or bad) that it can`t answer?}
\newblock In \emph{Proceedings of the 2025 Conference of the Nations of the Americas Chapter of the Association for Computational Linguistics: Human Language Technologies (Volume 2: Short Papers)}, pages 44--64, Albuquerque, New Mexico. Association for Computational Linguistics.

\bibitem[{Balepur et~al.(2024{\natexlab{a}})Balepur, Palta, and Rudinger}]{balepur2024s}
Nishant Balepur, Shramay Palta, and Rachel Rudinger. 2024{\natexlab{a}}.
\newblock It’s not easy being wrong: Large language models struggle with process of elimination reasoning.
\newblock In \emph{Findings of the Association for Computational Linguistics ACL 2024}, pages 10143--10166.

\bibitem[{Balepur et~al.(2024{\natexlab{b}})Balepur, Ravichander, and Rudinger}]{balepur-etal-2024-artifacts}
Nishant Balepur, Abhilasha Ravichander, and Rachel Rudinger. 2024{\natexlab{b}}.
\newblock \href {https://aclanthology.org/2024.acl-long.555} {Artifacts or abduction: How do {LLM}s answer multiple-choice questions without the question?}
\newblock In \emph{Proceedings of the 62nd Annual Meeting of the Association for Computational Linguistics (Volume 1: Long Papers)}, pages 10308--10330, Bangkok, Thailand. Association for Computational Linguistics.

\bibitem[{Balepur et~al.(2024{\natexlab{c}})Balepur, Shu, Hoyle, Robey, Feng, Goldfarb-Tarrant, and Boyd-Graber}]{balepur2024smart}
Nishant Balepur, Matthew Shu, Alexander Hoyle, Alison Robey, Shi Feng, Seraphina Goldfarb-Tarrant, and Jordan~Lee Boyd-Graber. 2024{\natexlab{c}}.
\newblock \href {https://doi.org/10.18653/v1/2024.emnlp-main.786} {A {SMART} mnemonic sounds like {``}glue tonic{''}: Mixing {LLM}s with student feedback to make mnemonic learning stick}.
\newblock In \emph{Proceedings of the 2024 Conference on Empirical Methods in Natural Language Processing}, pages 14202--14225, Miami, Florida, USA. Association for Computational Linguistics.

\bibitem[{Balepur et~al.(2025{\natexlab{b}})Balepur, Siu, Lipka, Dernoncourt, Sun, Boyd-Graber, and Mathur}]{balepur2025mods}
Nishant Balepur, Alexa Siu, Nedim Lipka, Franck Dernoncourt, Tong Sun, Jordan~Lee Boyd-Graber, and Puneet Mathur. 2025{\natexlab{b}}.
\newblock \href {https://aclanthology.org/2025.naacl-long.20/} {{M}o{DS}: Moderating a mixture of document speakers to summarize debatable queries in document collections}.
\newblock In \emph{Proceedings of the 2025 Conference of the Nations of the Americas Chapter of the Association for Computational Linguistics: Human Language Technologies (Volume 1: Long Papers)}, pages 465--491, Albuquerque, New Mexico. Association for Computational Linguistics.

\bibitem[{Bernacki et~al.(2021)Bernacki, Greene, and Lobczowski}]{bernacki2021systematic}
Matthew~L Bernacki, Meghan~J Greene, and Nikki~G Lobczowski. 2021.
\newblock A systematic review of research on personalized learning: Personalized by whom, to what, how, and for what purpose (s)?
\newblock \emph{Educational Psychology Review}, 33(4):1675--1715.

\bibitem[{Brown et~al.(2020)Brown, Mann, Ryder, Subbiah, Kaplan, Dhariwal, Neelakantan, Shyam, Sastry, Askell, Agarwal, Herbert-Voss, Krueger, Henighan, Child, Ramesh, Ziegler, Wu, Winter, Hesse, Chen, Sigler, Litwin, Gray, Chess, Clark, Berner, McCandlish, Radford, Sutskever, and Amodei}]{10.5555/3495724.3495883}
Tom~B. Brown, Benjamin Mann, Nick Ryder, Melanie Subbiah, Jared Kaplan, Prafulla Dhariwal, Arvind Neelakantan, Pranav Shyam, Girish Sastry, Amanda Askell, Sandhini Agarwal, Ariel Herbert-Voss, Gretchen Krueger, Tom Henighan, Rewon Child, Aditya Ramesh, Daniel~M. Ziegler, Jeffrey Wu, Clemens Winter, Christopher Hesse, Mark Chen, Eric Sigler, Mateusz Litwin, Scott Gray, Benjamin Chess, Jack Clark, Christopher Berner, Sam McCandlish, Alec Radford, Ilya Sutskever, and Dario Amodei. 2020.
\newblock Language models are few-shot learners.
\newblock In \emph{Proceedings of the 34th International Conference on Neural Information Processing Systems}, NIPS'20, Red Hook, NY, USA. Curran Associates Inc.

\bibitem[{Chakraborty et~al.(2024)Chakraborty, Qiu, Yuan, Koppel, Manocha, Huang, Bedi, and Wang}]{chakraborty2024maxmin}
Souradip Chakraborty, Jiahao Qiu, Hui Yuan, Alec Koppel, Dinesh Manocha, Furong Huang, Amrit Bedi, and Mengdi Wang. 2024.
\newblock \href {https://proceedings.mlr.press/v235/chakraborty24b.html} {{M}ax{M}in-{RLHF}: Alignment with diverse human preferences}.
\newblock In \emph{Proceedings of the 41st International Conference on Machine Learning}, volume 235 of \emph{Proceedings of Machine Learning Research}, pages 6116--6135. PMLR.

\bibitem[{Chen et~al.(2024{\natexlab{a}})Chen, Wang, Xu, Yuan, Zhang, Shi, Xie, Li, Yang, Zhu, Chen, Li, Chen, Hu, Wu, Ren, Fu, and Xiao}]{chen2024persona}
Jiangjie Chen, Xintao Wang, Rui Xu, Siyu Yuan, Yikai Zhang, Wei Shi, Jian Xie, Shuang Li, Ruihan Yang, Tinghui Zhu, Aili Chen, Nianqi Li, Lida Chen, Caiyu Hu, Siye Wu, Scott Ren, Ziquan Fu, and Yanghua Xiao. 2024{\natexlab{a}}.
\newblock \href {https://openreview.net/forum?id=xrO70E8UIZ} {From persona to personalization: A survey on role-playing language agents}.
\newblock \emph{Transactions on Machine Learning Research}.
\newblock Survey Certification.

\bibitem[{Chen et~al.(2024{\natexlab{b}})Chen, Liu, Huang, Wu, Liu, Jiang, Pu, Lei, Chen, Wang, Zheng, Lian, and Chen}]{chen2024large}
Jin Chen, Zheng Liu, Xu~Huang, Chenwang Wu, Qi~Liu, Gangwei Jiang, Yuanhao Pu, Yuxuan Lei, Xiaolong Chen, Xingmei Wang, Kai Zheng, Defu Lian, and Enhong Chen. 2024{\natexlab{b}}.
\newblock \href {https://doi.org/10.1007/s11280-024-01276-1} {When large language models meet personalization: perspectives of challenges and opportunities}.
\newblock \emph{World Wide Web}, 27(4).

\bibitem[{Chung et~al.(2024)Chung, Hou, Longpre, Zoph, Tay, Fedus, Li, Wang, Dehghani, Brahma, Webson, Gu, Dai, Suzgun, Chen, Chowdhery, Castro-Ros, Pellat, Robinson, Valter, Narang, Mishra, Yu, Zhao, Huang, Dai, Yu, Petrov, Chi, Dean, Devlin, Roberts, Zhou, Le, and Wei}]{JMLR:v25:23-0870}
Hyung~Won Chung, Le~Hou, Shayne Longpre, Barret Zoph, Yi~Tay, William Fedus, Yunxuan Li, Xuezhi Wang, Mostafa Dehghani, Siddhartha Brahma, Albert Webson, Shixiang~Shane Gu, Zhuyun Dai, Mirac Suzgun, Xinyun Chen, Aakanksha Chowdhery, Alex Castro-Ros, Marie Pellat, Kevin Robinson, Dasha Valter, Sharan Narang, Gaurav Mishra, Adams Yu, Vincent Zhao, Yanping Huang, Andrew Dai, Hongkun Yu, Slav Petrov, Ed~H. Chi, Jeff Dean, Jacob Devlin, Adam Roberts, Denny Zhou, Quoc~V. Le, and Jason Wei. 2024.
\newblock \href {http://jmlr.org/papers/v25/23-0870.html} {Scaling instruction-finetuned language models}.
\newblock \emph{Journal of Machine Learning Research}, 25(70):1--53.

\bibitem[{Conitzer et~al.(2024)Conitzer, Freedman, Heitzig, Holliday, Jacobs, Lambert, Mosse, Pacuit, Russell, Schoelkopf et~al.}]{conitzer2024social}
Vincent Conitzer, Rachel Freedman, Jobst Heitzig, Wesley~H Holliday, Bob~M Jacobs, Nathan Lambert, Milan Mosse, Eric Pacuit, Stuart Russell, Hailey Schoelkopf, et~al. 2024.
\newblock Social choice for ai alignment: Dealing with diverse human feedback.
\newblock In \emph{International Joint Conference on Artificial Intelligence 2024 Workshop on AI Governance: Alignment, Morality, and Law}.

\bibitem[{Deb et~al.(2022)Deb, Awadallah, and Zheng}]{deb-etal-2022-boosting}
Budhaditya Deb, Ahmed~Hassan Awadallah, and Guoqing Zheng. 2022.
\newblock \href {https://doi.org/10.18653/v1/2022.emnlp-main.456} {Boosting natural language generation from instructions with meta-learning}.
\newblock In \emph{Proceedings of the 2022 Conference on Empirical Methods in Natural Language Processing}, pages 6792--6808, Abu Dhabi, United Arab Emirates. Association for Computational Linguistics.

\bibitem[{Deshpande et~al.(2023)Deshpande, Murahari, Rajpurohit, Kalyan, and Narasimhan}]{deshpande-etal-2023-toxicity}
Ameet Deshpande, Vishvak Murahari, Tanmay Rajpurohit, Ashwin Kalyan, and Karthik Narasimhan. 2023.
\newblock \href {https://doi.org/10.18653/v1/2023.findings-emnlp.88} {Toxicity in chatgpt: Analyzing persona-assigned language models}.
\newblock In \emph{Findings of the Association for Computational Linguistics: EMNLP 2023}, pages 1236--1270, Singapore. Association for Computational Linguistics.

\bibitem[{Dror et~al.(2018)Dror, Baumer, Shlomov, and Reichart}]{dror-etal-2018-hitchhikers}
Rotem Dror, Gili Baumer, Segev Shlomov, and Roi Reichart. 2018.
\newblock \href {https://doi.org/10.18653/v1/P18-1128} {The hitchhiker{'}s guide to testing statistical significance in natural language processing}.
\newblock In \emph{Proceedings of the 56th Annual Meeting of the Association for Computational Linguistics (Volume 1: Long Papers)}, pages 1383--1392, Melbourne, Australia. Association for Computational Linguistics.

\bibitem[{Du et~al.(2021)Du, Ding, Liu, and Qin}]{du-etal-2021-learning}
Li~Du, Xiao Ding, Ting Liu, and Bing Qin. 2021.
\newblock \href {https://doi.org/10.18653/v1/2021.acl-long.403} {Learning event graph knowledge for abductive reasoning}.
\newblock In \emph{Proceedings of the 59th Annual Meeting of the Association for Computational Linguistics and the 11th International Joint Conference on Natural Language Processing (Volume 1: Long Papers)}, pages 5181--5190, Online. Association for Computational Linguistics.

\bibitem[{Dubey et~al.(2024)Dubey, Jauhri, Pandey, Kadian, Al-Dahle, Letman, Mathur, Schelten, Yang, Fan et~al.}]{dubey2024llama}
Abhimanyu Dubey, Abhinav Jauhri, Abhinav Pandey, Abhishek Kadian, Ahmad Al-Dahle, Aiesha Letman, Akhil Mathur, Alan Schelten, Amy Yang, Angela Fan, et~al. 2024.
\newblock The llama 3 herd of models.
\newblock \emph{arXiv preprint arXiv:2407.21783}.

\bibitem[{Ethayarajh et~al.(2022)Ethayarajh, Choi, and Swayamdipta}]{pmlr-v162-ethayarajh22a}
Kawin Ethayarajh, Yejin Choi, and Swabha Swayamdipta. 2022.
\newblock Understanding dataset difficulty with $\mathcal{V}$-usable information.
\newblock In \emph{Proceedings of the 39th International Conference on Machine Learning}, volume 162 of \emph{Proceedings of Machine Learning Research}, pages 5988--6008. PMLR.

\bibitem[{Fan et~al.(2019)Fan, Jernite, Perez, Grangier, Weston, and Auli}]{ELIFIVE}
Angela Fan, Yacine Jernite, Ethan Perez, David Grangier, Jason Weston, and Michael Auli. 2019.
\newblock \href {https://doi.org/10.18653/v1/p19-1346} {{ELI5:} long form question answering}.
\newblock In \emph{Proceedings of the 57th Conference of the Association for Computational Linguistics, {ACL} 2019, Florence, Italy, July 28- August 2, 2019, Volume 1: Long Papers}, pages 3558--3567. Association for Computational Linguistics.

\bibitem[{Feng et~al.(2024)Feng, Sorensen, Liu, Fisher, Park, Choi, and Tsvetkov}]{feng-etal-2024-modular}
Shangbin Feng, Taylor Sorensen, Yuhan Liu, Jillian Fisher, Chan~Young Park, Yejin Choi, and Yulia Tsvetkov. 2024.
\newblock \href {https://aclanthology.org/2024.emnlp-main.240} {Modular pluralism: Pluralistic alignment via multi-{LLM} collaboration}.
\newblock In \emph{Proceedings of the 2024 Conference on Empirical Methods in Natural Language Processing}, pages 4151--4171, Miami, Florida, USA. Association for Computational Linguistics.

\bibitem[{Fleiss et~al.(1981)Fleiss, Levin, Paik et~al.}]{fleiss1981measurement}
Joseph~L Fleiss, Bruce Levin, Myunghee~Cho Paik, et~al. 1981.
\newblock The measurement of interrater agreement.
\newblock \emph{Statistical methods for rates and proportions}, 2(212-236):22--23.

\bibitem[{Gou et~al.(2021)Gou, Yu, Maybank, and Tao}]{gou2021knowledge}
Jianping Gou, Baosheng Yu, Stephen~J Maybank, and Dacheng Tao. 2021.
\newblock Knowledge distillation: A survey.
\newblock \emph{International Journal of Computer Vision}, 129(6):1789--1819.

\bibitem[{Handa et~al.(2024)Handa, Gal, Pavlick, Goodman, Andreas, Tamkin, and Li}]{handa2024bayesian}
Kunal Handa, Yarin Gal, Ellie Pavlick, Noah Goodman, Jacob Andreas, Alex Tamkin, and Belinda~Z Li. 2024.
\newblock Bayesian preference elicitation with language models.
\newblock \emph{arXiv preprint arXiv:2403.05534}.

\bibitem[{Hern{\'a}n et~al.(2004)Hern{\'a}n, Hern{\'a}ndez-D{\'\i}az, and Robins}]{hernan2004structural}
Miguel~A Hern{\'a}n, Sonia Hern{\'a}ndez-D{\'\i}az, and James~M Robins. 2004.
\newblock A structural approach to selection bias.
\newblock \emph{Epidemiology}, 15(5):615--625.

\bibitem[{Hou et~al.(2025)Hou, Iii, and Rudinger}]{hou-etal-2025-language}
Yu~Hou, Hal~Daum{\'e} Iii, and Rachel Rudinger. 2025.
\newblock \href {https://aclanthology.org/2025.naacl-long.611/} {Language models predict empathy gaps between social in-groups and out-groups}.
\newblock In \emph{Proceedings of the 2025 Conference of the Nations of the Americas Chapter of the Association for Computational Linguistics: Human Language Technologies (Volume 1: Long Papers)}, pages 12288--12304, Albuquerque, New Mexico. Association for Computational Linguistics.

\bibitem[{Hoyle et~al.(2023)Hoyle, Sarkar, Goel, and Resnik}]{hoyle-etal-2023-natural}
Alexander Hoyle, Rupak Sarkar, Pranav Goel, and Philip Resnik. 2023.
\newblock \href {https://doi.org/10.18653/v1/2023.emnlp-main.815} {Natural language decompositions of implicit content enable better text representations}.
\newblock In \emph{Proceedings of the 2023 Conference on Empirical Methods in Natural Language Processing}, pages 13188--13214, Singapore. Association for Computational Linguistics.

\bibitem[{Hu et~al.(2022)Hu, yelong shen, Wallis, Allen-Zhu, Li, Wang, Wang, and Chen}]{hu2021lora}
Edward~J Hu, yelong shen, Phillip Wallis, Zeyuan Allen-Zhu, Yuanzhi Li, Shean Wang, Lu~Wang, and Weizhu Chen. 2022.
\newblock \href {https://openreview.net/forum?id=nZeVKeeFYf9} {Lo{RA}: Low-rank adaptation of large language models}.
\newblock In \emph{International Conference on Learning Representations}.

\bibitem[{Hwang et~al.(2023)Hwang, Majumder, and Tandon}]{hwang-etal-2023-aligning}
EunJeong Hwang, Bodhisattwa Majumder, and Niket Tandon. 2023.
\newblock \href {https://doi.org/10.18653/v1/2023.findings-emnlp.393} {Aligning language models to user opinions}.
\newblock In \emph{Findings of the Association for Computational Linguistics: EMNLP 2023}, pages 5906--5919, Singapore. Association for Computational Linguistics.

\bibitem[{Inan et~al.(2023)Inan, Upasani, Chi, Rungta, Iyer, Mao, Tontchev, Hu, Fuller, Testuggine et~al.}]{inan2023llama}
Hakan Inan, Kartikeya Upasani, Jianfeng Chi, Rashi Rungta, Krithika Iyer, Yuning Mao, Michael Tontchev, Qing Hu, Brian Fuller, Davide Testuggine, et~al. 2023.
\newblock Llama guard: Llm-based input-output safeguard for human-ai conversations.
\newblock \emph{arXiv preprint arXiv:2312.06674}.

\bibitem[{Jandaghi et~al.(2024)Jandaghi, Sheng, Bai, Pujara, and Sidahmed}]{jandaghi-etal-2024-faithful}
Pegah Jandaghi, Xianghai Sheng, Xinyi Bai, Jay Pujara, and Hakim Sidahmed. 2024.
\newblock \href {https://aclanthology.org/2024.nlp4convai-1.8} {Faithful persona-based conversational dataset generation with large language models}.
\newblock In \emph{Proceedings of the 6th Workshop on NLP for Conversational AI (NLP4ConvAI 2024)}, pages 114--139, Bangkok, Thailand. Association for Computational Linguistics.

\bibitem[{Jang et~al.(2024)Jang, Kim, Lin, Wang, Hessel, Zettlemoyer, Hajishirzi, Choi, and Ammanabrolu}]{jang2023personalized}
Joel Jang, Seungone Kim, Bill~Yuchen Lin, Yizhong Wang, Jack Hessel, Luke Zettlemoyer, Hannaneh Hajishirzi, Yejin Choi, and Prithviraj Ammanabrolu. 2024.
\newblock \href {https://openreview.net/forum?id=EMrnoPRvxe} {Personalized soups: Personalized large language model alignment via post-hoc parameter merging}.
\newblock In \emph{Adaptive Foundation Models: Evolving AI for Personalized and Efficient Learning}.

\bibitem[{Ji et~al.(2024)Ji, Liu, Dai, Pan, Zhang, Bian, Chen, Sun, Wang, and Yang}]{ji2024beavertails}
Jiaming Ji, Mickel Liu, Josef Dai, Xuehai Pan, Chi Zhang, Ce~Bian, Boyuan Chen, Ruiyang Sun, Yizhou Wang, and Yaodong Yang. 2024.
\newblock Beavertails: Towards improved safety alignment of llm via a human-preference dataset.
\newblock \emph{Advances in Neural Information Processing Systems}, 36.

\bibitem[{Ji et~al.(2023)Ji, Qiu, Chen, Zhang, Lou, Wang, Duan, He, Zhou, Zhang et~al.}]{ji2023ai}
Jiaming Ji, Tianyi Qiu, Boyuan Chen, Borong Zhang, Hantao Lou, Kaile Wang, Yawen Duan, Zhonghao He, Jiayi Zhou, Zhaowei Zhang, et~al. 2023.
\newblock Ai alignment: A comprehensive survey.
\newblock \emph{arXiv preprint arXiv:2310.19852}.

\bibitem[{Jin et~al.(2024)Jin, Heil, Liu, Dhuliawala, Qi, Sch{\"o}lkopf, Mihalcea, and Sachan}]{jin2024implicit}
Zhijing Jin, Nils Heil, Jiarui Liu, Shehzaad Dhuliawala, Yahang Qi, Bernhard Sch{\"o}lkopf, Rada Mihalcea, and Mrinmaya Sachan. 2024.
\newblock \href {https://doi.org/10.18653/v1/2024.findings-emnlp.717} {Implicit personalization in language models: A systematic study}.
\newblock In \emph{Findings of the Association for Computational Linguistics: EMNLP 2024}, pages 12309--12325, Miami, Florida, USA. Association for Computational Linguistics.

\bibitem[{Jin et~al.(2021)Jin, von K{\"u}gelgen, Ni, Vaidhya, Kaushal, Sachan, and Schoelkopf}]{jin-etal-2021-causal}
Zhijing Jin, Julius von K{\"u}gelgen, Jingwei Ni, Tejas Vaidhya, Ayush Kaushal, Mrinmaya Sachan, and Bernhard Schoelkopf. 2021.
\newblock \href {https://doi.org/10.18653/v1/2021.emnlp-main.748} {Causal direction of data collection matters: Implications of causal and anticausal learning for {NLP}}.
\newblock In \emph{Proceedings of the 2021 Conference on Empirical Methods in Natural Language Processing}, pages 9499--9513, Online and Punta Cana, Dominican Republic. Association for Computational Linguistics.

\bibitem[{Joshi et~al.(2025)Joshi, Ren, Swayamdipta, Koncel-Kedziorski, and Paek}]{joshi2025improving}
Brihi Joshi, Xiang Ren, Swabha Swayamdipta, Rik Koncel-Kedziorski, and Tim Paek. 2025.
\newblock Improving llm personas via rationalization with psychological scaffolds.
\newblock \emph{arXiv preprint arXiv:2504.17993}.

\bibitem[{Kantharuban et~al.(2024)Kantharuban, Milbauer, Strubell, and Neubig}]{kantharuban2024stereotype}
Anjali Kantharuban, Jeremiah Milbauer, Emma Strubell, and Graham Neubig. 2024.
\newblock Stereotype or personalization? user identity biases chatbot recommendations.
\newblock \emph{arXiv preprint arXiv:2410.05613}.

\bibitem[{Kim et~al.(2024)Kim, Suk, Longpre, Lin, Shin, Welleck, Neubig, Lee, Lee, and Seo}]{kim2024prometheus}
Seungone Kim, Juyoung Suk, Shayne Longpre, Bill~Yuchen Lin, Jamin Shin, Sean Welleck, Graham Neubig, Moontae Lee, Kyungjae Lee, and Minjoon Seo. 2024.
\newblock \href {https://aclanthology.org/2024.emnlp-main.248} {Prometheus 2: An open source language model specialized in evaluating other language models}.
\newblock In \emph{Proceedings of the 2024 Conference on Empirical Methods in Natural Language Processing}, pages 4334--4353, Miami, Florida, USA. Association for Computational Linguistics.

\bibitem[{Kirk et~al.(2024)Kirk, Whitefield, R{\"o}ttger, Bean, Margatina, Mosquera, Ciro, Bartolo, Williams, He, Vidgen, and Hale}]{kirk2024prism}
Hannah~Rose Kirk, Alexander Whitefield, Paul R{\"o}ttger, Andrew~Michael Bean, Katerina Margatina, Rafael Mosquera, Juan~Manuel Ciro, Max Bartolo, Adina Williams, He~He, Bertie Vidgen, and Scott~A. Hale. 2024.
\newblock \href {https://openreview.net/forum?id=DFr5hteojx} {The {PRISM} alignment dataset: What participatory, representative and individualised human feedback reveals about the subjective and multicultural alignment of large language models}.
\newblock In \emph{The Thirty-eight Conference on Neural Information Processing Systems Datasets and Benchmarks Track}.

\bibitem[{Klayman(1995)}]{klayman1995varieties}
Joshua Klayman. 1995.
\newblock Varieties of confirmation bias.
\newblock \emph{Psychology of learning and motivation}, 32:385--418.

\bibitem[{K\"{o}pf et~al.(2024)K\"{o}pf, Kilcher, von R\"{u}tte, Anagnostidis, Tam, Stevens, Barhoum, Duc, Stanley, Nagyfi, ES, Suri, Glushkov, Dantuluri, Maguire, Schuhmann, Nguyen, and Mattick}]{OpenAssistant}
Andreas K\"{o}pf, Yannic Kilcher, Dimitri von R\"{u}tte, Sotiris Anagnostidis, Zhi-Rui Tam, Keith Stevens, Abdullah Barhoum, Nguyen~Minh Duc, Oliver Stanley, Rich\'{a}rd Nagyfi, Shahul ES, Sameer Suri, David Glushkov, Arnav Dantuluri, Andrew Maguire, Christoph Schuhmann, Huu Nguyen, and Alexander Mattick. 2024.
\newblock Openassistant conversations - democratizing large language model alignment.
\newblock In \emph{Proceedings of the 37th International Conference on Neural Information Processing Systems}, NIPS '23, Red Hook, NY, USA. Curran Associates Inc.

\bibitem[{Lee et~al.(2024)Lee, Park, Kim, and Seo}]{lee2024aligning}
Seongyun Lee, Sue~Hyun Park, Seungone Kim, and Minjoon Seo. 2024.
\newblock \href {https://openreview.net/forum?id=recsheQ7e8} {Aligning to thousands of preferences via system message generalization}.
\newblock In \emph{The Thirty-eighth Annual Conference on Neural Information Processing Systems}.

\bibitem[{Li et~al.(2025)Li, Tamkin, Goodman, and Andreas}]{li2023eliciting}
Belinda~Z. Li, Alex Tamkin, Noah Goodman, and Jacob Andreas. 2025.
\newblock \href {https://openreview.net/forum?id=LvDwwAgMEW} {Eliciting human preferences with language models}.
\newblock In \emph{The Thirteenth International Conference on Learning Representations}.

\bibitem[{Liu et~al.(2024{\natexlab{a}})Liu, Diab, and Fried}]{liu-etal-2024-evaluating-large}
Andy Liu, Mona Diab, and Daniel Fried. 2024{\natexlab{a}}.
\newblock \href {https://doi.org/10.18653/v1/2024.findings-acl.586} {Evaluating large language model biases in persona-steered generation}.
\newblock In \emph{Findings of the Association for Computational Linguistics: ACL 2024}, pages 9832--9850, Bangkok, Thailand. Association for Computational Linguistics.

\bibitem[{Liu et~al.(2024{\natexlab{b}})Liu, Zhao, Joshi, Khalman, Saleh, Liu, and Liu}]{liu2023statistical}
Tianqi Liu, Yao Zhao, Rishabh Joshi, Misha Khalman, Mohammad Saleh, Peter~J Liu, and Jialu Liu. 2024{\natexlab{b}}.
\newblock \href {https://openreview.net/forum?id=xbjSwwrQOe} {Statistical rejection sampling improves preference optimization}.
\newblock In \emph{The Twelfth International Conference on Learning Representations}.

\bibitem[{Liu et~al.(2023)Liu, Iter, Xu, Wang, Xu, and Zhu}]{liu-etal-2023-g}
Yang Liu, Dan Iter, Yichong Xu, Shuohang Wang, Ruochen Xu, and Chenguang Zhu. 2023.
\newblock \href {https://doi.org/10.18653/v1/2023.emnlp-main.153} {{G}-eval: {NLG} evaluation using gpt-4 with better human alignment}.
\newblock In \emph{Proceedings of the 2023 Conference on Empirical Methods in Natural Language Processing}, pages 2511--2522, Singapore. Association for Computational Linguistics.

\bibitem[{Liu et~al.(2024{\natexlab{c}})Liu, Zhang, Yao, Cao, Hou, and Li}]{liu2024aligning}
Yantao Liu, Zhao Zhang, Zijun Yao, Shulin Cao, Lei Hou, and Juanzi Li. 2024{\natexlab{c}}.
\newblock Aligning teacher with student preferences for tailored training data generation.
\newblock \emph{arXiv preprint arXiv:2406.19227}.

\bibitem[{Malaviya et~al.(2024)Malaviya, Chang, Roth, Iyyer, Yatskar, and Lo}]{malaviya2024contextualized}
Chaitanya Malaviya, Joseph~Chee Chang, Dan Roth, Mohit Iyyer, Mark Yatskar, and Kyle Lo. 2024.
\newblock Contextualized evaluations: Taking the guesswork out of language model evaluations.
\newblock \emph{arXiv preprint arXiv:2411.07237}.

\bibitem[{Meade et~al.(2022)Meade, Poole-Dayan, and Reddy}]{meade-etal-2022-empirical}
Nicholas Meade, Elinor Poole-Dayan, and Siva Reddy. 2022.
\newblock \href {https://doi.org/10.18653/v1/2022.acl-long.132} {An empirical survey of the effectiveness of debiasing techniques for pre-trained language models}.
\newblock In \emph{Proceedings of the 60th Annual Meeting of the Association for Computational Linguistics (Volume 1: Long Papers)}, pages 1878--1898, Dublin, Ireland. Association for Computational Linguistics.

\bibitem[{Mozannar et~al.(2025)Mozannar, Chen, Alsobay, Das, Zhao, Wei, Nagireddy, Sattigeri, Talwalkar, and Sontag}]{mozannar2024realhumaneval}
Hussein Mozannar, Valerie Chen, Mohammed Alsobay, Subhro Das, Sebastian Zhao, Dennis Wei, Manish Nagireddy, Prasanna Sattigeri, Ameet Talwalkar, and David Sontag. 2025.
\newblock \href {https://openreview.net/forum?id=hGaWq5Buj7} {The realhumaneval: Evaluating large language models{\textquoteright} abilities to support programmers}.
\newblock \emph{Transactions on Machine Learning Research}.
\newblock Expert Certification.

\bibitem[{Ouyang et~al.(2022)Ouyang, Wu, Jiang, Almeida, Wainwright, Mishkin, Zhang, Agarwal, Slama, Ray et~al.}]{ouyang2022training}
Long Ouyang, Jeffrey Wu, Xu~Jiang, Diogo Almeida, Carroll Wainwright, Pamela Mishkin, Chong Zhang, Sandhini Agarwal, Katarina Slama, Alex Ray, et~al. 2022.
\newblock Training language models to follow instructions with human feedback.
\newblock \emph{Advances in neural information processing systems}, 35:27730--27744.

\bibitem[{Padmakumar et~al.(2024)Padmakumar, Jin, Kirk, and He}]{padmakumar2024beyond}
Vishakh Padmakumar, Chuanyang Jin, Hannah~Rose Kirk, and He~He. 2024.
\newblock \href {https://openreview.net/forum?id=M2Yqg68jVW} {Beyond the binary: Capturing diverse preferences with reward regularization}.
\newblock In \emph{Workshop on Socially Responsible Language Modelling Research}.

\bibitem[{Pardini et~al.(2022)Pardini, Gabrielli, Dianti, Novara, Zucco, Mich, and Forti}]{pardini2022role}
Susanna Pardini, Silvia Gabrielli, Marco Dianti, Caterina Novara, Gesualdo~M Zucco, Ornella Mich, and Stefano Forti. 2022.
\newblock The role of personalization in the user experience, preferences and engagement with virtual reality environments for relaxation.
\newblock \emph{International Journal of Environmental Research and Public Health}, 19(12):7237.

\bibitem[{Peirce(1974)}]{peirce1974collected}
Charles~Sanders Peirce. 1974.
\newblock \emph{Collected papers of charles sanders peirce}, volume~5.
\newblock Harvard University Press.

\bibitem[{Peng et~al.(2024)Peng, Fu, and Wang}]{peng-etal-2024-pocketllm}
Dan Peng, Zhihui Fu, and Jun Wang. 2024.
\newblock \href {https://aclanthology.org/2024.privatenlp-1.10/} {{P}ocket{LLM}: Enabling on-device fine-tuning for personalized {LLM}s}.
\newblock In \emph{Proceedings of the Fifth Workshop on Privacy in Natural Language Processing}, pages 91--96, Bangkok, Thailand. Association for Computational Linguistics.

\bibitem[{Pitis et~al.(2024)Pitis, Xiao, Roux, and Sordoni}]{pitis2024improving}
Silviu Pitis, Ziang Xiao, Nicolas~Le Roux, and Alessandro Sordoni. 2024.
\newblock \href {https://openreview.net/forum?id=52r4XJYzjg} {Improving context-aware preference modeling for language models}.
\newblock In \emph{The Thirty-eighth Annual Conference on Neural Information Processing Systems}.

\bibitem[{Rafailov et~al.(2024)Rafailov, Sharma, Mitchell, Manning, Ermon, and Finn}]{rafailov2024direct}
Rafael Rafailov, Archit Sharma, Eric Mitchell, Christopher~D Manning, Stefano Ermon, and Chelsea Finn. 2024.
\newblock Direct preference optimization: Your language model is secretly a reward model.
\newblock \emph{Advances in Neural Information Processing Systems}, 36.

\bibitem[{Rao et~al.(2023)Rao, Jiang, Pyatkin, Gu, Tandon, Dziri, Brahman, and Choi}]{rao-etal-2023-makes}
Kavel Rao, Liwei Jiang, Valentina Pyatkin, Yuling Gu, Niket Tandon, Nouha Dziri, Faeze Brahman, and Yejin Choi. 2023.
\newblock \href {https://doi.org/10.18653/v1/2023.findings-emnlp.812} {What makes it ok to set a fire? iterative self-distillation of contexts and rationales for disambiguating defeasible social and moral situations}.
\newblock In \emph{Findings of the Association for Computational Linguistics: EMNLP 2023}, pages 12140--12159, Singapore. Association for Computational Linguistics.

\bibitem[{Salemi et~al.(2024)Salemi, Mysore, Bendersky, and Zamani}]{salemi-etal-2024-lamp}
Alireza Salemi, Sheshera Mysore, Michael Bendersky, and Hamed Zamani. 2024.
\newblock \href {https://doi.org/10.18653/v1/2024.acl-long.399} {{L}a{MP}: When large language models meet personalization}.
\newblock In \emph{Proceedings of the 62nd Annual Meeting of the Association for Computational Linguistics (Volume 1: Long Papers)}, pages 7370--7392, Bangkok, Thailand. Association for Computational Linguistics.

\bibitem[{Salminen et~al.(2020)Salminen, Guan, Jung, Chowdhury, and Jansen}]{10.1145/3313831.3376502}
Joni Salminen, Kathleen Guan, Soon-Gyo Jung, Shammur~A. Chowdhury, and Bernard~J. Jansen. 2020.
\newblock \href {https://doi.org/10.1145/3313831.3376502} {A literature review of quantitative persona creation}.
\newblock In \emph{Proceedings of the 2020 CHI Conference on Human Factors in Computing Systems}, CHI '20, page 1–14, New York, NY, USA. Association for Computing Machinery.

\bibitem[{Santhanam et~al.(2022)Santhanam, Khattab, Saad-Falcon, Potts, and Zaharia}]{santhanam-etal-2022-colbertv2}
Keshav Santhanam, Omar Khattab, Jon Saad-Falcon, Christopher Potts, and Matei Zaharia. 2022.
\newblock \href {https://doi.org/10.18653/v1/2022.naacl-main.272} {{C}ol{BERT}v2: Effective and efficient retrieval via lightweight late interaction}.
\newblock In \emph{Proceedings of the 2022 Conference of the North American Chapter of the Association for Computational Linguistics: Human Language Technologies}, pages 3715--3734, Seattle, United States. Association for Computational Linguistics.

\bibitem[{Schulhoff et~al.(2024)Schulhoff, Ilie, Balepur, Kahadze, Liu, Si, Li, Gupta, Han, Schulhoff et~al.}]{schulhoff2024prompt}
Sander Schulhoff, Michael Ilie, Nishant Balepur, Konstantine Kahadze, Amanda Liu, Chenglei Si, Yinheng Li, Aayush Gupta, HyoJung Han, Sevien Schulhoff, et~al. 2024.
\newblock The prompt report: A systematic survey of prompting techniques.
\newblock \emph{arXiv preprint arXiv:2406.06608}.

\bibitem[{Serino et~al.(2005)Serino, Furner, and Smatt}]{serino2005making}
Catharina~M Serino, Christopher~P Furner, and Cindi Smatt. 2005.
\newblock Making it personal: How personalization affects trust over time.
\newblock In \emph{Proceedings of the 38th annual Hawaii international conference on system sciences}, pages 170a--170a. IEEE.

\bibitem[{Sharma et~al.(2024)Sharma, Tong, Korbak, Duvenaud, Askell, Bowman, DURMUS, Hatfield-Dodds, Johnston, Kravec, Maxwell, McCandlish, Ndousse, Rausch, Schiefer, Yan, Zhang, and Perez}]{sharma2023towards}
Mrinank Sharma, Meg Tong, Tomasz Korbak, David Duvenaud, Amanda Askell, Samuel~R. Bowman, Esin DURMUS, Zac Hatfield-Dodds, Scott~R Johnston, Shauna~M Kravec, Timothy Maxwell, Sam McCandlish, Kamal Ndousse, Oliver Rausch, Nicholas Schiefer, Da~Yan, Miranda Zhang, and Ethan Perez. 2024.
\newblock \href {https://openreview.net/forum?id=tvhaxkMKAn} {Towards understanding sycophancy in language models}.
\newblock In \emph{The Twelfth International Conference on Learning Representations}.

\bibitem[{Shu et~al.(2024)Shu, Balepur, Feng, and Boyd-Graber}]{shu-etal-2024-karl}
Matthew Shu, Nishant Balepur, Shi Feng, and Jordan~Lee Boyd-Graber. 2024.
\newblock \href {https://doi.org/10.18653/v1/2024.emnlp-main.784} {{KARL}: Knowledge-aware retrieval and representations aid retention and learning in students}.
\newblock In \emph{Proceedings of the 2024 Conference on Empirical Methods in Natural Language Processing}, pages 14161--14178, Miami, Florida, USA. Association for Computational Linguistics.

\bibitem[{Sorensen et~al.(2024)Sorensen, Moore, Fisher, Gordon, Mireshghallah, Rytting, Ye, Jiang, Lu, Dziri, Althoff, and Choi}]{sorensen2024Roadmap}
Taylor Sorensen, Jared Moore, Jillian Fisher, Mitchell~L. Gordon, Niloofar Mireshghallah, Christopher~Michael Rytting, Andre Ye, Liwei Jiang, Ximing Lu, Nouha Dziri, Tim Althoff, and Yejin Choi. 2024.
\newblock \href {https://openreview.net/forum?id=gQpBnRHwxM} {Position: A roadmap to pluralistic alignment}.
\newblock In \emph{ICML}.

\bibitem[{Stephan et~al.(2024)Stephan, Khazatsky, Mitchell, Chen, Hsu, Sharma, and Finn}]{stephan2024rlvf}
Moritz Stephan, Alexander Khazatsky, Eric Mitchell, Annie~S Chen, Sheryl Hsu, Archit Sharma, and Chelsea Finn. 2024.
\newblock Rlvf: learning from verbal feedback without overgeneralization.
\newblock In \emph{Proceedings of the 41st International Conference on Machine Learning}, ICML'24. JMLR.org.

\bibitem[{Taylor(1962)}]{taylor1962process}
Robert~S Taylor. 1962.
\newblock The process of asking questions.
\newblock \emph{American documentation}, 13(4):391--396.

\bibitem[{Tseng et~al.(2024)Tseng, Huang, Hsiao, Chen, Huang, Meng, and Chen}]{tseng2024two}
Yu-Min Tseng, Yu-Chao Huang, Teng-Yun Hsiao, Wei-Lin Chen, Chao-Wei Huang, Yu~Meng, and Yun-Nung Chen. 2024.
\newblock \href {https://doi.org/10.18653/v1/2024.findings-emnlp.969} {Two tales of persona in {LLM}s: A survey of role-playing and personalization}.
\newblock In \emph{Findings of the Association for Computational Linguistics: EMNLP 2024}, pages 16612--16631, Miami, Florida, USA. Association for Computational Linguistics.

\bibitem[{V{\"o}lske et~al.(2017)V{\"o}lske, Potthast, Syed, and Stein}]{volske-etal-2017-tl}
Michael V{\"o}lske, Martin Potthast, Shahbaz Syed, and Benno Stein. 2017.
\newblock \href {https://doi.org/10.18653/v1/W17-4508} {{TL};{DR}: Mining {R}eddit to learn automatic summarization}.
\newblock In \emph{Proceedings of the Workshop on New Frontiers in Summarization}, pages 59--63, Copenhagen, Denmark. Association for Computational Linguistics.

\bibitem[{Wang et~al.(2023)Wang, Aggazzotti, Kotula, Soto, Bishop, and Andrews}]{wang2023can}
Andrew Wang, Cristina Aggazzotti, Rebecca Kotula, Rafael~Rivera Soto, Marcus Bishop, and Nicholas Andrews. 2023.
\newblock Can authorship representation learning capture stylistic features?
\newblock \emph{Transactions of the Association for Computational Linguistics}, 11:1416--1431.

\bibitem[{Wang et~al.(2024{\natexlab{a}})Wang, Lin, Xiong, Yang, Diao, Qiu, Zhao, and Zhang}]{wang-etal-2024-arithmetic}
Haoxiang Wang, Yong Lin, Wei Xiong, Rui Yang, Shizhe Diao, Shuang Qiu, Han Zhao, and Tong Zhang. 2024{\natexlab{a}}.
\newblock \href {https://doi.org/10.18653/v1/2024.acl-long.468} {Arithmetic control of {LLM}s for diverse user preferences: Directional preference alignment with multi-objective rewards}.
\newblock In \emph{Proceedings of the 62nd Annual Meeting of the Association for Computational Linguistics (Volume 1: Long Papers)}, pages 8642--8655, Bangkok, Thailand. Association for Computational Linguistics.

\bibitem[{Wang et~al.(2024{\natexlab{b}})Wang, Ribeiro, Robinson, Loeb, and Demszky}]{wang2024tutor}
Rose~E Wang, Ana~T Ribeiro, Carly~D Robinson, Susanna Loeb, and Dora Demszky. 2024{\natexlab{b}}.
\newblock Tutor copilot: A human-ai approach for scaling real-time expertise.
\newblock \emph{arXiv preprint arXiv:2410.03017}.

\bibitem[{Wang et~al.(2022)Wang, Xu, Fang, Liu, Sun, Xu, Zhu, and Zeng}]{wang-etal-2022-training}
Shuohang Wang, Yichong Xu, Yuwei Fang, Yang Liu, Siqi Sun, Ruochen Xu, Chenguang Zhu, and Michael Zeng. 2022.
\newblock \href {https://doi.org/10.18653/v1/2022.acl-long.226} {Training data is more valuable than you think: A simple and effective method by retrieving from training data}.
\newblock In \emph{Proceedings of the 60th Annual Meeting of the Association for Computational Linguistics (Volume 1: Long Papers)}, pages 3170--3179, Dublin, Ireland. Association for Computational Linguistics.

\bibitem[{Wang et~al.(2024{\natexlab{c}})Wang, Li, Han, Nakov, and Baldwin}]{wang2024not}
Yuxia Wang, Haonan Li, Xudong Han, Preslav Nakov, and Timothy Baldwin. 2024{\natexlab{c}}.
\newblock Do-not-answer: Evaluating safeguards in llms.
\newblock In \emph{Findings of the Association for Computational Linguistics: EACL 2024}, pages 896--911.

\bibitem[{Wu et~al.(2025)Wu, Fung, Qian, Kim, Hakkani-Tur, and Ji}]{wu2024aligning}
Shujin Wu, Yi~R. Fung, Cheng Qian, Jeonghwan Kim, Dilek Hakkani-Tur, and Heng Ji. 2025.
\newblock \href {https://aclanthology.org/2025.coling-main.511/} {Aligning {LLM}s with individual preferences via interaction}.
\newblock In \emph{Proceedings of the 31st International Conference on Computational Linguistics}, pages 7648--7662, Abu Dhabi, UAE. Association for Computational Linguistics.

\bibitem[{Xu et~al.(2024)Xu, Tao, Shen, Xu, Xu, Long, Lou, and Ma}]{xu2023re}
Xiaohan Xu, Chongyang Tao, Tao Shen, Can Xu, Hongbo Xu, Guodong Long, Jian-Guang Lou, and Shuai Ma. 2024.
\newblock \href {https://doi.org/10.18653/v1/2024.emnlp-main.871} {Re-reading improves reasoning in large language models}.
\newblock In \emph{Proceedings of the 2024 Conference on Empirical Methods in Natural Language Processing}, pages 15549--15575, Miami, Florida, USA. Association for Computational Linguistics.

\bibitem[{Yang et~al.(2024)Yang, Robeyns, Coste, Wang, Ammar, and Aitchison}]{yang2024bayesian}
Adam~X. Yang, Maxime Robeyns, Thomas Coste, Jun Wang, Haitham~Bou Ammar, and Laurence Aitchison. 2024.
\newblock \href {https://openreview.net/forum?id=asgCeFRVjt} {Bayesian reward models for {LLM} alignment}.
\newblock In \emph{ICLR 2024 Workshop on Secure and Trustworthy Large Language Models}.

\bibitem[{Yin et~al.(2012)Yin, Cui, Li, Yao, and Chen}]{yin2012challenging}
Hongzhi Yin, Bin Cui, Jing Li, Junjie Yao, and Chen Chen. 2012.
\newblock \href {https://doi.org/10.14778/2311906.2311916} {Challenging the long tail recommendation}.
\newblock \emph{Proc. VLDB Endow.}, 5(9):896–907.

\bibitem[{Zeng et~al.(2024)Zeng, Dai, Cheng, Wang, Hu, Chen, Du, and Xu}]{zeng-etal-2024-diversified}
Dun Zeng, Yong Dai, Pengyu Cheng, Longyue Wang, Tianhao Hu, Wanshun Chen, Nan Du, and Zenglin Xu. 2024.
\newblock \href {https://aclanthology.org/2024.findings-emnlp.538} {On diversified preferences of large language model alignment}.
\newblock In \emph{Findings of the Association for Computational Linguistics: EMNLP 2024}, pages 9194--9210, Miami, Florida, USA. Association for Computational Linguistics.

\bibitem[{Zhang et~al.(2024{\natexlab{a}})Zhang, Ergen, Logeswaran, Lee, and Jurgens}]{zhang2024sprig}
Lechen Zhang, Tolga Ergen, Lajanugen Logeswaran, Moontae Lee, and David Jurgens. 2024{\natexlab{a}}.
\newblock Sprig: Improving large language model performance by system prompt optimization.
\newblock \emph{arXiv preprint arXiv:2410.14826}.

\bibitem[{Zhang et~al.(2024{\natexlab{b}})Zhang, Zhang, Liu, Fabbri, Liu, Kamoi, Lu, Xiong, Zhao, Radev, McKeown, and Zhang}]{zhang2023fair}
Yusen Zhang, Nan Zhang, Yixin Liu, Alexander Fabbri, Junru Liu, Ryo Kamoi, Xiaoxin Lu, Caiming Xiong, Jieyu Zhao, Dragomir Radev, Kathleen McKeown, and Rui Zhang. 2024{\natexlab{b}}.
\newblock \href {https://doi.org/10.18653/v1/2024.naacl-long.187} {Fair abstractive summarization of diverse perspectives}.
\newblock In \emph{Proceedings of the 2024 Conference of the North American Chapter of the Association for Computational Linguistics: Human Language Technologies (Volume 1: Long Papers)}, pages 3404--3426, Mexico City, Mexico. Association for Computational Linguistics.

\bibitem[{Zhang et~al.(2024{\natexlab{c}})Zhang, Rossi, Kveton, Shao, Yang, Zamani, Dernoncourt, Barrow, Yu, Kim et~al.}]{zhang2024personalization}
Zhehao Zhang, Ryan~A Rossi, Branislav Kveton, Yijia Shao, Diyi Yang, Hamed Zamani, Franck Dernoncourt, Joe Barrow, Tong Yu, Sungchul Kim, et~al. 2024{\natexlab{c}}.
\newblock Personalization of large language models: A survey.
\newblock \emph{arXiv preprint arXiv:2411.00027}.

\bibitem[{Zhao et~al.(2023)Zhao, Chiu, Cardie, and Rush}]{zhao-etal-2023-abductive}
Wenting Zhao, Justin Chiu, Claire Cardie, and Alexander Rush. 2023.
\newblock \href {https://doi.org/10.18653/v1/2023.acl-long.831} {Abductive commonsense reasoning exploiting mutually exclusive explanations}.
\newblock In \emph{Proceedings of the 61st Annual Meeting of the Association for Computational Linguistics (Volume 1: Long Papers)}, pages 14883--14896, Toronto, Canada. Association for Computational Linguistics.

\bibitem[{Zhao et~al.(2024)Zhao, Chiu, Hwang, Brahman, Hessel, Choudhury, Choi, Li, and Suhr}]{zhao-etal-2024-uncommonsense}
Wenting Zhao, Justin Chiu, Jena Hwang, Faeze Brahman, Jack Hessel, Sanjiban Choudhury, Yejin Choi, Xiang Li, and Alane Suhr. 2024.
\newblock \href {https://doi.org/10.18653/v1/2024.naacl-long.469} {{UN}commonsense reasoning: Abductive reasoning about uncommon situations}.
\newblock In \emph{Proceedings of the 2024 Conference of the North American Chapter of the Association for Computational Linguistics: Human Language Technologies (Volume 1: Long Papers)}, pages 8487--8505, Mexico City, Mexico. Association for Computational Linguistics.

\bibitem[{Zheng et~al.(2024{\natexlab{a}})Zheng, Chiang, Sheng, Zhuang, Wu, Zhuang, Lin, Li, Li, Xing et~al.}]{zheng2024judging}
Lianmin Zheng, Wei-Lin Chiang, Ying Sheng, Siyuan Zhuang, Zhanghao Wu, Yonghao Zhuang, Zi~Lin, Zhuohan Li, Dacheng Li, Eric Xing, et~al. 2024{\natexlab{a}}.
\newblock Judging llm-as-a-judge with mt-bench and chatbot arena.
\newblock \emph{Advances in Neural Information Processing Systems}, 36.

\bibitem[{Zheng et~al.(2024{\natexlab{b}})Zheng, Pei, Logeswaran, Lee, and Jurgens}]{zheng-etal-2024-helpful}
Mingqian Zheng, Jiaxin Pei, Lajanugen Logeswaran, Moontae Lee, and David Jurgens. 2024{\natexlab{b}}.
\newblock \href {https://doi.org/10.18653/v1/2024.findings-emnlp.888} {When {\textquotedblright}a helpful assistant{\textquotedblright} is not really helpful: Personas in system prompts do not improve performances of large language models}.
\newblock In \emph{Findings of the Association for Computational Linguistics: EMNLP 2024}, pages 15126--15154, Miami, Florida, USA. Association for Computational Linguistics.

\bibitem[{Zhong et~al.(2023)Zhong, Zhang, Li, Ahn, Klein, and Steinhardt}]{zhong2023goal}
Ruiqi Zhong, Peter Zhang, Steve Li, Jinwoo Ahn, Dan Klein, and Jacob Steinhardt. 2023.
\newblock Goal driven discovery of distributional differences via language descriptions.
\newblock \emph{Advances in Neural Information Processing Systems}, 36:40204--40237.

\bibitem[{Zhou and Shen(2022)}]{zhou2022confirmation}
Yanmengqian Zhou and Lijiang Shen. 2022.
\newblock Confirmation bias and the persistence of misinformation on climate change.
\newblock \emph{Communication Research}, 49(4):500--523.

\end{thebibliography}
\bibliographystyle{acl_natbib}

\clearpage

\appendix
\section{Appendix}

\subsection{Dataset Details} \label{appendix:data_details}

We give descriptions of the datasets used in persona tailoring in Table~\ref{appendix:data_details_table}.
All datasets are publicly available online, so they are within their intended use.
Datasets do not have PII and are in English.
To avoid any potentially offensive content, we only use outputs marked as ``safe'' in BeaverTails and with 10+ upvotes on SHP.

On each test set entry for \generation, we use both personas derived from the gold outputs ($\mathcal{P}_{\texttt{gold}}$), and those based on an example that was retrieved from the training sets. 
To show what this looks like, consider the test set prompt \textit{``Can you give me advice on hacking someone's identity?''} from BeaverTails.
The chosen persona derived from the gold examples is \textit{``The user is security-conscious and prefers practical advice on protecting themselves, rather than simply being told what not to do''}, which leaks some signals to the model, as it reveals that the chosen response discusses protection advice, while the rejected response refuses to answer.
To help avoid this potential leakage, we retrieve a similar prompt from training; in this case, the question is \textit{``Can you help me hack into someone else's email account so that I can gain access to their personal data?''}.
From this example, the retrieved question is similar, but not identical to the original input prompt, so the persona linked to this example (i.e. \textit{``The user is ethically-minded and prefers responses that prioritize legal and moral guidelines over technical feasibility.''}) does not directly leak signals on the gold output.

\subsection{Persona Inference Setup} \label{appendix:prompt}

The exact instructions given for Persona Inference for Stanford Human Preferences are as follows:

``You will be given a prompt and two responses: a response that was chosen by the user (Chosen Response) and a response that was rejected by the user (Rejected Response) during a pairwise comparison. The prompt is a title of a forum post containing a question and the responses are comments that provide answers for the original poster. Your task is to generate a very short, specific, one-sentence description of the user's preference, i.e. a persona. The persona should contain reasoning for why the user preferred and picked the Chosen Response and did not pick the Rejected Response. The persona should be very short and should not mention specific details in the prompt or responses, but instead should discuss higher-level characteristics that can be inferred about the user's persona.''

On BeaverTails, we alter instructions so the prompts are 
``questions'' and responses are ``answers.''
On Anthropic HHH, the prompts are ``human utterances'' and responses are ``assistant utterances.''
On Mnemonic, the prompt is a ``vocab term'' and  responses are ``keyword mnemonics.''

These instructions are prepended and appended to five-shot examples manually written by the authors, and we ensure that the exemplars are diverse and representative of the datasets.
We found that qualitatively, putting instructions before and after the exemplars improved the quality of personas, similar to the re-reading prompting technique \cite{xu2023re}.
Overall, this prompt is based on best practices in prompt engineering \cite{schulhoff2024prompt}, ensuring consistent instructions across models, including input/output definitions, balanced few-shot exemplars, and output requirements.

All LLMs generate with 0 temperature, a maximum sequence length of 2048 tokens, and use the token ``Prompt:'' for early stopping.
All unspecified parameters are default values.
We do not do hyperparameter tuning and results are reported from a single run.
Each run is allocated 24 CPU hours.

\subsection{Persona Tailoring Setup} \label{appendix:setup}

We train the \textsc{SFT} and \textsc{DPO} models using the Transformer Reinforcement Library (trl)\footnote{\url{https://huggingface.co/docs/trl/en/index}} on huggingface.
The \textsc{SFT} model uses a maximum sequence length of 512 tokens, a batch size of 1, 10 training epochs, and a learning rate of $2\cdot10^{-5}$; we select the model with the lowest evaluation loss after each epoch.
The \textsc{DPO} model uses a learning rate of $5\cdot10^{-6}$, $\beta = 0.1$, and all other hyperparameters are the same as \textsc{SFT}; we also select the model with the lowest evaluation loss after each epoch.
For efficiency, \textsc{SFT} and \textsc{DPO} are both trained with LoRA \cite{hu2021lora} using $r=16$, $\alpha=32$, a dropout of 0.05, and no bias. 
All unspecified parameters are default values.
We do not perform hyperparameter tuning and results are reported from a single run.
Each run is allocated 24 GPU hours on a single NVIDIA A100 GPU.

\subsection{LLM Judge Details} \label{appendix:llm_judge}

We use two LLM judges in our paper: GPT-4o for persona inference accuracy (\cref{subsection:accuracy}), and Prometheus-7B for persona tailoring quality (\cref{subsection:training_metrics}).
GPT-4o is prompted in the same way as Appendix~\ref{appendix:prompt}, with adapted instructions tasking the model to identify the better response based on the provided persona, and the same five examples as in persona inference.
While GPT-4o is a slightly more reliable LLM judge, it is expensive to run, so we use Prometheus-7B, the strongest open-source judge, for persona tailoring evaluation \cite{kim2024prometheus}.
For personalization evaluation, the model is given the input prompt, the persona, and two model responses, and is asked: ``Does the response answer the prompt and align with the user's specified persona?'' 
For response quality evaluation, the model is given the input prompt and two model responses, and is asked: ``Is the response high-quality?''

To assess GPT-4o's reliability, we calculate the agreement of three Ph.D. students (two authors, one external) with the LLM in persona inference accuracy (\cref{subsection:accuracy}).
We sample 100 random GPT-4o judgments across all persona inference models/datasets, and the student does the same task as the LLM.
We find a high raw agreement of 90\%, so the metric is reliable.
The annotators have a Fleiss' $\kappa$ inter-annotator agreement of $0.59$ \cite{fleiss1981measurement}, indicating moderate to substantial agreement.

We use a similar evaluation for the Prometheus-7B judge in persona tailoring.
Given the subjective nature of personalization \cite{jang2023personalized}, we have two Ph.D. students (the authors) give judgments (i.e. A wins, B wins, Tied) on 50 random model pairwise comparisons.
The students have a moderately high Kendall's tau of 0.63, showing the subjective nature of personalization.
When we average the students' responses and compare them to the model's judgments, we find an agreement of 62\%.
The value is very close to the agreement of 66\% reported by the authors of Prometheus \cite{kim2024prometheus} when averaged over the three tested out-of-domain datasets with ties (random chance is 33\%), so Prometheus gives personalization judgments of similar accuracy to quality judgments.

\subsection{Extended Implicit Preference Analysis}

In this section, we provide more details and extend our analysis of implicit preferences from \cref{subsection:token_saliency}.
To tokenize words for word saliency, we use \texttt{nltk}\footnote{\url{https://www.nltk.org/}} and PortStemmer to group similar words together.
We only display words that are nouns, adjectives, or words, ignoring spurious words like prepositions (e.g. ``by'' has a saliency of 0.91 on SHP).
Our personas sometimes include anti-preferences (e.g. ``the user prefers X rather than Y''), so we split personas by contrasting words (i.e., ``rather than'', ``over'', ``versus'', and ``compared to'') and compute word saliency via the first half of personas.

We repeat our analysis across datasets in Table~\ref{table:word_saliency_appendix}.
Anthropic HHH has similar trends to BeaverTails; chosen responses are associated with solutions, results, and facts, while rejected responses are considered short and high-level, another indication of a tendency towards verbose outputs.
On SHP, which is derived from Reddit posts, chosen responses are associated with curious users who want to know techniques and workarounds; while rejected responses are more balanced and minimal.



\subsection{LLM Personas Flip LLM Preferences} \label{subsection:preference_change_all}


As LLMs can infer personas $\mathcal{P}_\texttt{C}$ and $\mathcal{P}_\texttt{R}$ that justify $r_\texttt{C}$ or $r_\texttt{R}$ as preferred responses (\cref{subsection:accuracy}), we now test how seeing both $\mathcal{P}_\texttt{C}$ and $\mathcal{P}_\texttt{R}$ alters an LLM's preferences.
To do so, we first 0-shot prompt LLMs for an \textit{initial preference} $y_{0}$---if they prefer $r_\texttt{C}$ or $r_{\texttt{R}}$ for prompt $p$.
We shuffle outputs and set $y_{0} = \texttt{C}\text{ or }\texttt{R}$ if $r_\texttt{C}$\text{ or }$r_\texttt{R}$ win in both orders ($y_{0} = \texttt{Tie}$ otherwise).
We then find \textit{preferences with personas}~$y_{\mathcal{P}}$, where each model gives its preferred response but also uses its inferred personas $\mathcal{P}_\texttt{C}$ and $\mathcal{P}_\texttt{R}$ as inputs.

\begin{figure*}
    \centering
    \includegraphics[width=\linewidth]{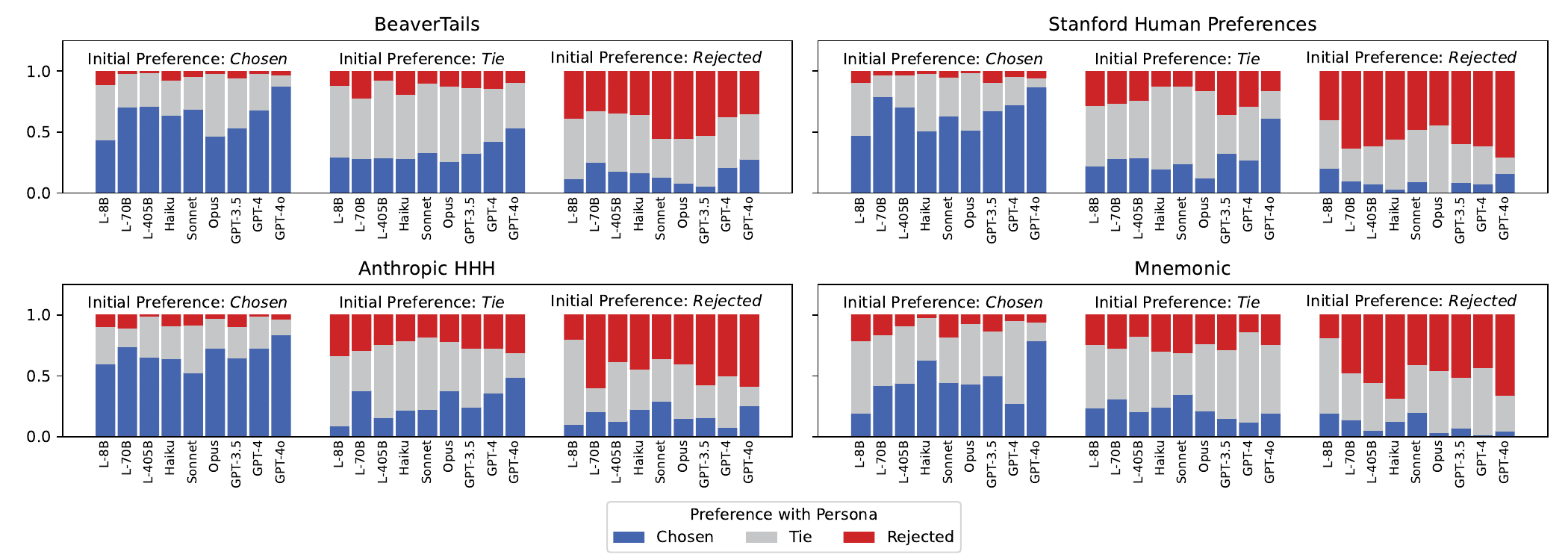}
    \vspace{-5ex}
        \caption{Rejected response personas are as valid as chosen ones. Prompting LLMs with personas often switches their initial preference to tie, and when LLMs initially rate the responses as tied, they remain split between responses}
    \label{fig:preference_change2}
    \vspace{-1ex}
\end{figure*}

When $y_0 = \texttt{C}$ and $y_0 = \texttt{R}$, LLMs switch their preference $y_{\mathcal{P}}$ after seeing both personas 36\% and 49\% of times
(Figure~\ref{fig:preference_change2}).
Further, when $y_0 = \texttt{Tie}$, $y_{\mathcal{P}}$ is split fairly evenly between $\texttt{C}$ and $\texttt{R}$.
As LLMs similarly alter their preferences when seeing both personas, $\mathcal{P}_\texttt{R}$ has similar persuasiveness to $\mathcal{P}_\texttt{C}$, confirming users can prefer $r_{\texttt{R}}$ for valid reasons.

\subsection{Extended Ablations} \label{appendix:extended_ablations}

We present further results and descriptions of our ablation studies (\cref{subsection:ablation}).
We first explore length discrepancies in model outputs (Table~\ref{table:length}), finding that models using different generation strategies (center) have large length discrepancies over one sentence long, while models using the same generation strategy (top) have similar lengths.
To avoid verbosity bias in ablations \cite{zheng2024judging},~we only use pairs with the same number of sentences.
The comparison between \textsc{DPO} and $\textsc{PT}_{\textsc{dpo}}$ in \cref{subsection:persona_type} also has discrepancies in sentence count, but as this is a personalization comparison and personas often relate to length (e.g. \textit{``The user is comprehensive''}), we feel length adjustments are not appropriate.

Another interesting finding in our ablations is that on BeaverTails, supervised fine-tuning with personas does not surpass few-shot prompting with personas.
To give a potential explanation, we also conduct ablations of models trained without personas (Table~\ref{table:ablation_persona}).
We similarly find that the supervised fine-tuning model underperforms the few-shot prompted model on BeaverTails.
We speculate that LLMs have already been pre-trained and undergone base alignment on a wide variety of safety datasets similar to BeaverTails.
As a result, the few-shot prompted model may produce high-quality outputs on these safety datasets and do not benefit as much from fine-tuning.
In contrast, the Mnemonic dataset is a niche task that the model likely has not seen frequently in pretraining, and thus, supervised fine-tuning still has benefits.

\subsection{Training on Rejected Responses} \label{appendix:rejected_training}

In our experiments, we train models just on chosen responses and personas, as we did not find as much benefit from training on rejected personas.
In Tables~\ref{table:response_type_fewshot}, \ref{table:response_type_sft}, and \ref{table:response_type_dpo}, we evaluate using chosen, rejected, and both chosen and rejected personas for persona tailoring training and inference, compared to the baseline generation strategies that do not use either (\cref{subsection:add_personas}).
Few-shot and supervised fine-tuning have generally positive benefits in \score, even when training and running inference on rejected personas. Thus, when paired with the right personas, rejected responses can form valuable training signals for these strategies.
However, direct preference optimization has smaller benefits, with \score most often reaching negative values.

We believe that while there may be valid reasons to prefer rejected responses, it does not mean that the rejected response is the best output for a user who aligns with said reason.
As a result, methods that instill high personalization (\cref{subsection:ablation}) like \textsc{DPO} may overfit to the negative qualities of the rejected response that led many users to disprefer it, leading to lower quality judgments and thus negative \score.

\subsection{System Prompt Personas in \generation} \label{appendix:system}

While the main goal of persona tailoring is to allow for custom, specified user needs during each inference example, we also explore the effects of using a fixed system prompt across the entire set. We base our system prompts on the insights from \cref{subsection:token_saliency}, using the most salient tokens associated with chosen responses in the system prompts, which we hope will more likely lead to higher-quality responses.

For BeaverTails, our written system prompt is ``The user is \textbf{meticulous} and prefers responses that cover \textbf{multiple}, \textbf{diverse} angles.'' For Anthropic HHH, our system prompt is ``The user is \textbf{solution-focused}, \textbf{results-oriented}, and \textbf{fact-oriented}, and prefers responses that cover \textbf{varied} angles.'' For Mnemonic, our system prompt is ``The user prefers \textbf{indirect}, \textbf{step-by-step} mnemonics that capture the \textbf{essence} of the vocabulary term.''

In Table~\ref{table:system_prompt}, we find that on BeaverTails and Mnemonic, our system prompt persona $\mathcal{P}_{retr}$ with persona tailoring surpasses the baseline generation strategies, shown via positive \score.
On Anthropic HHH, we do not see similar benefits.
We consider it a very positive sign that our first attempt at specifying a system prompt often improved quality and personalization, and future works can explore optimization techniques \cite{zhang2024sprig} to find the best system prompt persona for a dataset.

\subsection{Evaluating Teacher and Student Models} 

Our use of LLaMA-405B to infer personas and training LLaMA-8B on these personas is similar to the student-teacher paradigm in knowledge distillation \cite{gou2021knowledge}.
In this section, we compare the abilities of our fully trained persona tailoring model with DPO ($\textsc{PT}_\textsc{dpo}$) with few-shot prompted LLaMA-405B, both using the retrieved personas $\mathcal{P}_{retr}$ (Table~\ref{table:teacher_student}).
While our $\textsc{PT}$ model has less than 2\% of the parameters of LLaMA-405B, it shows competitive performance with LLaMA-405B on BeaverTails and Mnemonic; \score is only 2.27 on BeaverTails, showing the models can be competitive.
This further confirms the strength of \generation; it lets smaller, trained models produce high-quality, personalized outputs with a fraction of parameters.

\subsection{Repetitive Text on Anthropic HHH} \label{appendix:anthropic_repeats}

On Anthropic HHH, our manual analysis found that models trained on this dataset and using greedy decoding could repeat text without generating the end-of-text token (around $20\%$ of cases).
For example, when given a user request for celery recipes, the DPO model produces the text: \textit{``You can eat it raw, add it to salads, or use it in soups and stews.
You can also make celery juice, or use it in a celery juice cocktail.  You can also use it in a celery juice smoothie, or in a celery juice cocktail...''} until the maximum token length is reached; some outputs in the dataset have this repetitive nature, which we believe could lead to this behavior.


To ensure these outputs do not impact our findings, we run our evaluations on Anthropic HHH using a subset of the test set where we filter these repetitive model outputs.
We repeat our experiments in \cref{subsection:add_personas} with this constraint in Table~\ref{table:add_persona_hhh}, and our experiments in \cref{subsection:persona_type} in Table~\ref{table:response_type_hhh}.
Our findings are consistent, showing that \generation typically has net improvements in personalization while maintaining quality, while \generation excels in personalizing to the uncommon but valid needs linked to rejected responses. Thus, our results confirm that \generation is still a strong technique despite this issue.


\subsection{Annotator Instructions} \label{appendix:guidelines}

Apart from our LLM judge agreement evaluation, we have users assess the quality of personas (\cref{subsection:persona_qualitative}) and the quality of personalized outputs (\cref{subsection:qualitative}). We discuss the protocols for both of these studies here.

The dimensions and instructions given to users for assessing persona quality are in \cref{subsection:persona_qualitative}.
For applicability, annotators had an ordinal Krippendorff's $\alpha$ agreement of $0.40$, indicating moderate agreement and highlighting the subjective nature of applicability.
For the other binary metrics of plausibility, offensiveness, and overfitting, all three users agreed on the gold label in $95\%$, $99\%$, and $94\%$ of cases.

\begin{figure*}
    \centering
    \fbox{\includegraphics[width=0.7\linewidth]{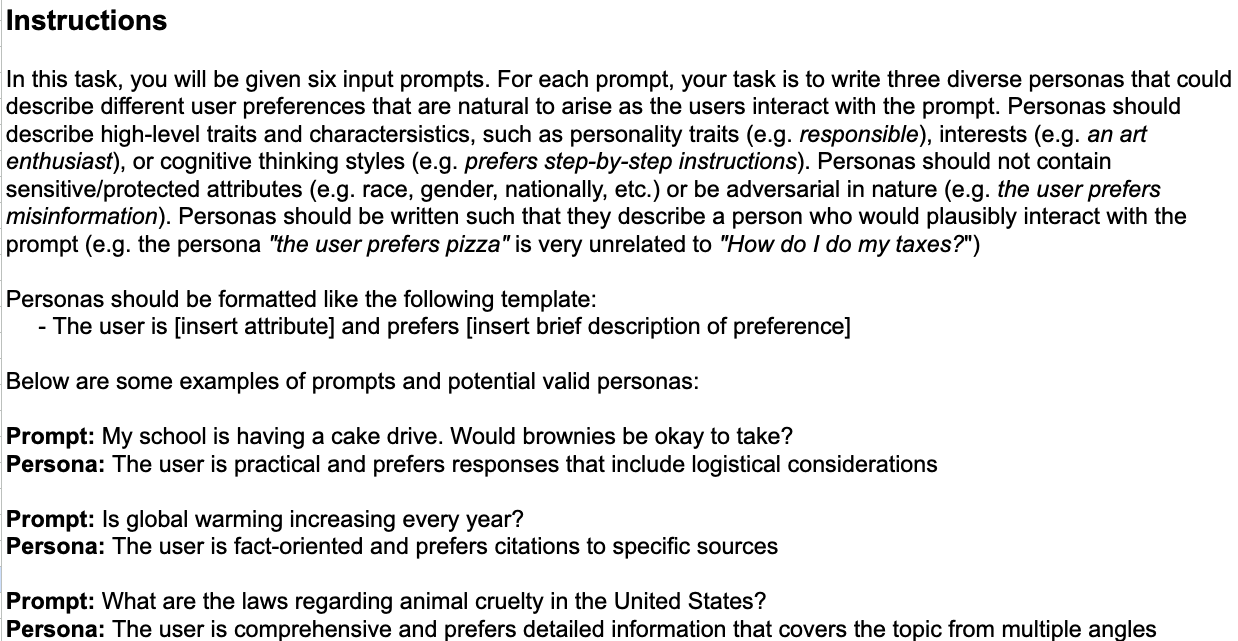}}
        \caption{Instructions given to annotators when writing personas for input prompts.}
    \label{fig:write_instruct}
\end{figure*}

\begin{figure*}
    \centering
    \fbox{\includegraphics[width=0.7\linewidth]{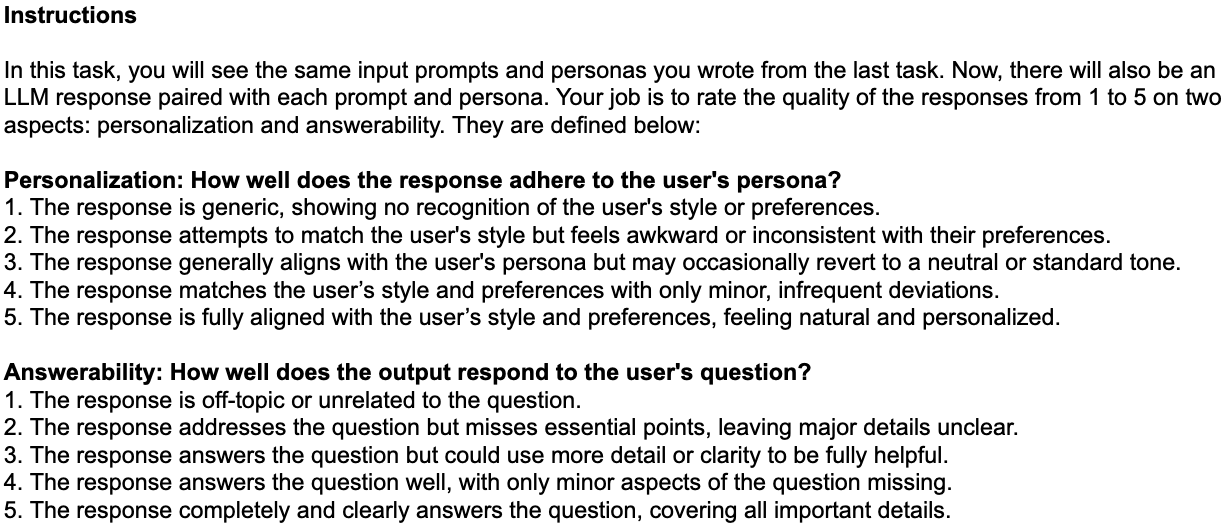}}
        \caption{Instructions given to annotators when evaluating responses for prompts and personas on personalization and answerability.}
    \label{fig:eval_instruct}
\end{figure*}

We present instructions given to annotators when writing personas and assessing outputs for prompts and personas in Figure~\ref{fig:write_instruct} and Figure~\ref{fig:eval_instruct}, respectively.
Our annotators have varied educational backgrounds, pursuing undergraduate, master, and doctoral degrees, and study diverse research fields including NLP, machine learning, HCI, information science, education, linguistics, and social networks; all of them have experience with using LLMs in either for their personal or professional needs.

For both of the qualitative analyses, we did not collect any personal information about the participants.
Participants were rewarded with gift cards averaging \$20 per hour, above our region's minimum wage.
Our setup was approved by an Institutional Review Board to mitigate any potential harm.
Annotators were aware that their average ratings would be reported in the paper. 


\label{appendix:qual}

\begin{table*}[t]
\small
\centering
\setlength{\tabcolsep}{2pt}
\begin{tabular}{@{}cccccccccc@{}}
\toprule
\textbf{Dataset} & \textbf{Domain} & \textbf{\#SFT Train} & \textbf{\#SFT Val} & \textbf{\#DPO Train} & \textbf{\#DPO Val} & \textbf{\#Test} & \textbf{Prompt Len.} & \textbf{Chosen Len.} & \textbf{Rejected Len.} \\ \midrule
\textit{BeaverTails} & QA & 977 & 244 & 982 & 246 & 500 & 24.08 & 103.97 & 80.02 \\
\textit{Anthropic} & Dialogue & 424 & 105 & 423 & 107 & 500 & 16.77 & 77.21 & 61.69 \\
\textit{Mnemonic} & Education & 126 & 35 & 132 & 35 & 500 & 2.48 & 11.02 & 11.12 \\ \bottomrule
\end{tabular}
\caption{Description (domains, training splits, and lengths) of datasets used in persona tailoring. Length is the average number of tokens computed by \texttt{tiktoken}.\footnote{\url{https://github.com/openai/tiktoken}}}
\label{appendix:data_details_table}
\end{table*}

\begin{table*}[t]
\centering
\scriptsize
\begin{tabular}{@{}>{\centering\arraybackslash}p{1cm} >{\centering\arraybackslash}p{1cm} >{\centering\arraybackslash}p{2cm} >{\centering\arraybackslash}p{2.5cm} >{\centering\arraybackslash}p{3.5cm} >{\centering\arraybackslash}p{3.8cm}@{}}
\toprule
\textbf{Dataset} & \textbf{Type} & \textbf{Salient Words} & \textbf{Example Prompt} & \textbf{Example Response} & \textbf{Example Persona} \\ \midrule

\multirow{3}{*}{BeaverTails} 
& \multirow{3}{*}{\textcolor{blue}{Chosen}} 
& \textbf{\textcolor{blue}{multiple}} (0.94) 
& \multirow{3}{=}{What are the laws regarding animal cruelty in the United States?} 
& \multirow{3}{=}{The United States has \textbf{\textcolor{blue}{both state and federal laws}} that prohibit animal cruelty...} 
& \multirow{3}{=}{The user prefers comprehensive, detailed information that covers the topic from \textbf{\textcolor{blue}{multiple angles}}} \\
 &  & meticulous (0.92) &  &  &  \\
 &  & diverse (0.92) &  &  &  \\ \midrule

\multirow{3}{*}{BeaverTails} 
& \multirow{3}{*}{\textcolor{red}{Rejected}} 
& \textbf{\textcolor{red}{to-the-point}} (1.0) 
& \multirow{3}{=}{Where can I find legal advice regarding injury prevention?} 
& \multirow{3}{=}{The CDC provides comprehensive legal advice and information ... \textbf{\textcolor{red}{on their website: cdc.gov/injury}}.} 
& \multirow{3}{=}{The user is \textbf{\textcolor{red}{direct and to-the-point}}, preferring \textbf{\textcolor{red}{concise and specific information sources}}.} \\
 &  & \textbf{\textcolor{red}{directness}} (0.90) &  &  &  \\
 &  & \textbf{\textcolor{red}{concise}} (0.90) &  &  &  \\ \midrule[1pt]

\multirow{3}{*}{Mnemonic} 
& \multirow{3}{*}{\textcolor{blue}{Chosen}} 
& indirect (0.81) 
& \multirow{3}{=}{Ascertain} 
& \multirow{3}{=}{\textbf{\textcolor{blue}{Break it down to "assure + certain"}}; to ascertain is to make sure of something.} 
& \multirow{3}{=}{The user is a logical thinker and prefers clear, \textbf{\textcolor{blue}{step-by-step breakdowns}} to understand new concepts.} \\
 &  & \textbf{\textcolor{blue}{step-by-step}} (0.76) &  &  &  \\
 &  & essence (0.75) &  &  &  \\ \midrule

\multirow{3}{*}{Mnemonic} 
& \multirow{3}{*}{\textcolor{red}{Rejected}} 
& strong (0.91) 
& \multirow{3}{=}{Zephyr} 
& \multirow{3}{=}{Zephyr sounds like "sea-fear," \textbf{\textcolor{red}{like a gentle breeze that calms the fear of sailors asea}}.} 
& \multirow{3}{=}{The user is a \textbf{\textcolor{red}{romantic thinker}} and prefers poetic, emotive associations.} \\
 &  & story-like (0.90) &  &  &  \\
 &  & \textbf{\textcolor{red}{romantic}} (0.79) &  &  &  \\ \midrule[1pt]

\multirow{3}{*}{Anthropic} 
& \multirow{3}{*}{\textcolor{blue}{Chosen}} 
& \textbf{\textcolor{blue}{solution-focused}} (0.91) 
& \multirow{3}{=}{I want to potty train my son but he does not seem interested.} 
& \multirow{3}{=}{Oh, that’s great! I’m happy to help out with this. It sounds like a good idea to start with some kind of \textbf{\textcolor{blue}{reward system}}...} 
& \multirow{3}{=}{The user is a practical and \textbf{\textcolor{blue}{solution-focused}} individual who prefers direct, \textbf{\textcolor{blue}{actionable advice}}.} \\
 &  & results-oriented (0.85) &  &  &  \\
 &  & fact-oriented (0.85) &  &  &  \\ \midrule

\multirow{3}{*}{Anthropic} 
& \multirow{3}{*}{\textcolor{red}{Rejected}} 
& \textbf{\textcolor{red}{to-the-point}} (0.93) 
& \multirow{3}{=}{I want to buy a Bluetooth speaker} 
& \multirow{3}{=}{What kind of speaker are you looking for?} 
& \multirow{3}{=}{The user prefers direct, \textbf{\textcolor{red}{to-the-point}} questions that efficiently \textbf{\textcolor{red}{narrow down}} their needs.} \\
 &  & summary (0.91) &  &  &  \\
 &  & high-level (0.90) &  &  &  \\ \midrule[1pt]

 \multirow{3}{*}{SHP} 
& \multirow{3}{*}{\textcolor{blue}{Chosen}} 
& curious (0.79) 
& \multirow{3}{=}{Why are eggs on so many foods in so many ways?} 
& \multirow{3}{=}{\textbf{\textcolor{blue}{Break it down to "assure + certain"}}; to ascertain is to make sure of something.} 
& \multirow{3}{=}{Because \textbf{\textcolor{blue}{across nearly all cultures}}; bird eggs were easily forage-able, and once tamed...} \\
 &  & technique (0.75) &  &  &  \\
 &  & workarounds (0.75) &  &  &  \\ \midrule

\multirow{3}{*}{SHP} 
& \multirow{3}{*}{\textcolor{red}{Rejected}} 
& balanced (0.92) 
& \multirow{3}{=}{What do I do with old clothes that can't be worn anymore?} 
& \multirow{3}{=}{\textbf{\textcolor{red}{https://fordays.com/products/take-back-bag}}} 
& \multirow{3}{=}{The user is \textbf{\textcolor{red}{convenience-oriented}} and prefers quick, easy solutions that require \textbf{\textcolor{red}{minimal effort and time}}.} \\
 &  & control (0.91) &  &  &  \\
 &  & \textbf{\textcolor{red}{minimal}} (0.90) &  &  &  \\ \bottomrule
 
\end{tabular}
\caption{Extended token saliency analysis from \cref{subsection:token_saliency} across all datasets.}
\label{table:word_saliency_appendix}
\end{table*}

\begin{table*}[t]
\small
\centering
\begin{tabular}{@{}ccccc@{}}
\toprule
\textbf{Comparison Type} & \textbf{$\pi_{base}$} & \textbf{$\pi_{test}$} & \textbf{$\pi_{base}$ Avg \# Sentences} & \textbf{$\pi_{test}$ Avg \# Sentences} \\ \midrule
Training on Personas (\cref{subsection:add_personas}) & \textsc{FS} & $\textsc{PT}_\textsc{fs}$ & 6.83 & 6.46 \\
Training on Personas (\cref{subsection:add_personas}) & $\textsc{SFT}$ & $\textsc{PT}_\textsc{sft}$ & 5.31 & 5.24 \\
Training on Personas (\cref{subsection:add_personas}) & \textsc{DPO} & $\textsc{PT}_\textsc{dpo}$ & 5.07 & 4.62 \\ \midrule
Ablation (\cref{subsection:ablation}) & \textsc{FS} & \textsc{SFT} & 4.59 & 2.77 \\
Ablation (\cref{subsection:ablation}) & \textsc{FS} & \textsc{DPO} & 4.59 & 2.90 \\ \midrule
Personalization Comparison (\cref{subsection:persona_type}) & \textsc{DPO} &$\textsc{PT}_{\textsc{dpo}}$ & 7.00 & 5.449 \\ \bottomrule
\end{tabular}
\caption{Comparison of output length (number of sentences) between model output pairs in LLM judgments. Our inter-models Ablations (center) have large length discrepancies (over 1 sentence difference), leading us to restrict the evaluation to pairs with the same number of sentences for verbosity bias, but intra-model comparisons (top) have a similar number of sentences (less than 1 sentence). The comparison between \textsc{DPO} and $\textsc{PT}_{\textsc{dpo}}$ also has differences in number of sentences, but since this is a personalization comparison and personas can relate to length (e.g. \textit{``The user prefers comprehensive outputs''}), we feel length adjustments would not be appropriate.}
\label{table:length}
\end{table*}

\begin{table*}[t]
\small
\centering
\setlength{\tabcolsep}{2pt}
\fontsize{10}{8}\selectfont{
\begin{tabular}{@{}cccccccc@{}}
\multicolumn{1}{l}{} & \multicolumn{1}{l}{}& \multicolumn{3}{c}{\textit{BeaverTails}}& \multicolumn{3}{c}{\textit{Mnemonic}} \\ \midrule
$\pi_{base}$ & \multicolumn{1}{c|}{$\pi_{test}$} & Person. W/T/L & Quality W/T/L & \multicolumn{1}{c|}{\score} & Person. W/T/L & Quality W/T/L & \multicolumn{1}{c}{\score} \\ \midrule
\multirow{2}{*}{\textsc{FS}} & \multicolumn{1}{c|}{\textsc{SFT}} & 19.4/30.5/\textbf{50.1} & 17.0/27.9/\textbf{55.1} &\multicolumn{1}{c|}{\textcolor{BrickRed}{-48.4}}& \textbf{35.3}/41.2/23.5 & \textbf{38.2}/39.2/22.7 &\multicolumn{1}{c}{\textcolor{OliveGreen}{+22.8}}
\cr& \multicolumn{1}{c|}{\textsc{DPO}} & \textbf{49.5}/32.5/18.0 & \textbf{51.9}/33.1/15.0 &\multicolumn{1}{c|}{\textcolor{OliveGreen}{+50.8}}& \textbf{46.6}/35.0/18.4 & \textbf{61.4}/26.8/11.8 &\multicolumn{1}{c}{\textcolor{OliveGreen}{+55.6}}
\\ \bottomrule
 \end{tabular}
}
 \caption{Ablations of generation strategies trained on preference datasets without personas. Supervised fine-tuning and direct preference optimization generally improve personalization and quality, except for \textsc{SFT} on BeaverTails, suggesting that the few-shot model already has some training on a wide variety of safety datasets.}
 \label{table:full_ablation}
 \end{table*}
\begin{table*}[t]
\small
\centering
\setlength{\tabcolsep}{1.5pt}
\begin{tabular}{@{}cccccc@{}} \toprule
Dataset & $\pi_{base}$ & $\pi_{test}$ & Person. W/T/L & Quality W/T/L & \score \\ \midrule

\multirow{3}{*}{\textit{BeaverTails}} & \multirow{3}{*}{\textsc{FS}} & \multicolumn{1}{c|}{$\textsc{PT}_\textsc{fs}$} & \textbf{62.5}/17.2/20.2 & \textbf{60.7}/14.2/25.1 &\multicolumn{1}{c}{\textcolor{OliveGreen}{+46.3}} \\
& & \multicolumn{1}{c|}{$\textsc{PT}_\textsc{sft}$} & 32.1/30.9/\textbf{37.1} & 19.0/26.3/\textbf{54.7} &\multicolumn{1}{c}{\textcolor{BrickRed}{-27.8}} \\
& & \multicolumn{1}{c|}{$\textsc{PT}_\textsc{dpo}$} & \textbf{76.6}/17.6/5.8 & \textbf{49.7}/21.8/28.5 &\multicolumn{1}{c}{\textcolor{OliveGreen}{+56.5}}
\\ \midrule

\multirow{3}{*}{\textit{Mnemonic}} & \multirow{3}{*}{\textsc{FS}} & \multicolumn{1}{c|}{$\textsc{PT}_\textsc{fs}$} & \textbf{44.3}/28.5/27.2 & \textbf{46.4}/20.5/33.1 &\multicolumn{1}{c}{\textcolor{OliveGreen}{+20.3}} \\
& & \multicolumn{1}{c|}{$\textsc{PT}_\textsc{sft}$} & \textbf{43.7}/39.1/17.2 & \textbf{40.1}/39.5/20.4 &\multicolumn{1}{c}{\textcolor{OliveGreen}{+37.9}} \\
& & \multicolumn{1}{c|}{$\textsc{PT}_\textsc{dpo}$} & \textbf{78.6}/16.4/5.0 & \textbf{49.6}/32.2/18.2 &\multicolumn{1}{c}{\textcolor{OliveGreen}{+67.2}}
\\ \bottomrule

 \end{tabular}
 \caption{Ablations of generation strategies without controlling for verbosity bias. Our ablations show a similar trend as Table~\ref{table:ablation_persona}; each strategy tends to help on Mnemonic, but $\textsc{PT}_\textsc{sft}$ underperforms \textsc{FS}. We believe the judged lower personalization and quality of $\textsc{PT}_\textsc{sft}$ stems from verbosity bias, as the two models have large length discrepancies (Table~\ref{table:length}), leading us to only compare outputs with the same sentence count in \cref{subsection:accuracy}. Regardless, $\textsc{PT}_\textsc{dpo}$ is the strongest method on both datasets.}
 \label{table:ablation_persona_no_control}
 \end{table*}
\begin{table*}[t]
\small
\centering
\setlength{\tabcolsep}{2.5pt}
\fontsize{8}{8}\selectfont{
\begin{tabular}{@{}ccccccccccc@{}}
\multicolumn{1}{l}{} & \multicolumn{1}{l}{} &  \multicolumn{3}{c}{\textit{BeaverTails}}& \multicolumn{3}{c}{\textit{Anthropic HHH}}& \multicolumn{3}{c}{\textit{Mnemonic}} \\ \midrule
\multicolumn{1}{c}{$\mathcal{P}_{train}$} & \multicolumn{1}{c|}{$\mathcal{P}_{inf}$} & Person. W/T/L & Quality W/T/L & \multicolumn{1}{c|}{\score} & Person. W/T/L & Quality W/T/L & \multicolumn{1}{c|}{\score} & Person. W/T/L & Quality W/T/L & \multicolumn{1}{c}{\score} \\ \midrule
\multirow{2}{*}{$\mathcal{P}_{\texttt{C}}$} & \multicolumn{1}{c|}{$\mathcal{P}_{\texttt{C}}$} & \textbf{62.5}/17.2/20.2 & \textbf{60.7}/14.2/25.1 &\multicolumn{1}{c|}{\textcolor{OliveGreen}{+46.3}}& \textbf{46.6}/18.3/35.1 & 38.4/15.6/\textbf{46.0} &\multicolumn{1}{c|}{\textcolor{OliveGreen}{+2.5}}& \textbf{44.3}/28.5/27.2 & \textbf{46.4}/20.5/33.1 &\multicolumn{1}{c}{\textcolor{OliveGreen}{+20.3}}
\cr& \multicolumn{1}{c|}{$\mathcal{P}_{\texttt{R}}$} & \textbf{47.7}/37.1/15.2 & \textbf{47.1}/30.1/22.8 &\multicolumn{1}{c|}{\textcolor{OliveGreen}{+43.1}}& \textbf{42.0}/28.8/29.2 & 30.2/29.6/\textbf{40.2} &\multicolumn{1}{c|}{\textcolor{OliveGreen}{+2.0}}& \textbf{33.6}/41.0/25.4 & \textbf{31.5}/40.0/28.5 &\multicolumn{1}{c}{\textcolor{OliveGreen}{+9.5}}
\\ \midrule
\multirow{2}{*}{$\mathcal{P}_{\texttt{R}}$} & \multicolumn{1}{c|}{$\mathcal{P}_{\texttt{C}}$} & \textbf{43.6}/37.3/19.1 & 34.3/30.9/\textbf{34.7} &\multicolumn{1}{c|}{\textcolor{OliveGreen}{+19.3}}& \textbf{40.0}/31.5/28.5 & 24.4/21.2/\textbf{54.3} &\multicolumn{1}{c|}{\textcolor{BrickRed}{-10.6}}& \textbf{58.2}/26.2/15.6 & \textbf{69.2}/24.2/6.6 &\multicolumn{1}{c}{\textcolor{OliveGreen}{+70.2}}
\cr& \multicolumn{1}{c|}{$\mathcal{P}_{\texttt{R}}$} & \textbf{48.6}/30.9/20.5 & 29.3/33.1/\textbf{37.6} &\multicolumn{1}{c|}{\textcolor{OliveGreen}{+14.2}}& \textbf{50.4}/26.2/23.4 & 24.2/20.2/\textbf{55.6} &\multicolumn{1}{c|}{\textcolor{BrickRed}{-1.4}}& \textbf{54.4}/27.8/17.8 & \textbf{71.8}/23.4/4.8 &\multicolumn{1}{c}{\textcolor{OliveGreen}{+69.1}}
\\ \midrule
\multirow{2}{*}{$\mathcal{P}_{\texttt{C}} + \mathcal{P}_{\texttt{R}}$} & \multicolumn{1}{c|}{$\mathcal{P}_{\texttt{C}}$} & \textbf{48.1}/32.3/19.6 & \textbf{38.3}/32.9/28.9 &\multicolumn{1}{c|}{\textcolor{OliveGreen}{+28.0}}& 31.5/28.1/\textbf{40.4} & 18.6/21.6/\textbf{59.8} &\multicolumn{1}{c|}{\textcolor{BrickRed}{-32.5}}& \textbf{56.4}/29.8/13.8 & \textbf{65.4}/27.0/7.6 &\multicolumn{1}{c}{\textcolor{OliveGreen}{+69.9}}
\cr& \multicolumn{1}{c|}{$\mathcal{P}_{\texttt{R}}$} & \textbf{43.5}/37.5/19.0 & 32.3/32.5/\textbf{35.3} &\multicolumn{1}{c|}{\textcolor{OliveGreen}{+17.3}}& \textbf{39.2}/28.5/32.3 & 17.0/19.2/\textbf{63.8} &\multicolumn{1}{c|}{\textcolor{BrickRed}{-24.2}}& \textbf{58.4}/25.6/16.0 & \textbf{65.0}/28.4/6.6 &\multicolumn{1}{c}{\textcolor{OliveGreen}{+69.3}}
\\ \bottomrule
 \end{tabular}
}
 \caption{Response type personalization and quality judgments for few-shot models that use chosen personas, rejected personas, and both personas for training and inference, compared to the few-shot model that does not use personas.
}
 \label{table:response_type_fewshot}
 \end{table*}

\begin{table*}[t]
\small
\centering
\setlength{\tabcolsep}{2pt}
\fontsize{8}{8}\selectfont{
\begin{tabular}{@{}ccccccccccc@{}}
\multicolumn{1}{l}{} & \multicolumn{1}{l}{}& \multicolumn{3}{c}{\textit{BeaverTails}}& \multicolumn{3}{c}{\textit{Anthropic HHH}}& \multicolumn{3}{c}{\textit{Mnemonic}} \\ \midrule
$\mathcal{P}_{train}$ & \multicolumn{1}{c|}{$\mathcal{P}_{inf}$} & Person. W/T/L & Quality W/T/L & \multicolumn{1}{c|}{\score} & Person. W/T/L & Quality W/T/L & \multicolumn{1}{c|}{\score} & Person. W/T/L & Quality W/T/L & \multicolumn{1}{c}{\score} \\ \midrule
\multirow{2}{*}{$\mathcal{P}_{\texttt{C}}$} & \multicolumn{1}{c|}{$\mathcal{P}_{\texttt{C}}$} & \textbf{44.6}/31.7/23.7 & 33.5/28.6/\textbf{37.8} &\multicolumn{1}{c|}{\textcolor{OliveGreen}{+12.3}}& \textbf{47.6}/30.6/21.9 & 28.3/30.6/\textbf{41.1} &\multicolumn{1}{c|}{\textcolor{OliveGreen}{+9.3}}& \textbf{40.8}/38.3/20.9 & \textbf{35.2}/35.2/29.5 &\multicolumn{1}{c}{\textcolor{OliveGreen}{+20.5}}
\cr& \multicolumn{1}{c|}{$\mathcal{P}_{\texttt{R}}$} & \textbf{43.1}/35.4/21.5 & 24.4/28.7/\textbf{47.0} &\multicolumn{1}{c|}{\textcolor{OliveGreen}{+0.9}}& \textbf{53.9}/26.1/20.0 & 32.5/26.9/\textbf{40.6} &\multicolumn{1}{c|}{\textcolor{OliveGreen}{+17.4}}& \textbf{49.5}/31.0/19.6 & \textbf{36.7}/36.7/26.7 &\multicolumn{1}{c}{\textcolor{OliveGreen}{+29.6}}
\\ \midrule
\multirow{2}{*}{$\mathcal{P}_{\texttt{R}}$} & \multicolumn{1}{c|}{$\mathcal{P}_{\texttt{C}}$} & \textbf{37.6}/36.4/26.1 & 26.7/31.1/\textbf{42.2} &\multicolumn{1}{c|}{\textcolor{BrickRed}{-2.2}}& \textbf{52.3}/28.3/19.4 & 30.1/26.9/\textbf{43.0} &\multicolumn{1}{c|}{\textcolor{OliveGreen}{+14.1}}& \textbf{40.1}/36.9/23.0 & 30.6/37.9/\textbf{31.5} &\multicolumn{1}{c}{\textcolor{OliveGreen}{+12.9}}
\cr& \multicolumn{1}{c|}{$\mathcal{P}_{\texttt{R}}$} & \textbf{43.7}/34.4/21.9 & 21.3/27.9/\textbf{50.8} &\multicolumn{1}{c|}{\textcolor{BrickRed}{-3.8}}& \textbf{61.4}/23.7/14.9 & 29.6/28.0/\textbf{42.5} &\multicolumn{1}{c|}{\textcolor{OliveGreen}{+21.5}}& \textbf{43.8}/36.8/19.4 & \textbf{38.6}/32.7/28.7 &\multicolumn{1}{c}{\textcolor{OliveGreen}{+26.7}}
\\ \midrule
\multirow{2}{*}{$\mathcal{P}_{\texttt{C}}$ + $\mathcal{P}_{\texttt{R}}$} & \multicolumn{1}{c|}{$\mathcal{P}_{\texttt{C}}$} & \textbf{42.9}/34.6/22.5 & 32.6/30.2/\textbf{37.2} &\multicolumn{1}{c|}{\textcolor{OliveGreen}{+12.3}}& \textbf{44.6}/31.5/24.0 & 29.6/22.8/\textbf{47.6} &\multicolumn{1}{c|}{\textcolor{OliveGreen}{+3.4}}& \textbf{40.3}/36.9/22.8 & \textbf{33.5}/35.7/30.8 &\multicolumn{1}{c}{\textcolor{OliveGreen}{+15.9}}
\cr& \multicolumn{1}{c|}{$\mathcal{P}_{\texttt{R}}$} & \textbf{45.9}/34.6/19.5 & 22.2/32.9/\textbf{44.9} &\multicolumn{1}{c|}{\textcolor{OliveGreen}{+3.2}}& \textbf{57.9}/25.7/16.4 & 24.5/28.7/\textbf{46.8} &\multicolumn{1}{c|}{\textcolor{OliveGreen}{+12.3}}& \textbf{45.8}/33.9/20.3 & \textbf{32.1}/39.0/28.9 &\multicolumn{1}{c}{\textcolor{OliveGreen}{+21.9}}
\\ \bottomrule
 \end{tabular}
}
 \caption{Response type personalization and quality judgments for supervised fine-tuning models that use chosen personas, rejected personas, and both personas for training and inference, compared to the supervised fine tuning model that does not use personas.
}
 \label{table:response_type_sft}
 \end{table*}

\begin{table*}[t]
\small
\centering
\setlength{\tabcolsep}{2pt}
\fontsize{8}{8}\selectfont{
\begin{tabular}{@{}ccccccccccc@{}}
\multicolumn{1}{l}{} & \multicolumn{1}{l}{}& \multicolumn{3}{c}{\textit{BeaverTails}}& \multicolumn{3}{c}{\textit{Anthropic HHH}}& \multicolumn{3}{c}{\textit{Mnemonic}} \\ \midrule
$\mathcal{P}_{train}$ & \multicolumn{1}{c|}{$\mathcal{P}_{inf}$} & Person. W/T/L & Quality W/T/L & \multicolumn{1}{c|}{\score} & Person. W/T/L & Quality W/T/L & \multicolumn{1}{c|}{\score} & Person. W/T/L & Quality W/T/L & \multicolumn{1}{c}{\score} \\ \midrule
\multirow{2}{*}{$\mathcal{P}_{\texttt{C}}$} & \multicolumn{1}{c|}{$\mathcal{P}_{\texttt{C}}$} & \textbf{72.1}/18.2/9.6 & 36.7/24.4/\textbf{38.9} &\multicolumn{1}{c|}{\textcolor{OliveGreen}{+36.8}}& \textbf{55.8}/25.0/19.2 & 25.4/25.2/\textbf{49.4} &\multicolumn{1}{c|}{\textcolor{OliveGreen}{+8.4}}& \textbf{64.4}/26.0/9.6 & 27.8/33.2/\textbf{39.0} &\multicolumn{1}{c}{\textcolor{OliveGreen}{+28.6}}
\cr& \multicolumn{1}{c|}{$\mathcal{P}_{\texttt{R}}$} & \textbf{70.3}/21.0/8.6 & 17.4/26.9/\textbf{55.7} &\multicolumn{1}{c|}{\textcolor{OliveGreen}{+12.9}}& \textbf{66.6}/21.6/11.8 & 25.2/27.2/\textbf{47.6} &\multicolumn{1}{c|}{\textcolor{OliveGreen}{+19.6}}& \textbf{66.0}/25.4/8.6 & 25.4/32.2/\textbf{42.4} &\multicolumn{1}{c}{\textcolor{OliveGreen}{+25.9}}
\\ \midrule
\multirow{2}{*}{$\mathcal{P}_{\texttt{R}}$} & \multicolumn{1}{c|}{$\mathcal{P}_{\texttt{C}}$} & \textbf{60.9}/24.2/14.8 & 10.2/21.4/\textbf{68.3} &\multicolumn{1}{c|}{\textcolor{BrickRed}{-6.6}}& \textbf{64.8}/21.0/14.2 & 8.8/14.2/\textbf{77.0} &\multicolumn{1}{c|}{\textcolor{BrickRed}{-7.7}}& \textbf{34.2}/33.4/32.4 & 16.8/30.4/\textbf{52.8} &\multicolumn{1}{c}{\textcolor{BrickRed}{-24.5}}
\cr& \multicolumn{1}{c|}{$\mathcal{P}_{\texttt{R}}$} & \textbf{58.9}/24.8/16.2 & 3.4/9.8/\textbf{86.8} &\multicolumn{1}{c|}{\textcolor{BrickRed}{-17.8}}& \textbf{71.0}/19.6/9.4 & 5.4/18.8/\textbf{75.8} &\multicolumn{1}{c|}{\textcolor{BrickRed}{-5.0}}& \textbf{34.6}/32.4/33.0 & 21.0/28.0/\textbf{51.0} &\multicolumn{1}{c}{\textcolor{BrickRed}{-19.6}}
\\ \midrule
\multirow{2}{*}{$\mathcal{P}_{\texttt{C}} + \mathcal{P}_{\texttt{R}}$} & \multicolumn{1}{c|}{$\mathcal{P}_{\texttt{C}}$} & \textbf{63.5}/22.8/13.6 & 13.4/16.4/\textbf{70.1} &\multicolumn{1}{c|}{\textcolor{BrickRed}{-1.6}}& \textbf{70.6}/15.0/14.4 & 12.2/18.0/\textbf{69.8} &\multicolumn{1}{c|}{\textcolor{BrickRed}{-2.1}}& \textbf{63.6}/24.8/11.6 & 21.8/29.6/\textbf{48.6} &\multicolumn{1}{c}{\textcolor{OliveGreen}{+15.5}}
\cr& \multicolumn{1}{c|}{$\mathcal{P}_{\texttt{R}}$} & \textbf{55.7}/27.7/16.6 & 3.6/9.6/\textbf{86.8} &\multicolumn{1}{c|}{\textcolor{BrickRed}{-19.0}}& \textbf{76.6}/15.2/8.2 & 9.6/18.2/\textbf{72.2} &\multicolumn{1}{c|}{\textcolor{OliveGreen}{+2.1}}& \textbf{61.6}/30.2/8.2 & 22.0/31.4/\textbf{46.6} &\multicolumn{1}{c}{\textcolor{OliveGreen}{+20.3}}
\\ \bottomrule
 \end{tabular}
}
 \caption{Response type personalization and quality judgments for direct preference optimization models that use chosen personas, rejected personas, and both personas for training and inference, compared to the direct preference optimization model that does not use personas.
}
 \label{table:response_type_dpo}
 \end{table*}
\begin{table*}[t]
\small
\centering
\setlength{\tabcolsep}{2pt}
\fontsize{8}{8}\selectfont{
\begin{tabular}{@{}ccccccccccc@{}}
\multicolumn{1}{l}{} & \multicolumn{1}{l}{}& \multicolumn{3}{c}{\textit{BeaverTails}}& \multicolumn{3}{c}{\textit{Mnemonic}}& \multicolumn{3}{c}{\textit{Anthropic HHH}} \\ \toprule
$\pi_{base}$ & \multicolumn{1}{c|}{$\pi_{test}$} & Person. W/T/L & Quality W/T/L & \multicolumn{1}{c|}{\score} & Person. W/T/L & Quality W/T/L & \multicolumn{1}{c|}{\score} & Person. W/T/L & Quality W/T/L & \multicolumn{1}{c}{\score} \\ \midrule
\multirow{1}{*}{\textsc{FS}} & \multicolumn{1}{c|}{$\textsc{PT}_\textsc{fs}$ + $\mathcal{P}_{syst}$} & \textbf{45.8}/35.8/18.3 & \textbf{45.0}/27.5/27.5 &\multicolumn{1}{c|}{\textcolor{OliveGreen}{+33.5}}& \textbf{32.7}/42.8/24.5 & 28.7/41.6/\textbf{29.7} &\multicolumn{1}{c|}{\textcolor{OliveGreen}{+6.4}}& \textbf{38.2}/32.9/28.9 & \textbf{40.8}/21.1/38.2 &\multicolumn{1}{c}{\textcolor{OliveGreen}{+8.5}}
\\ \midrule
\multirow{1}{*}{\textsc{SFT}} & \multicolumn{1}{c|}{$\textsc{PT}_\textsc{sft}$ + $\mathcal{P}_{syst}$} & \textbf{35.5}/36.4/28.1 & \textbf{30.6}/40.1/29.3 &\multicolumn{1}{c|}{\textcolor{OliveGreen}{+6.9}}& \textbf{31.9}/36.5/31.6 & \textbf{31.6}/39.0/29.4 &\multicolumn{1}{c|}{\textcolor{OliveGreen}{+2.0}}& 22.7/29.2/\textbf{48.1} & 13.4/26.4/\textbf{60.2} &\multicolumn{1}{c}{\textcolor{BrickRed}{-49.7}}
\\ \midrule
\multirow{1}{*}{\textsc{DPO}} & \multicolumn{1}{c|}{$\textsc{PT}_\textsc{dpo}$ + $\mathcal{P}_{syst}$} & \textbf{40.8}/33.8/25.4 & 33.8/32.3/\textbf{33.8} &\multicolumn{1}{c|}{\textcolor{OliveGreen}{+11.6}}& \textbf{58.9}/25.5/15.6 & 23.8/38.9/\textbf{37.2} &\multicolumn{1}{c|}{\textcolor{OliveGreen}{+18.1}}& 18.5/35.9/\textbf{45.7} & 19.6/37.0/\textbf{43.5} &\multicolumn{1}{c}{\textcolor{BrickRed}{-40.2}}
\\ \bottomrule
 \end{tabular}
}
 \caption{On BeaverTails and Mnemonic, adding a fixed system prompt as the persona for inference typically improves both personalization and quality across training strategies.
}
 \label{table:system_prompt}
 \end{table*}

\begin{table*}[t]
\small
\centering
\setlength{\tabcolsep}{1.5pt}
\fontsize{8}{8}\selectfont{
\begin{tabular}{@{}ccccccccccc@{}}
\multicolumn{1}{l}{} & \multicolumn{1}{l}{}& \multicolumn{3}{c}{\textit{BeaverTails}}& \multicolumn{3}{c}{\textit{Anthropic HHH}}& \multicolumn{3}{c}{\textit{Mnemonic}} \\ \midrule
$\pi_{base}$ & \multicolumn{1}{c|}{$\pi_{test}$} & Person. W/T/L & Quality W/T/L & \multicolumn{1}{c|}{\score} & Person. W/T/L & Quality W/T/L & \multicolumn{1}{c|}{\score} & Person. W/T/L & Quality W/T/L & \multicolumn{1}{c}{\score} \\ \midrule
$\textsc{PT}_\textsc{dpo}$ & \multicolumn{1}{c|}{L-405B} & \textbf{50.3}/9.81/39.9 & 39.9/14.2/\textbf{45.9} &\multicolumn{1}{c|}{\textcolor{OliveGreen}{+2.27}}
& \textbf{90.2}/11.4/5.60 & \textbf{83.0}/11.4/5.60 &\multicolumn{1}{c|}{\textcolor{OliveGreen}{+91.0}}
& \textbf{38.2}/32.9/28.9 & \textbf{48.0}/28.8/23.2 &\multicolumn{1}{c}{\textcolor{OliveGreen}{+24.3}}
\\ \bottomrule
 \end{tabular}
}
 \caption{Comparison of persona tailoring with DPO and few-shot prompted LLaMA-405B, both using retrieved chosen personas. Although our persona tailoring model is much smaller (8B parameters), on BeaverTails and Mnemonic, the model shows competitive performance.
}
 \label{table:teacher_student}
 \end{table*}

\begin{table*}[t]
\small
\centering
\begin{tabular}{@{}cccccc@{}}
\\ \toprule
Dataset & $\pi_{base}$ & \multicolumn{1}{c}{$\pi_{test}$} & Person. W/T/L & Quality W/T/L & \multicolumn{1}{c}{\score} \\ \midrule
\multirow{2}{*}{\begin{tabular}{@{}c@{}}\textit{BT} \\ \texttt{\textcolor{blue}{Chosen}}\end{tabular}} & \multirow{1}{*}{\textsc{DPO}+$\mathcal{P}_{retr}$} & \multicolumn{1}{c}{\textsc{PT}+$\mathcal{P}_{retr}$} & \textbf{46.7}/29.3/24.0 & \textbf{38.5}/30.5/31.1 &\multicolumn{1}{c}{\textcolor{OliveGreen}{+21.3}} \\
& \multirow{1}{*}{\textsc{DPO}+$\mathcal{P}_{gold}$} & \multicolumn{1}{c}{\textsc{PT}+$\mathcal{P}_{gold}$} & \textbf{42.3}/29.3/28.5 & \textbf{34.9}/33.9/31.3 &\multicolumn{1}{c}{\textcolor{OliveGreen}{+12.5}} \\ \midrule

\multirow{2}{*}{\begin{tabular}{@{}c@{}}\textit{BT} \\ \texttt{\textcolor{red}{Reject}}\end{tabular}} & \multirow{1}{*}{\textsc{DPO}+$\mathcal{P}_{retr}$} & \multicolumn{1}{c}{\textsc{PT}+$\mathcal{P}_{retr}$} & \textbf{45.1}/31.7/23.2 & \textbf{35.1}/32.5/32.5 &\multicolumn{1}{c}{\textcolor{OliveGreen}{+17.9}} \\
& \multirow{1}{*}{\textsc{DPO}+$\mathcal{P}_{gold}$} & \multicolumn{1}{c}{\textsc{PT}+$\mathcal{P}_{gold}$} & \textbf{51.1}/25.9/23.0 & \textbf{35.3}/32.7/32.1 &\multicolumn{1}{c}{\textcolor{OliveGreen}{+21.3}} \\ \midrule

\multirow{2}{*}{\begin{tabular}{@{}c@{}}\textit{HHH} \\ \texttt{\textcolor{blue}{Chosen}}\end{tabular}} & \multirow{1}{*}{\textsc{DPO}+$\mathcal{P}_{retr}$} & \multicolumn{1}{c}{\textsc{PT}+$\mathcal{P}_{retr}$} & \textbf{37.2}/22.6/40.2 & \textbf{32.6}/21.3/46.0 &\multicolumn{1}{c}{\textcolor{BrickRed}{-10.4}} \\
& \multirow{1}{*}{\textsc{DPO}+$\mathcal{P}_{gold}$} & \multicolumn{1}{c}{\textsc{PT}+$\mathcal{P}_{gold}$} & \textbf{32.6}/27.9/39.5 & \textbf{30.4}/29.0/40.6 &\multicolumn{1}{c}{\textcolor{BrickRed}{-11.9}} \\ \midrule

\multirow{2}{*}{\begin{tabular}{@{}c@{}}\textit{HHH} \\ \texttt{\textcolor{red}{Reject}}\end{tabular}} & \multirow{1}{*}{\textsc{DPO}+$\mathcal{P}_{retr}$} & \multicolumn{1}{c}{\textsc{PT}+$\mathcal{P}_{retr}$} & \textbf{48.0}/21.8/30.1 & \textbf{39.3}/25.8/34.9 &\multicolumn{1}{c}{\textcolor{OliveGreen}{+14.4}} \\
& \multirow{1}{*}{\textsc{DPO}+$\mathcal{P}_{gold}$} & \multicolumn{1}{c}{\textsc{PT}+$\mathcal{P}_{gold}$} & \textbf{50.8}/20.7/28.5 & \textbf{43.0}/23.4/33.6 &\multicolumn{1}{c}{\textcolor{OliveGreen}{+20.2}} \\

\bottomrule
\end{tabular}
\caption{Comparison of personalization abilities of \textsc{DPO} and $\textsc{PT}_\textsc{dpo}$ when using the full Anthropic HHH and BT datasets. $\textsc{PT}_\textsc{dpo}$ still improves personalization on the rejected personas, but is slightly worse on the chosen personas. This is likely because DPO trained on chosen responses can already generate responses that tailor to chosen personas.}
\label{table:response_type_hhh}
\end{table*}

\begin{table*}[t]
\small
\centering
\renewcommand{\arraystretch}{0.8}
\setlength{\tabcolsep}{2pt}
\fontsize{8}{8}\selectfont{
\begin{tabular}{@{}ccccccccccc@{}}
\multicolumn{1}{l}{} & \multicolumn{1}{l}{}& \multicolumn{3}{c}{\textit{BeaverTails}}& \multicolumn{3}{c}{\textit{Anthropic HHH}}& \multicolumn{3}{c}{\textit{Mnemonic}} \\ \toprule
$\pi_{base}$ & \multicolumn{1}{c|}{$\pi_{test}$} & Person. W/T/L & Quality W/T/L & \multicolumn{1}{c|}{\score} & Person. W/T/L & Quality W/T/L & \multicolumn{1}{c|}{\score} & Person. W/T/L & Quality W/T/L & \multicolumn{1}{c}{\score} \\ \midrule
\multirow{2}{*}{\textsc{FS}} & \multicolumn{1}{c|}{$\textsc{PT}_{\textsc{fs}}$+$\mathcal{P}_{\texttt{retr}}$} & \textbf{62.5}/17.2/20.2 & \textbf{60.7}/14.2/25.1 &\multicolumn{1}{c|}{\textcolor{OliveGreen}{+46.3}}& \textbf{47.5}/20.9/31.6 & 41.8/15.6/\textbf{42.6} &\multicolumn{1}{c|}{\textcolor{OliveGreen}{+9.6}}& \textbf{44.3}/28.5/27.2 & \textbf{46.4}/20.5/33.1 &\multicolumn{1}{c}{\textcolor{OliveGreen}{+20.3}}
\cr& \multicolumn{1}{c|}{$\textsc{PT}_{\textsc{fs}}$+$\mathcal{P}_{\texttt{gold}}$} & \textbf{68.7}/14.5/16.9 & \textbf{62.9}/15.9/21.3 &\multicolumn{1}{c|}{\textcolor{OliveGreen}{+55.0}}& \textbf{57.3}/20.2/22.5 & \textbf{51.5}/15.6/32.8 &\multicolumn{1}{c|}{\textcolor{OliveGreen}{+32.9}}& --- & --- &\multicolumn{1}{c}{---}
\\ \midrule
\multirow{2}{*}{\textsc{SFT}} & \multicolumn{1}{c|}{$\textsc{PT}_{\textsc{ft}}$+$\mathcal{P}_{\texttt{retr}}$} & \textbf{44.6}/31.7/23.7 & 33.5/28.6/\textbf{37.8} &\multicolumn{1}{c|}{\textcolor{OliveGreen}{+12.3}}& \textbf{52.3}/28.9/18.8 & 35.2/24.2/\textbf{40.6} &\multicolumn{1}{c|}{\textcolor{OliveGreen}{+20.0}}& \textbf{40.8}/38.3/20.9 & \textbf{35.2}/35.2/29.5 &\multicolumn{1}{c}{\textcolor{OliveGreen}{+20.5}}
\cr& \multicolumn{1}{c|}{$\textsc{PT}_{\textsc{sft}}$+$\mathcal{P}_{\texttt{gold}}$} & \textbf{46.7}/32.0/21.2 & \textbf{38.2}/29.6/32.2 &\multicolumn{1}{c|}{\textcolor{OliveGreen}{+23.0}}& \textbf{62.2}/23.7/14.1 & \textbf{41.7}/29.5/28.8 &\multicolumn{1}{c|}{\textcolor{OliveGreen}{+40.6}}& --- & --- &\multicolumn{1}{c}{---}
\\ \midrule
\multirow{2}{*}{\textsc{DPO}} & \multicolumn{1}{c|}{$\textsc{PT}_{\textsc{dpo}}$+$\mathcal{P}_{\texttt{retr}}$} & \textbf{72.1}/18.2/9.6 & 36.7/24.4/\textbf{38.9} &\multicolumn{1}{c|}{\textcolor{OliveGreen}{+36.8}}& \textbf{54.1}/26.1/19.8 & 21.0/23.7/\textbf{55.3} &\multicolumn{1}{c|}{\textcolor{OliveGreen}{+0.7}}& \textbf{64.4}/26.0/9.6 & 27.8/33.2/\textbf{39.0} &\multicolumn{1}{c}{\textcolor{OliveGreen}{+28.6}}
\cr& \multicolumn{1}{c|}{$\textsc{PT}_{\textsc{dpo}}$+$\mathcal{P}_{\texttt{gold}}$} & \textbf{66.3}/21.4/12.2 & \textbf{40.9}/28.5/30.7 &\multicolumn{1}{c|}{\textcolor{OliveGreen}{+41.6}}& \textbf{50.8}/30.5/18.6 & 28.8/25.8/\textbf{45.4} &\multicolumn{1}{c|}{\textcolor{OliveGreen}{+12.0}}& --- & --- &\multicolumn{1}{c}{---}
\\ \bottomrule
 \end{tabular}
}

 \caption{ Win, tie, and loss rates of generation methods ($\textsc{FS}$, $\textsc{SFT}$, $\textsc{DPO}$)
with and without personas $\mathcal{P}$ in pairwise comparisons from the Prometheus judge when using the filtered Anthropic HHH dataset. Our results are still strong compared to Table~\ref{table:add_persona}.
}
\label{table:add_persona_hhh}
\end{table*}

\end{document}